%% file: Arxiv.tex
\documentclass[11pt]{article}

\date{}

\input{math_commands.tex}

\usepackage[utf8]{inputenc} %
\usepackage[T1]{fontenc}    %
\usepackage[colorlinks=true,linkcolor=blue,allcolors=blue]{hyperref}       %
\usepackage{url}            %
\usepackage{booktabs}       %
\usepackage{amsfonts}       %
\usepackage{nicefrac}       %
\usepackage{microtype,nicefrac}      %
\usepackage{xspace}
\usepackage{natbib}

\usepackage{epsfig}
\usepackage{graphicx}
\usepackage{wrapfig}
\usepackage{amsmath}
\usepackage{amssymb}
\usepackage{subcaption}

\usepackage[margin=1in]{geometry}
\usepackage{longtable}
\usepackage{booktabs}
\usepackage{bbm}
\usepackage{adjustbox}
\usepackage{multirow}
\def\1{\mathbbm{1}}
\usepackage{placeins}

\usepackage[toc]{appendix}
\usepackage{minitoc}
\allowdisplaybreaks
\makeatletter

\title{Exploring the design space of deep-learning-based weather forecasting systems}

\author{ Shoaib Ahmed Siddiqui \\ \texttt{msas3@cam.ac.uk} \\
 University of Cambridge, UK 
\and
 Jean Kossaifi \\ \texttt{jkossaifi@nvidia.com} \\
 NVIDIA Research, USA 
\and
 Boris Bonev \\ \texttt{bbonev@nvidia.com} \\
 NVIDIA Research, Switzerland 
\and
 Christopher Choy \\ \texttt{cchoy@nvidia.com} \\
 NVIDIA Research, USA 
\and
 Jan Kautz \\ \texttt{jkautz@nvidia.com} \\
 NVIDIA Research, USA 
\and
 David Krueger \\ \texttt{dsk30@cam.ac.uk} \\
 University of Cambridge, UK 
\and
 Kamyar Azizzadenesheli \\ \texttt{kamyara@nvidia.com} \\
 NVIDIA Research, USA}

\newcommand{\fix}{\marginpar{FIX}}

\begin{document}
\doparttoc %
\faketableofcontents %

\maketitle

\input{sections/abstract}
\input{sections/introduction}
\input{sections/related_work}
\input{sections/methods}
\input{sections/experiments}
\input{sections/conclusion}
\input{sections/acknowledgements}

\bibliography{main}
\bibliographystyle{tmlr}

\input{sections/appendix}

\end{document}


\title{Supporting Information for "Insert Title"}

\authors{=Authors=}

\affiliation{=number=}{=Affiliation Address=}

\begin{article}

\noindent\textbf{Contents of this file}
\begin{enumerate}
\item Text S1 to Sx
\item Figures S1 to Sx
\item Tables S1 to Sx
\end{enumerate}
\noindent\textbf{Additional Supporting Information (Files uploaded separately)}
\begin{enumerate}
\item Captions for Datasets S1 to Sx
\item Captions for large Tables S1 to Sx (if larger than 1 page, upload as separate excel file)
\item Captions for Movies S1 to Sx
\item Captions for Audio S1 to Sx
\end{enumerate}

\noindent\textbf{Introduction}

\noindent\textbf{Text S1.}

\noindent\textbf{Data Set S1.} %

\noindent\textbf{Movie S1.} %

\noindent\textbf{Audio S1.} %

\end{article}
\clearpage

%% file: math_commands.tex
\usepackage{amsmath,amsfonts,bm}

\def\eqref#1{equation~\ref{#1}}

\def\1{\bm{1}}

\DeclareMathAlphabet{\mathsfit}{\encodingdefault}{\sfdefault}{m}{sl}
\SetMathAlphabet{\mathsfit}{bold}{\encodingdefault}{\sfdefault}{bx}{n}

%% file: sections/abstract.tex
\begin{abstract}
Despite tremendous progress in developing deep-learning-based weather forecasting systems, their design space, including the impact of different design choices, is yet to be well understood.
This paper aims to fill this knowledge gap by systematically analyzing these choices including architecture, problem formulation, pretraining scheme, use of image-based pretrained models, loss functions, noise injection, multi-step inputs, additional static masks, multi-step finetuning (including larger stride models), as well as training on a larger dataset.
We study fixed-grid architectures such as UNet, fully convolutional architectures, and transformer-based models, along with grid-invariant architectures, including graph-based and operator-based models.
Our results show that fixed-grid architectures outperform grid-invariant architectures, indicating a need for further architectural developments in grid-invariant models such as neural operators.
We therefore propose a hybrid system that combines the strong performance of fixed-grid models with the flexibility of grid-invariant architectures. 
We further show that multi-step fine-tuning is essential for most deep-learning models to work well in practice, which has been a common practice in the past.
Pretraining objectives degrade performance in comparison to supervised training, while image-based pretrained models provide useful inductive biases in some cases in comparison to training the model from scratch.
Interestingly, we see a strong positive effect of using a larger dataset when training a smaller model as compared to training on a smaller dataset for longer.
Larger models, on the other hand, primarily benefit from just an increase in the computational budget.
We believe that these results will aid in the design of better weather forecasting systems in the future.
\end{abstract}

%% file: sections/introduction.tex
\section{Introduction}

Deep learning for weather forecasting has been a very active area of research in the recent past, with many newer deep-learning-based models approaching the performance of numerical weather prediction systems while being orders of magnitude faster and energy efficient~\citep{pathak2022fourcastnet,bi2022pangu,lam2022graphcast,price2023gencast}.
Despite these advances, the justification for the design decisions, including model architectures, is rarely described~\citep{pathak2022fourcastnet,bi2022pangu,lam2022graphcast,price2023gencast}. While it is likely that most of these decisions are based on experimental evidence that was not directly presented, comparing these choices on an even playing field is important to understand the marginal contribution of each of these design choices on the resulting model.

\begin{figure}[t]
    \centering
    \includegraphics[width=0.49\textwidth,clip]{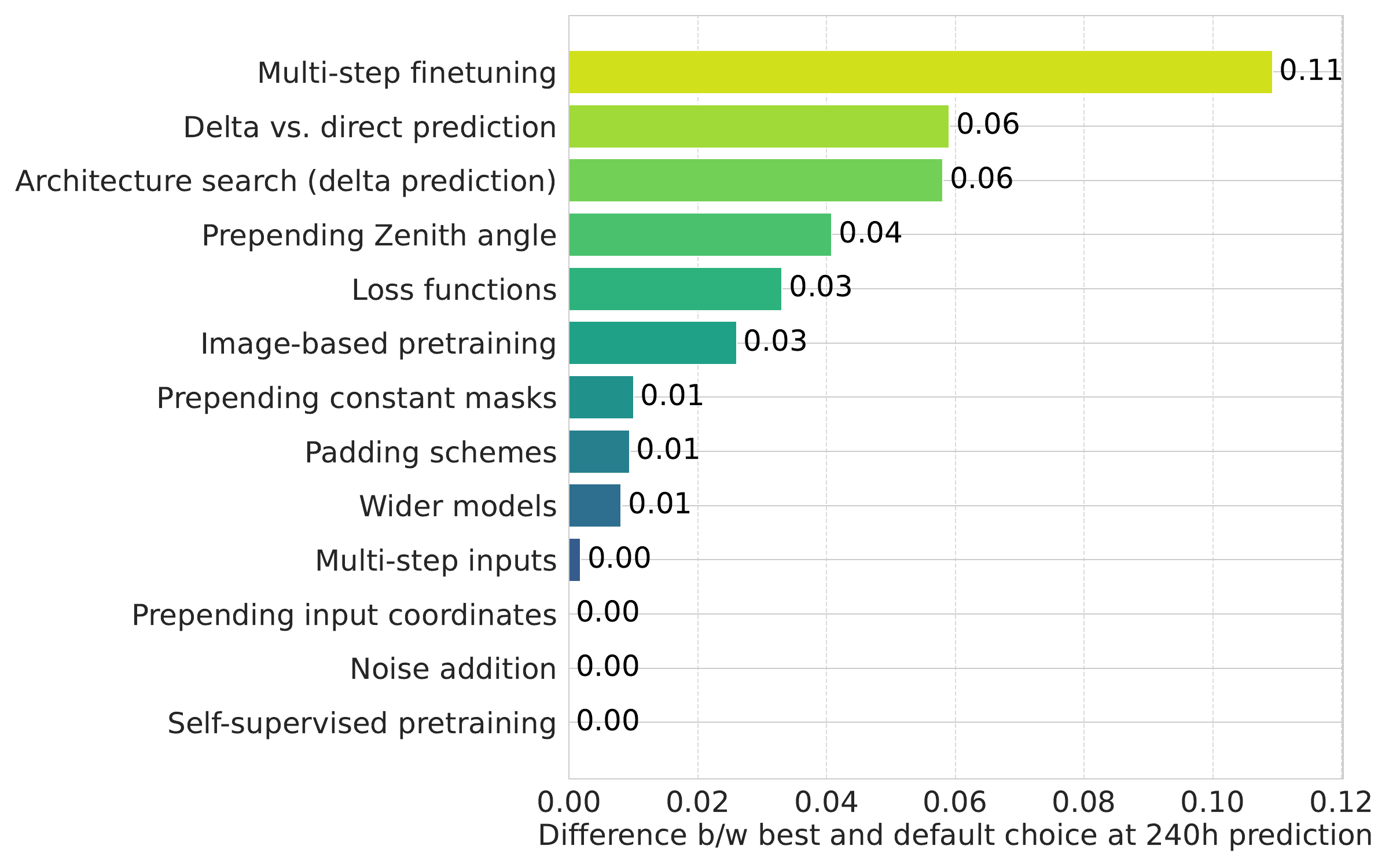}
    \includegraphics[width=0.49\textwidth]{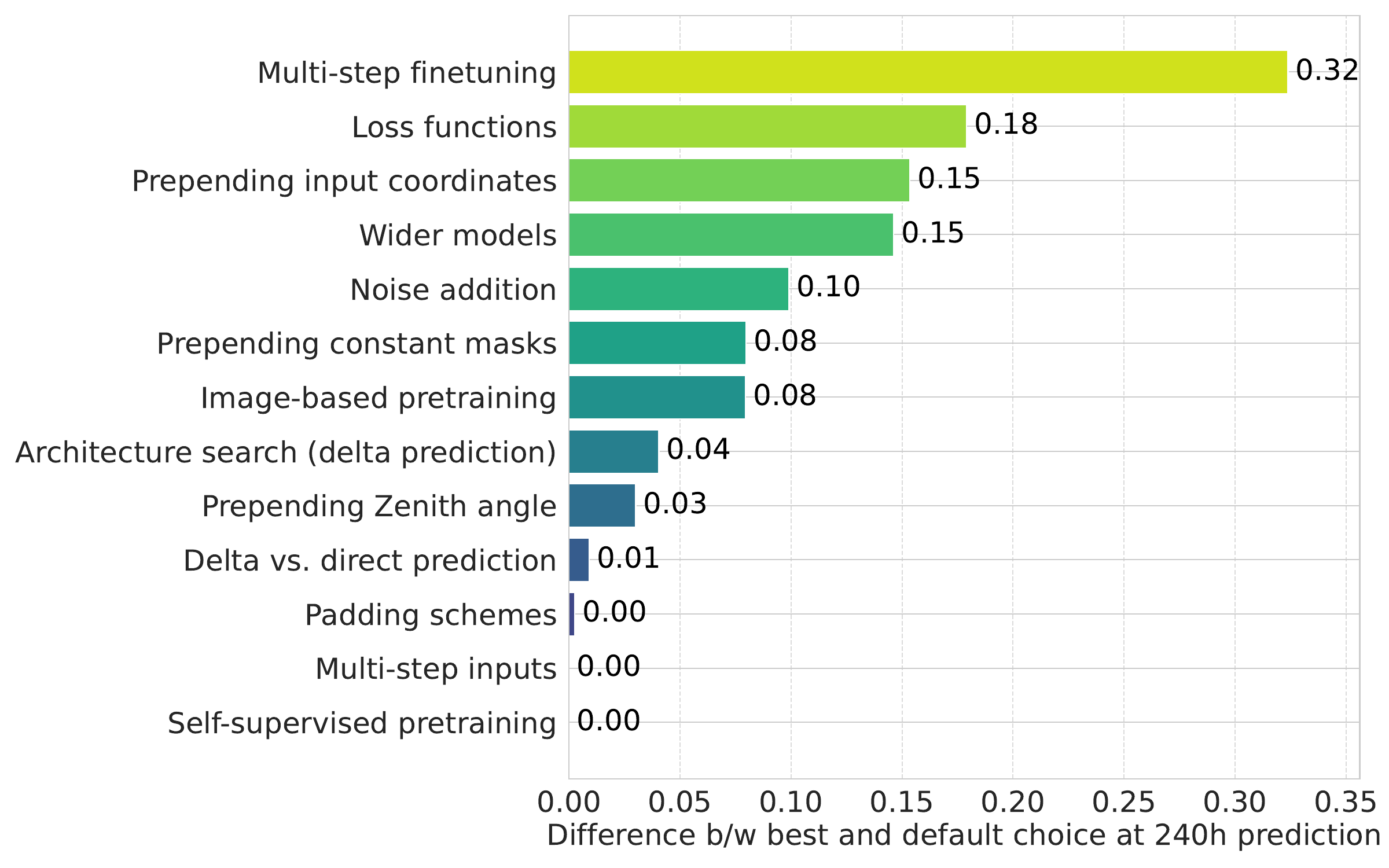}
    \caption{\textbf{Geometric ACC (left) and RMSE (right) for comparison between design choices.} This figure highlights the marginal contribution between the default choice (primarily based on our default 4-block UNet) and the best possible choice when considering different design decisions explored in this work. In cases where the default decision turned out to be the best, the marginal contribution is zero. Interestingly, there is a significant difference in the marginal contribution of different design choices between the two metrics.}
    \label{fig:overview_fig}
\end{figure}

In this work, we take a step back and ask: \textit{which of these choices are important and relevant for deep-learning-based weather forecasting systems?}
To answer this question, we evaluate different aspects of the design space given a fixed number of training steps, including output formulation (direct, or residual prediction), model architecture, pretraining objective, image-based pretraining, loss functions, noise-based augmentations, zenith angle, constant masks, multi-step inputs, multi-step fine-tuning, as well as training on larger datasets.
Our results predominantly indicate that highly-optimized computer-vision-based fixed-grid models are still effective for weather forecasting when evaluated on fixed grids, and the final performance of the models is more sensitive to some choices compared to others. These results call for extensive advancements in operator learning architecture design, for which we propose a new architecture through directly generalizing fixed-grid models~\citep{liu2024neural,li2024geometry}.

An overview of our results is presented in Fig.~\ref{fig:overview_fig} where we compare the best choice in each category with the default choice (which is primarily based on our 4-block UNet architecture).
Our results show that some of these choices have a non-trivial impact on the resulting model.
Furthermore, we see that the two metrics i.e., geometric ACC and RMSE behave differently under different choices.
In particular, our findings can be summarized as:

\begin{itemize}
    \item For short-horizon prediction such as 6h considered in this work, training the model to predict the residual to the input state is much more efficient as compared to directly predicting the output for a range of different architectures (Section~\ref{subsec:results_formulation}).
    \item Fixed-grid architectures including convolutional (such as UNet~\citep{ronneberger2015unet}, and ResNet-50 with fully convolutional decoder~\citep{long2015fcn}) and transformer-based models (such as Segformer~\citep{zheng2021setr}, and SETR~\citep{zheng2021setr}) are more performant for weather forecasting on ERA5 dataset in contrast to grid-invariant models including graph-based~\citep{lam2022graphcast}, and operator-based~\citep{pathak2022fourcastnet,bonev2023sfno} models when considering a fixed computational budget of 10 epochs, indicating a great need for further architectural developments in grid-invariant models. We therefore present a hybrid architecture that combines the flexibility of grid-invariant architectures with the performance of grid-invariant models through a direct generalization of fixed-grid models~\citep{liu2024neural,li2024geometry} (Section~\ref{subsec:results_arch}).
    \item Supervised training achieves superior performance compared to self-supervised pretraining using different objectives such as auto-encoding~\citep{erhan2009denoisingae}, denoising auto-encoding~\citep{erhan2009denoisingae}, or masked auto-encoding~\citep{he2022masked} (Section~\ref{subsec:results_pretraining_schemes}).
    \item Image-based pertaining can provide significant performance gains in some cases, specifically for transformers, as the model can still leverage its understanding of locality and attention while reinitializing the input and output heads to match the dimensionality of the ERA5 inputs as compared to RGB images (Section~\ref{subsec:results_image_pretraining}).
    \item Taking multiple previous input steps as input can increase the model's propensity to overfit the task, and hence, degrade performance rather than improve it in many cases (Section~\ref{subsec:results_multi_step_inputs}).
    \item Zenith angle captures the location of the sun w.r.t. the query location, and hence, provides a notion of time for the model. Therefore, providing the zenith angle as an additional channel to the model provides significant gains at all prediction horizons (Section~\ref{subsec:results_zenith}).
    \item Information about the geometry of the sphere can be encoded in the form of padding schemes used for image-based models. Using the circular pad for the longitude (x-axis) and zero/reflection padding scheme for the latitude (y-axis) for convolutional layers provides the best performance (Section~\ref{subsec:results_padding_scheme}).
    \item Injection of noise during training (which was primarily used for ensemble generation in prior work~\citep{pathak2022fourcastnet,bi2022pangu}) can improve forecasting performance by increasing the model's stability during long auto-regressive rollouts (Section~\ref{subsec:results_noise_addition}).
    \item Providing information regarding the 3D coordinates on the sphere provides stability in the long-horizon auto-regressive rollouts potentially due to the model leveraging information regarding oversampling at the poles (Section~\ref{subsec:results_input_loc}).
    \item Variants of $L_1$ loss function are more stable when doing longer auto-regressive rollouts as compared to MSE potentially due to them being less susceptible to outliers (Section~\ref{subsec:results_loss_fn}).
    \item Inclusion of constant masks such as topography, soil type, and land-sea mask as commonly included in prior work~\citep{bi2022pangu} mostly degrades performance, potentially due to an increased propensity of the model to overfit (Section~\ref{subsec:results_constant_masks}).
    \item Wider models outperform their narrower counterparts, similar to the findings in the computer vision literature~\citep{zagoruyko2016wideresnets}  (Section~\ref{subsec:results_wider_models}).
    \item Multi-step fine-tuning on longer horizons helps in reducing model artifacts during long auto-regressive rollouts. This sequential fine-tuning outperforms models trained directly on a larger temporal resolution, as proposed in~\citet{bi2022pangu}. Furthermore, intermediate-step supervision is useful in some cases (particularly for our larger model) as compared to just last-step supervision (Section~\ref{subsec:results_multistep_ft}).
    \item Increasing the amount of data used for training improves performance. In some cases, this gain in performance is proportional to the gain in performance achieved by appropriately scaling the number of training epochs on the smaller dataset to match the total training steps in the two cases (Section~\ref{subsec:results_hourly_training}).
\end{itemize}

We believe that our findings would enable a better understanding of the design space of deep-learning-based weather forecasting systems and hence, the development of more efficient and effective systems in the future.

%% file: sections/related_work.tex
\section{Related Work}

\citet{weyn2020forecastconvnet} established some of the initial results on deep-learning-based weather forecasting. Their model was trained on $2^{\circ} \times 2^{\circ}$, and consumed two previous input states to produce a forecast for the next two states, each with a 6h stride.
Their architecture was based on UNet, borrowing motivation from \citet{larraondo2019weatherforecastunet} who compared three different encoder-decoder segmentation models to establish the utility of the UNet model for predicting precipitation fields given only geopotential height field on small input size of $80 \times 120$ (focusing only on Europe).
\citet{rasp2021resnetweather} used a ResNet model on WeatherBench dataset~\citep{rasp2020weatherbench}, achieving state-of-the-art performance for a data-driven method at a small resolution of $5.625^{\circ}$ and a forecasting horizon of 5 days.
These works showed that deep-learning-based models fall short of the performance of the operational IFS system.

FourCastNet~\citep{pathak2022fourcastnet} was one of the first deep-learning-based weather forecasting systems to match the performance of the operational IFS system on spatially high-resolution inputs ($0.25^{\circ} \times 0.25^{\circ}$) while being orders of magnitude faster at inference time. FourCastNet was based on (Adaptive) Fourier Neural Operator~\citep{li2020fno,guibas2021afno} which is particularly designed to learn solution operators of differential equations, owing to its ability to be grid invariant.
The model was trained for 6h prediction and fine-tuned for multi-step auto-regressive prediction.
They trained the model using latitude-weighted MSE and summed up the losses from the two-step rollout during fine-tuning.
They explored using additive Gaussian noise to the input state to generate an ensemble prediction~\citep{graubner2022calibration}.

Pangu-weather~\citep{bi2022pangu} was based on the Swin transformer architecture~\citep{liu2021swin}, replacing the general position embedding with an earth-specific positional embedding.
They trained 4 models with different prediction horizons including 6h, 12h, 18h, and 24h.
The final prediction was based on a greedy combination of these fixed-stride models.
The model was trained for 100 epochs on hourly sampled data.
An ensemble of model predictions was constructed by adding Perlin~\citep{perlin2002noise} noise to the input states.

Spherical Fourier Neural Operator (SFNO)~\citep{bonev2023sfno} proposed an extension of the Fourier neural operator based on spherical harmonics, taking into account the geometrical properties of Earth, which ultimately resulted in more stable prediction over longer horizons. Such a model is discretization agnostic, meaning it can be trained on various discretization of input functions, and the output functions can be queried at any discretization.
The model was again pretrained for 6h prediction, and subsequently fine-tuned for multi-step auto-regressive predictions.

FuXi~\citep{chen2023fuxi} focused on a 15-day forecasting task, and proposed a cascaded ML weather forecasting system that produced 15-day global forecasting on par with the ECMWF ensemble mean at a temporal resolution of 6h and spatial resolution of $0.25^{\circ} \times 0.25^{\circ}$.
FuXi ensemble was created by perturbing both the initial states using Perlin noise~\citep{perlin2002noise} as well as the model parameters using dropout~\citep{srivastava2014dropout}.
FuXi used the two previous time-steps to forecast the next step.
Due to the 6h forecasting horizon, it required 60 auto-regressive steps for the model to generate a 15-day forecast.
Focusing on previous work which argued that a single model is sub-optimal for both short-term and long-term horizons, they proposed a cascade model architecture using pretrained FuXi models finetuned for optimal performance in specific 5-day forecast time windows: FuXi short (0-5 days), FuXi medium (5-10 days), and FuXi long (10-15 days).
FuXi used cube embedding (similar to PanguWeather where they used cube patches) and a U-transformer~\citep{petit2021utransformer}.
They trained the model for 40k steps on 8 GPUs.

Based on the success of graph-based message-passing networks~\citep{gilmer2017mpgnn}, \citet{lam2022graphcast} presented GraphCast model powered by a message-passing GNN, using a fixed multi-level mesh for efficient message-passing at multiple levels~\citep{li2020multipole}.
GraphCast used 5 surface variables and 6 variables at 37 pressure levels, resulting in a total of 227 input channels.
GraphCast took two previous time-steps as input and was trained for 300k steps with one auto-regressive step and subsequently fine-tuned for 1000 training steps with an increasing number of auto-regressive steps up to 12 steps (equivalent to a 3-day rollout).

More recently, GenCast~\citep{price2023gencast} trained GraphCast architecture~\citep{lam2022graphcast} using the diffusion objective, providing probabilistic predictions instead of just deterministic outputs.
The noise for the diffusion process is sampled from a sphere.
Training a diffusion model produces detailed outputs with physically consistent power spectra, and avoids the over-smoothing of predictions returned by GraphCast (and all other deterministic forecasting systems) at longer prediction horizons.
They trained this model on a lower resolution of $1^{\circ} \times 1^{\circ}$ with a 12h prediction horizon using two previous steps as input.
They used only 6 variables at 13 pressure levels instead of the 37 levels used in the original GraphCast model.
One surprising thing to note is that this formulation allowed the model to work well even without multi-step fine-tuning.
Similarly,~\citet{li2024generative} used diffusion models as probabilistic forecasting systems, allowing natural generation of prediction ensembles using diffusion sampling in contrast to input state perturbation as done in prior work~\citep{pathak2022fourcastnet,bi2022pangu}.

NeuralGCM~\citep{kochkov2023neuralgcm} attempted to learn General Circulation Models (GCMs), and proposed splitting the forecasting system into a dynamical core that simulates large-scale fluid motion and thermodynamics, and a learned physics model that predicts the effects of unresolved processes such as cloud formulation and precipitation, which are then fed to an ODE solver.
This achieves performance comparable to other ML-based systems at shorter horizons (1-10 days), while still being effective at climate modeling spanning over multiple decades.

CoreDiff~\citep{mardani2023residual} proposed a two-step regional forecasting system where a UNet first predicts the mean, which is subsequently corrected by a diffusion model.
This formulation allows them to focus on km-scale atmospheric downscaling, while prior works focused on $\sim 27.75$ km global weather prediction ($0.25^{\circ} \times 0.25^{\circ}$).
Many national weather agencies couple km-scale numerical weather models in a limited domain to coarser resolution models, known as `dynamical downscaling`.

DiffDA~\citep{huang2024diffda} is a method designed for data assimilation that assimilates atmospheric variables using predicted states and sparse observations.
They used GraphCast as a pretrained model for diffusion.
DiffDA was designed to produce assimilated global atmospheric data consistent with observations at $0.25^{\circ} \times 0.25^{\circ}$ resolution globally.
They used a conditional diffusion model where conditioning is on the current state, and the target is to predict the state after taking into account additional sparse observations.
They considered 6 pressure-level variables and 4 surface variables, with a horizontal resolution of $0.25^{\circ}$ and 13 vertical levels $13 \times 6 + 4 = 82$ which matches the resolution of the WeatherBench-2~\citep{rasp2023weatherbench} dataset.
The model was trained for 20 epochs on 48 NVIDIA A100 GPUs using a batch size of 48.

\citet{bodnar2024aurora} proposed `Aurora', which is a foundation model for atmospheric modeling, including weather forecasting, leveraging 3D Swin Transformer~\citep{liu2021swin}.
The model is simultaneously trained on multiple datasets with different resolutions, variables, and pressure levels.
The model is then fine-tuned using LoRA for downstream tasks.
Aurora~\citep{bodnar2024aurora} outperforms GraphCast~\citep{lam2022graphcast} on the vast majority of targets.

\citet{vaughan2024aardvark} proposed an end-to-end weather forecasting system, `Aardvark`, to replace the entire operational NWP pipeline.
Aardvark directly ingests raw observations and is capable of outputting global gridded forecasts, as well as local station forecasts.

In contrast to all prior work that attempts to propose the best predictive model for weather forecasting, our aim is to develop a thorough understanding of the different design decisions involved in designing these systems and their potential impact on weather forecasting performance.

%% file: sections/methods.tex
\section{Methods}
\label{sec:methods}

\subsection{Models}

Our evaluation considers a range of different architectures.
We use custom UNet implementation~\citep{ronneberger2015unet}, UNO~\citep{ashiqur2022uno} implementation from NeuralOperator package\footnote{\url{https://neuraloperator.github.io/neuraloperator/dev/index.html}}, official SFNO implementation from the Makani codebase\footnote{\url{https://github.com/NVIDIA/modulus-makani}},
FourCastNet~\citep{pathak2022fourcastnet}, and GraphCast~\citep{lam2022graphcast} from Modulus package\footnote{\url{https://github.com/NVIDIA/modulus}}, official pseudo-code translated to PyTorch~\citep{paszke2019pytorch} for Pangu-weather~\citep{bi2022pangu}, official code for Point Transformer v3~\citep{wu2023pointtranformerv3}, Octformer~\citep{wang2023octformer}, ISNet~\citep{qin2022isnet}, and Efficient ViT L2~\citep{cai2022efficientvit}, and MMSeg~\citep{mmseg2020} implementation of SETR~\citep{zheng2021setr}, ResNet-50 with fully-convolutional decoder~\citep{long2015fcn}, and Segformer (B5)~\citep{xie2021segformer}.
Note that some architectures operate on patches, hence producing a lower resolution output (one prediction per patch rather than per pixel).
We add a very lightweight upsampling network in these cases to upsample the outputs to match the input resolution.

We emphasize that prior work based on neural operators learn the operator map between function spaces and are capable of training and testing on various resolutions, a capability unique to these models. In this work, we mainly focus on a fixed resolution and do not explore the operator learning aspect of weather forecast. 

\subsection{Dataset}

We use $0.25^{\circ} \times 0.25^{\circ}$ resolution ERA-5 reanalysis dataset~\citep{hersbach2020era5} from 1980-2018 sub-sampled with a 6-hour stride following prior work with snapshots at 0000, 0600, 1200, and 1800 UTC~\citep{pathak2022fourcastnet,bi2022pangu,lam2022graphcast}.
We use data from 1980-2016 for model training, 2017 for validation, and 2018 for out-of-sample test set.
All evaluations presented in the paper are based on the 2018 out-of-sample test set.
We used 73 input variables including 8 surface variables, and 5 variables (q, t, u, v, z) sampled at 13 pressure levels, making the final input size for each timestamp to be: $721 \times 1440 \times 73$.
We standardize each input channel based on the per-channel mean and standard deviation computed on the entire dataset. 

\subsection{Additional channels}

Zenith angle is appended as an extra channel to the input when enabled.
Therefore, during auto-regressive rollouts, we append a new zenith angle for each timestamp.
The input location is similarly appended as a separate input channel.
We evaluate the use of three constant masks, including 
land mask, soil type, and topography following \citet{bi2022pangu}.
For multi-step inputs, multiple steps are concatenated in the channel dimension. Therefore, the number of input channels is multiplied by the number of input steps. Input location and constant masks are only appended once when using multi-step inputs as they are static channels.
These additional channels only affect the number of input channels, while leaving the number of output channels intact as these channels are not predicted by the model.

\subsection{Multi-step fine-tuning}

We backpropagate through multiple auto-regressive steps when fine-tuning.
By default, we supervise only the last step in the auto-regressive rollout.
We also evaluate intermediate-step supervision, where for each intermediate step, we calculate a (discounted) loss with a discount factor $\gamma$ of 0.9 starting from 1.0 for the first step i.e., the weight for step $i$ is computed as $w_i = \gamma^{i-1}$ for $i \ge 1$.
Furthermore, when using intermediate step supervision, we also experiment with the use of scheduled sampling~\citep{bengio2015scheduledsampling} i.e., the input steps are a convex combination of the previous model prediction as well as the actual target in the dataset with a linear schedule starting at 0.9 for the target, and annealing it to 0 for the last epoch i.e., the weight for the prediction at different epochs is 0.1, 0.2, 0.3, .., 1.0 for the 10 epochs.

\subsection{Optimization}

All models were trained via synchronous updates for 10 epochs on 128 H100 GPUs using a batch size of 1 per GPU (effective batch size of 128).
We used AdamW for optimization with an initial learning rate of 0.001, cosine learning rate decay, and weight decay of 0.0001.
Note that we did not tune the hyperparameters for each of the different architectures separately.

\subsection{Loss functions}

We evaluated the use of different loss functions for training including $L_2^2$ i.e., Mean-Squared Error (MSE), $L_1$, along with their linear composition, and their combinations such as Huber loss~\citep{huber1992robust}. For the ground truth next time-step weather state $\mathbf{y}$ and the model forecast $\mathbf{\hat{y}}$, $L_2^2$ and $L_1$ losses are defined as:

\begin{equation}
    \mathcal{L}_2^2(\mathbf{\hat{y}}, \mathbf{y}) = \frac{1}{H W C} \sum_{h=1}^{H} \sum_{w=1}^{W} \sum_{c=1}^{C} \; (\mathbf{\hat{y}}_{h,w,c}- \; \mathbf{y}_{h,w,c})^2
    \label{eq:l2_loss}
\end{equation}
\begin{equation}
    \mathcal{L}_{1}(\mathbf{\hat{y}}, \mathbf{y}) = \frac{1}{H W C} \sum_{h=1}^{H} \sum_{w=1}^{W} \sum_{c=1}^{C} \; |\mathbf{\hat{y}}_{h,w,c}- \; \mathbf{y}_{h,w,c}|
    \label{eq:l1_loss}
\end{equation}

\noindent Huber loss is defined as:

\begin{align}
    \mathcal{L}_H(\mathbf{\hat{y}}, \mathbf{y}) = \frac{1}{H W C} \sum_{h=1}^{H} \sum_{w=1}^{W} \sum_{c=1}^{C}
     \begin{cases}
       \frac{1}{2}(\mathbf{\hat{y}}_{h,w,c}- \; \mathbf{y}_{h,w,c})^2&\quad\text{if}\quad|\mathbf{\hat{y}}_{h,w,c}- \; \mathbf{y}_{h,w,c}|\leq \delta \\ 
       \delta(|\mathbf{\hat{y}}_{h,w,c}- \; \mathbf{y}_{h,w,c}|-\frac{\delta}{2})&\quad\quad\quad\text{otherwise}
     \end{cases}
\end{align}

Following the function view of the problem, i.e., the variables are in fact functions defined everywhere on the sphere for which we only have access to their pointwise evaluation on a grid, we also explored training using geometric $L_2^2$ and $L_1$ losses. For a given quadrature rule $w$:

\begin{equation}
    w(i) = \frac{\text{cos(lat}(i))}{\frac{1}{H} \sum_{j=1}^{H} \text{cos(lat}(j))}
    \label{eq:latitude_weights}
\end{equation}
\begin{equation}
    \mathcal{L}(\mathbf{\hat{y}}, \mathbf{y}) = \frac{1}{H W C} \sum_{h=1}^{H} w(h) \sum_{w=1}^{W} \sum_{c=1}^{C} \; \mathcal{L}'(\mathbf{\hat{y}}_{h,w,c}, \; \mathbf{y}_{h,w,c})
    \label{eq:latitude_weighted_loss}
\end{equation}

\noindent where inner $\mathcal{L}'$ is either $\frac{1}{2}(\cdot)^2$ for the $L_2^2$ loss on function spaces, and $|\cdot|$ for $L_1$ on function spaces.
These quadrature rules have been frequently used in the past~\citep{pathak2022fourcastnet,bi2022pangu,lam2022graphcast,nguyen2023stormer}.

\subsection{Metrics}

We report geometric (latitude-weighted) Anomaly Correlation Coefficient (ACC), and Root Mean-Squared Error (RMSE), where the latitudes are weighted according to the latitude value~\citep{rasp2020weatherbench,nguyen2023stormer} (see Eq.~\ref{eq:latitude_weighted_loss}). ACC is defined as:

\begin{equation}
    \mathbf{ACC}(\mathbf{\hat{y}}, \mathbf{y}) = \frac{1}{C} \sum_{c=1}^{C} \frac{\frac{1}{H W} \sum_{h=1}^{H} w(h) \sum_{w=1}^{W} (\mathbf{\hat{y}}_{h,w,c} \times \mathbf{y}_{h,w,c})}{\sqrt{\frac{1}{H W} \sum_{h=1}^{H} w(h) \sum_{w=1}^{W} \mathbf{\hat{y}}_{h,w,c}^{2}} \sqrt{\frac{1}{H W} \sum_{h=1}^{H} w(h) \sum_{w=1}^{W} \mathbf{y}_{h,w,c}^{2}}}
    \label{eq:latitude_weighted_acc}
\end{equation}

\noindent These two metrics allow for assessing the quality of the trained model. In particular, ACC considers the climatology-corrected correlation between the ground truth and model forecast, and RMSE considers the quantitative differences between the two fields.  
Since each channel is normalized, we report the mean RMSE and ACC across all 73 input channels.

%% file: sections/experiments.tex
\section{Experiments}

\input{sections/num_params_table}

\begin{figure}[t]
    \centering
    \includegraphics[width=0.37\textwidth,trim={0cm 0 18.8cm 0},clip]{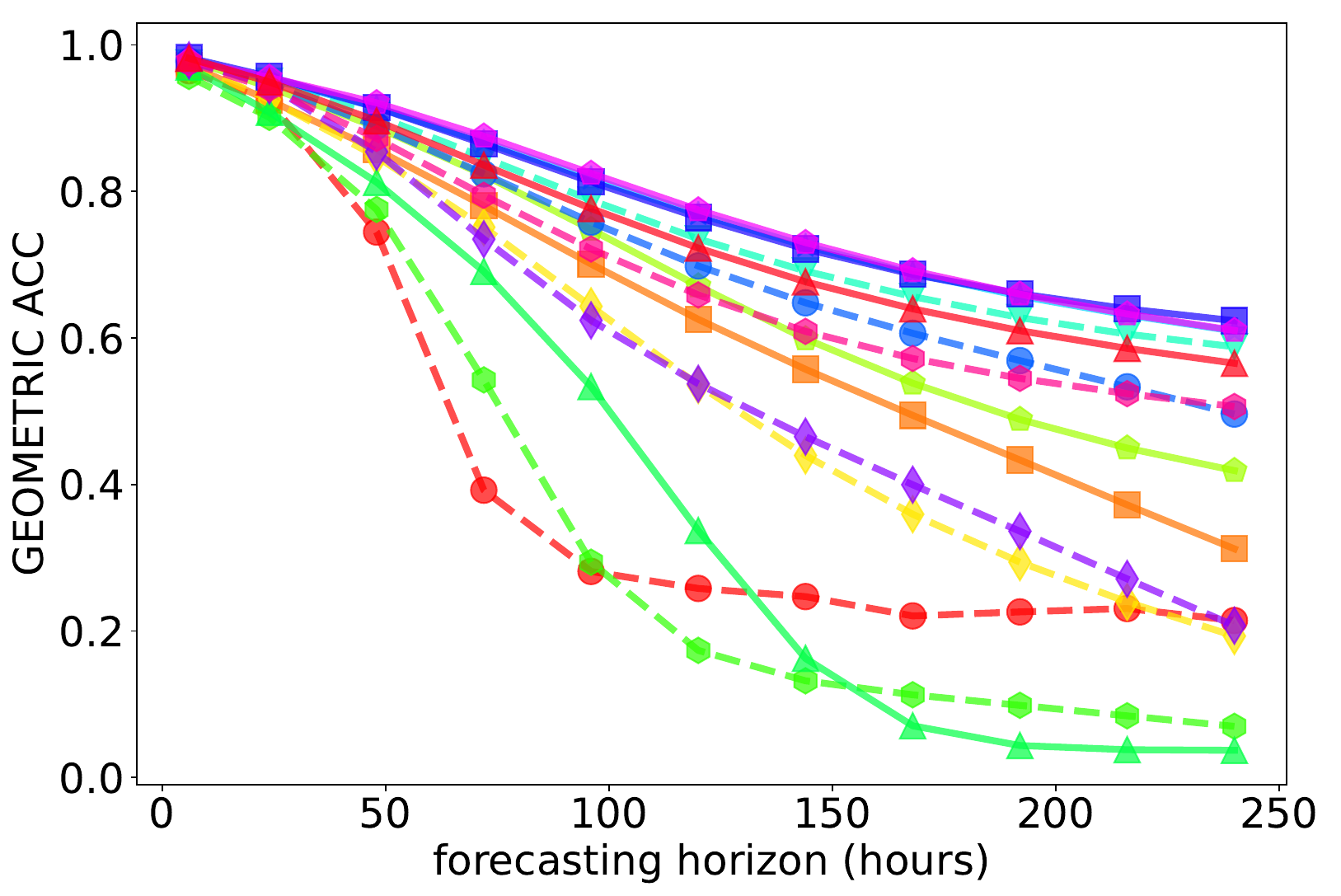}
    \includegraphics[width=0.62\textwidth]{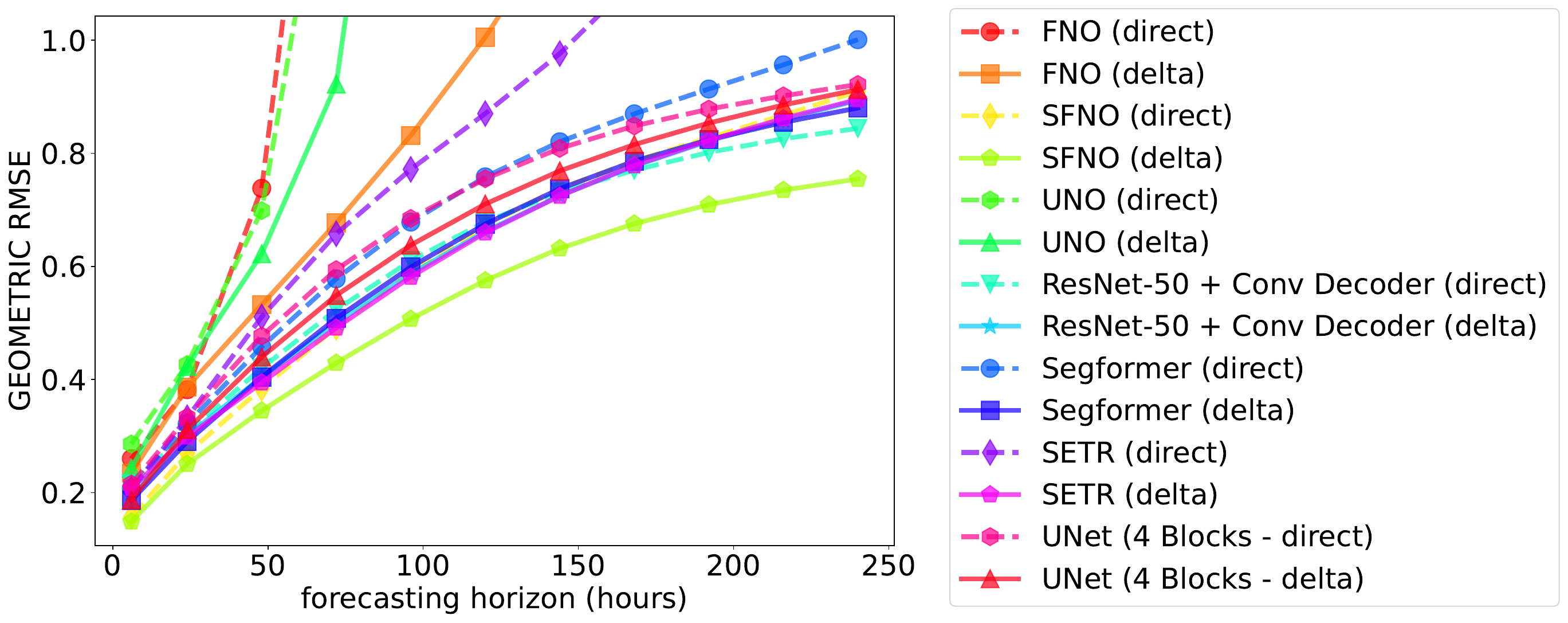}
    \caption{\textbf{Geometric ACC (left) and RMSE (right) for comparison between direct and delta prediction formulations.} The figure highlights that delta prediction is almost always superior in terms of performance as compared to direct prediction at the 6h prediction horizon (results in tabular form are presented in Table~\ref{direct_vs_delta_pred_ood_num_steps_test_loss.mean}).}
    \label{fig:direct_vs_delta_pred}
\end{figure}

\input{results_v3/direct_vs_delta_pred/direct_vs_delta_pred_ood_num_steps_test_loss.mean}

\begin{figure}[t]
    \centering
    \includegraphics[width=0.8\textwidth]{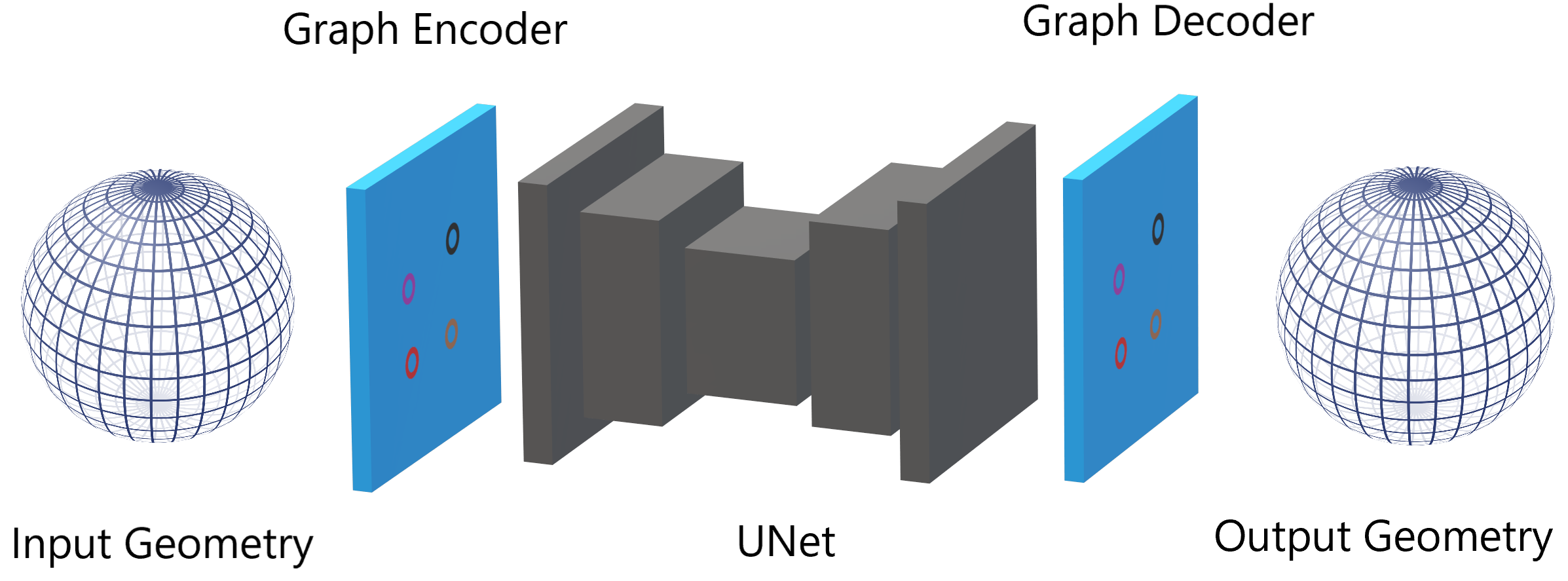}
    \caption{\textbf{Graph UNet architecture} where a fixed-grid UNet architecture is sandwiched between grid-invariant graph encoder and decoder layers. This provides the model with the flexibility of being grid-invariant while retaining the performance of fixed-grid models.}
    \label{fig:graph_unet}
\end{figure}

\begin{figure}[t]
    \centering
    \includegraphics[width=0.385\textwidth,trim={0cm 0 15.5cm 0},clip]{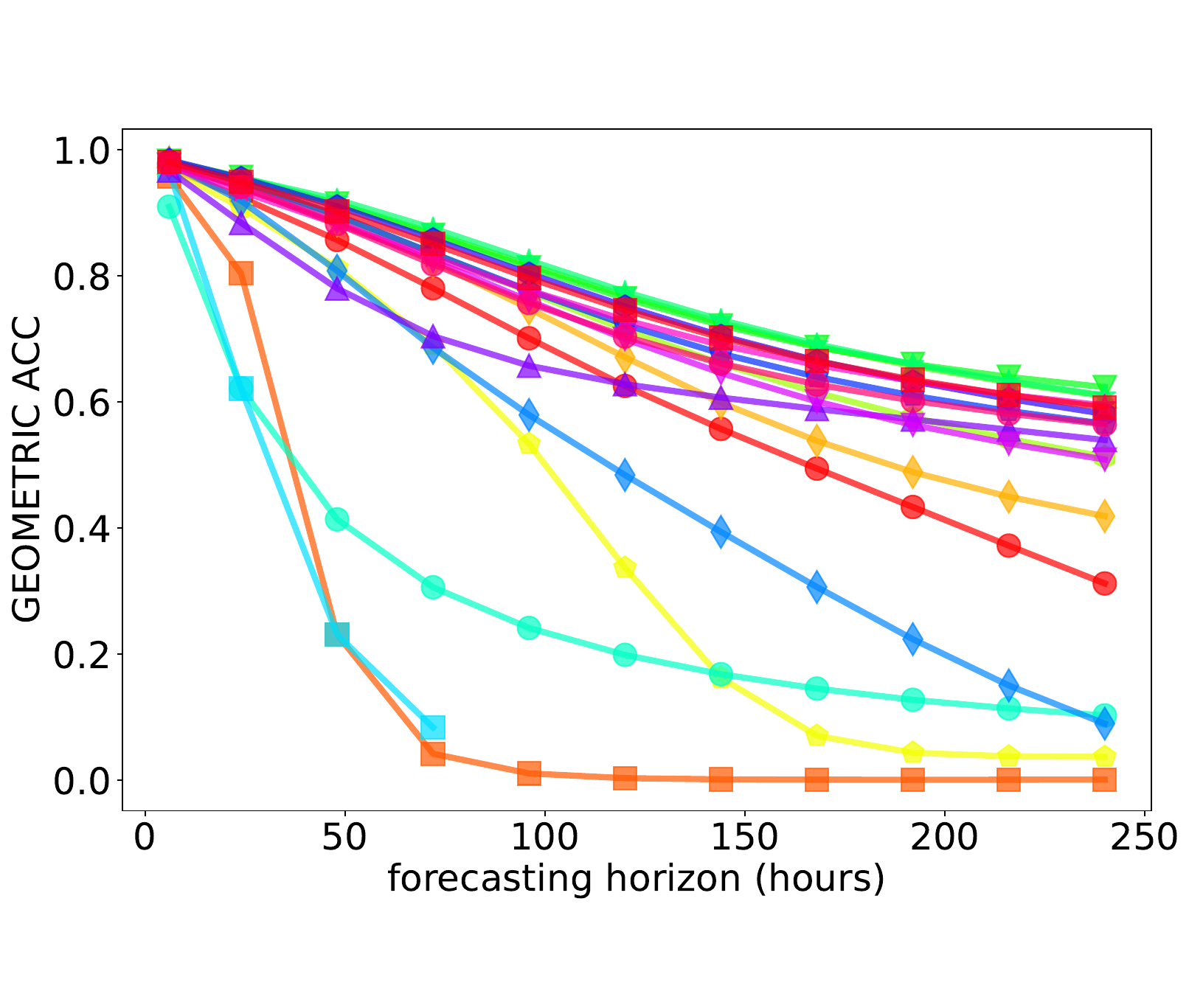}
    \includegraphics[width=0.605\textwidth]{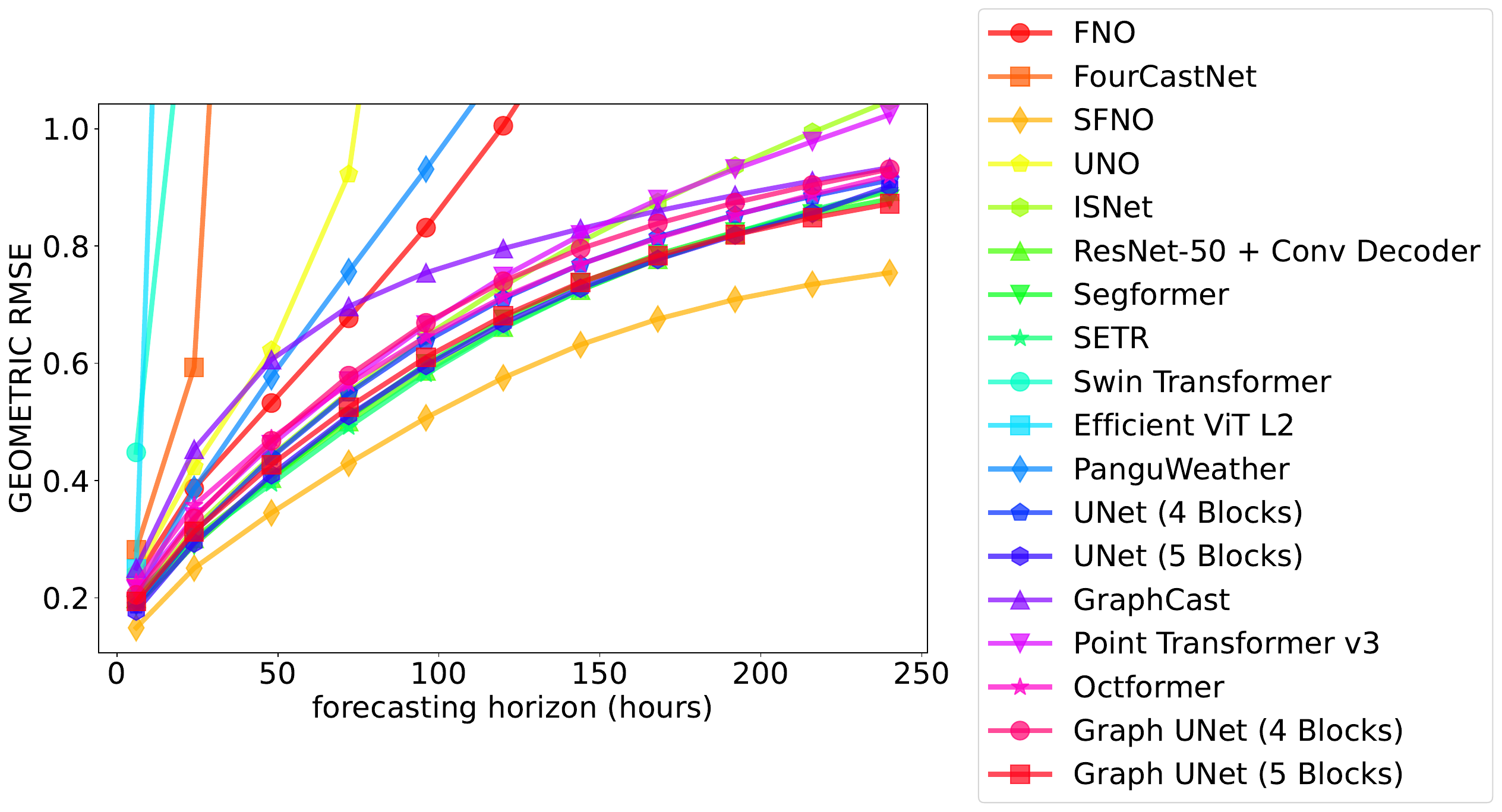}
    \caption{\textbf{Geometric ACC (left) and RMSE (right) for comparison of different architectures when using the delta prediction formulation.} The figure highlights that some architectures such as Segformer, SETR, ResNet-50 w/ convolutional decoder, and UNet are more efficient and effective for weather forecasting given a small and fixed number of training steps that we focused on in this work (results in tabular form are presented in Table~\ref{delta_pred_ood_num_steps_test_loss.mean}).}
    \label{fig:delta_pred}
\end{figure}

\input{results_v3/delta_pred/delta_pred_ood_num_steps_test_loss.mean}

\begin{figure}[t]
    \centering
    \includegraphics[width=0.6\textwidth,trim={0cm 18cm 0cm 0cm},clip]{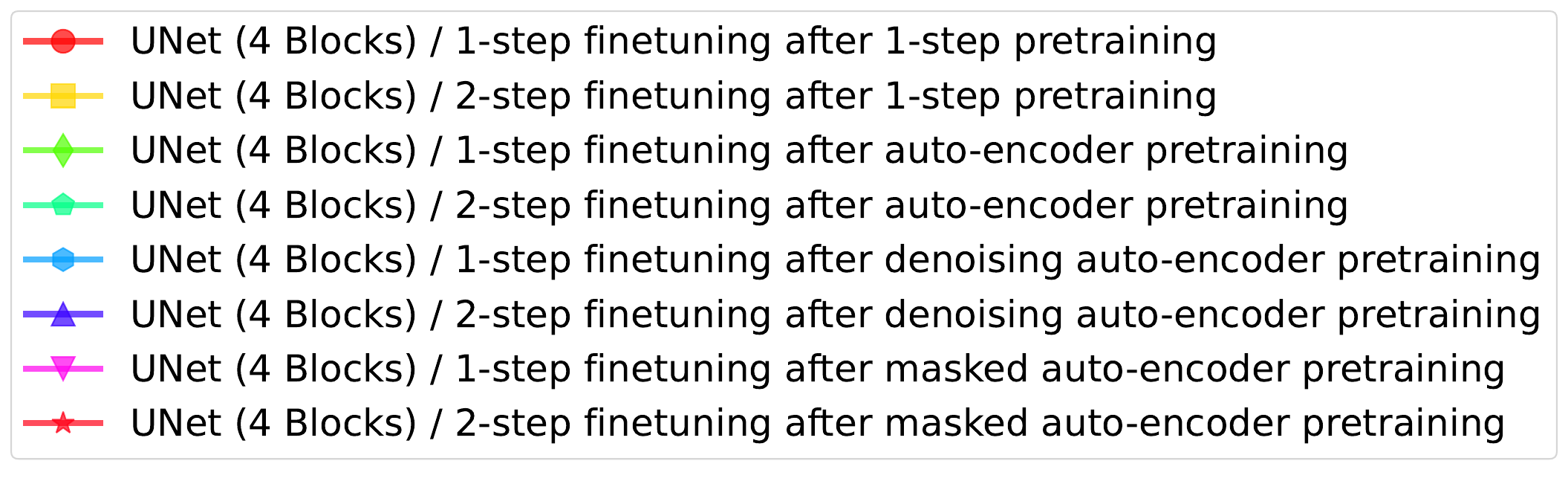}

    \includegraphics[width=0.495\textwidth,trim={3cm 0cm 5cm 11cm},clip]{results_v2/pretraining_4layers/pretraining_4layers_ood_num_steps_test_loss.geometric_acc.mean.pdf}
    \includegraphics[width=0.495\textwidth,trim={3cm 0cm 5cm 11cm},clip]{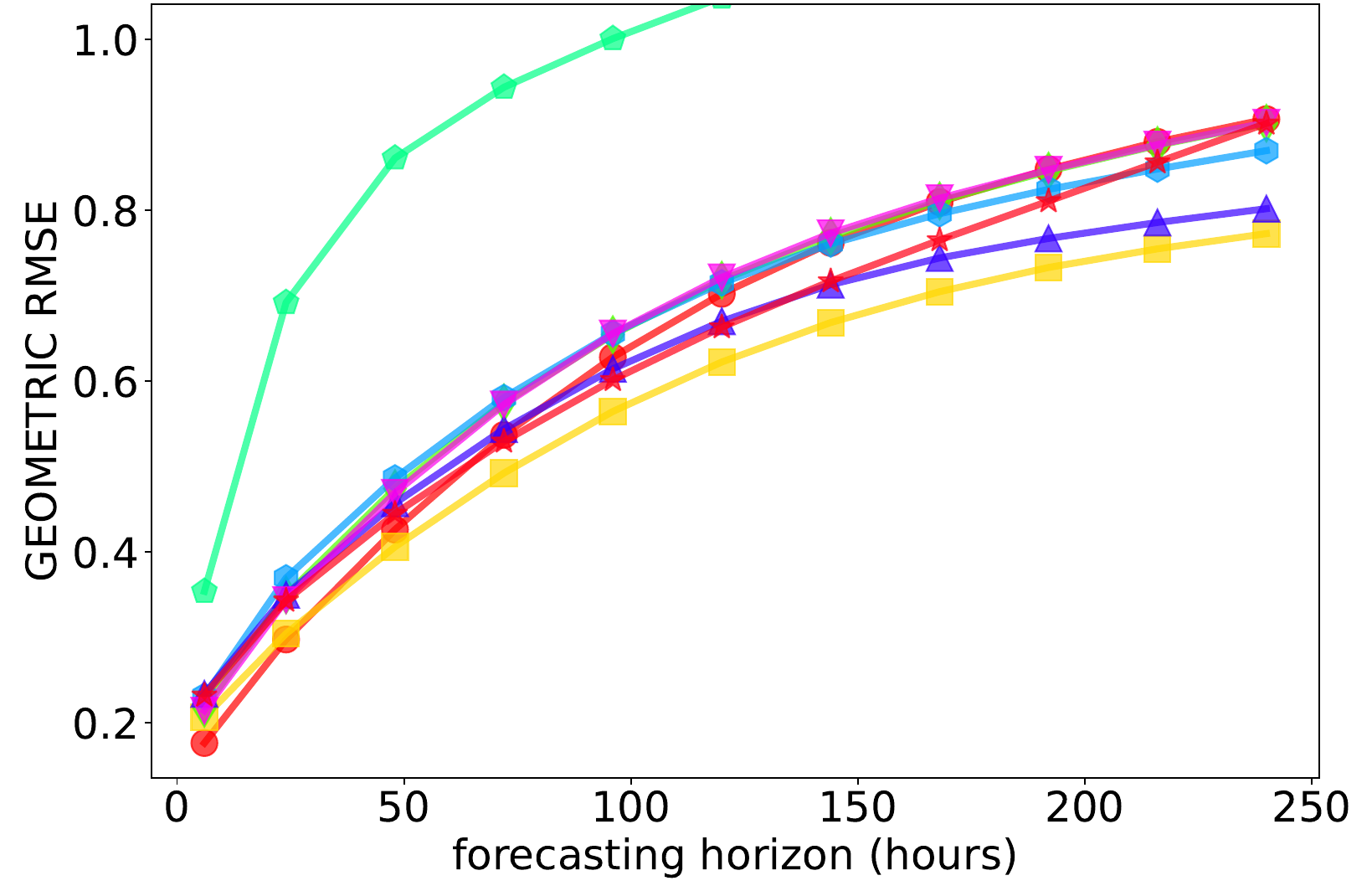}

    \caption{\textbf{Geometric ACC (left) and RMSE (right) for comparison of different pretraining objectives on UNet with 4 blocks.} The figure highlights that supervised pretraining achieves better performance in comparison to self-supervised pretraining using different objectives evaluated in our case (results in tabular form are presented in Table~\ref{pretraining_4layers_ood_num_steps_test_loss.mean}).}
    \label{fig:pretraining_4layers}
\end{figure}

\input{results_v2/pretraining_4layers/pretraining_4layers_ood_num_steps_test_loss.mean}

\begin{figure}[t]
    \centering
    \includegraphics[width=0.6\textwidth,trim={0cm 18.25cm 0cm 0cm},clip]{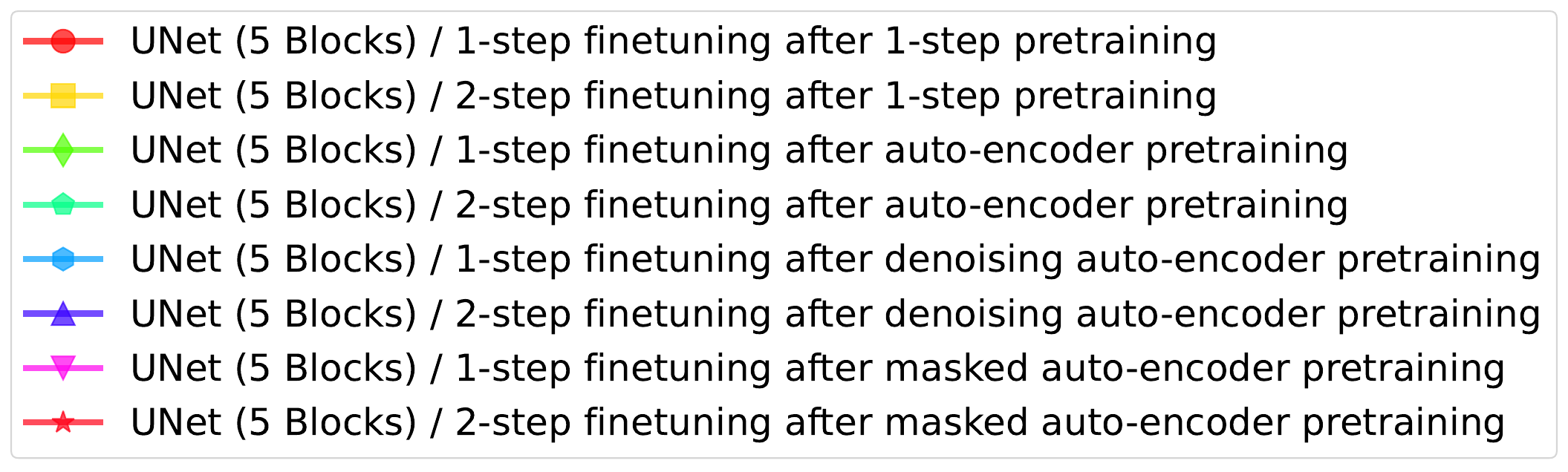}

    \includegraphics[width=0.495\textwidth,trim={3cm 0cm 5cm 11cm},clip]{results_v2/pretraining_5layers/pretraining_5layers_ood_num_steps_test_loss.geometric_acc.mean.pdf}
    \includegraphics[width=0.495\textwidth,trim={3cm 0cm 5cm 11cm},clip]{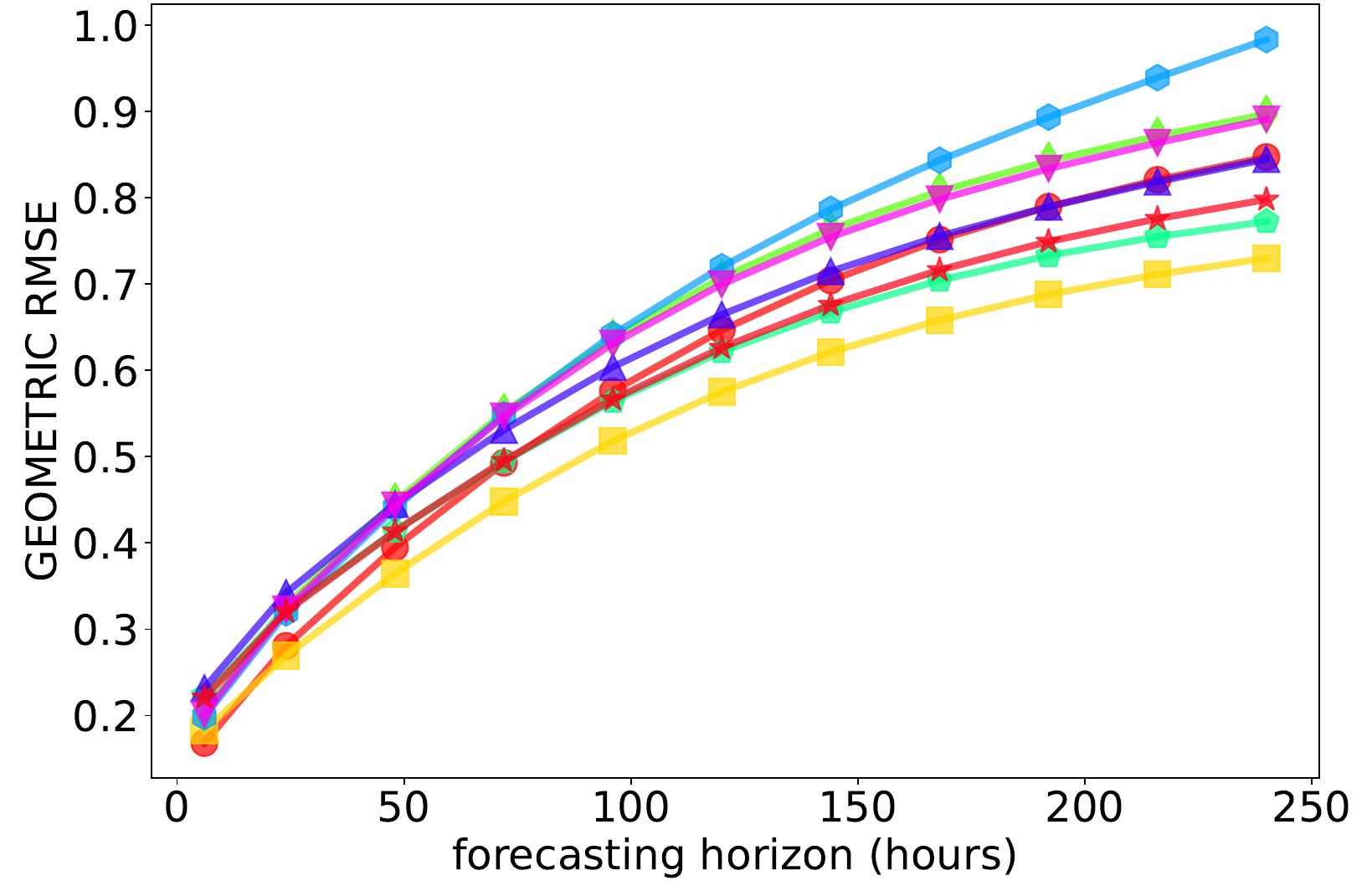}

    \caption{\textbf{Geometric ACC (left) and RMSE (right) for comparison of different pretraining objectives on UNet with 5 blocks.} The figure highlights that supervised pretraining achieves better performance in comparison to self-supervised pretraining using different objectives evaluated in this work (results in tabular form are presented in Table~\ref{pretraining_5layers_ood_num_steps_test_loss.mean}).}
    \label{fig:pretraining_5layers}
\end{figure}

\input{results_v2/pretraining_5layers/pretraining_5layers_ood_num_steps_test_loss.mean}

\begin{figure}[t]
    \centering
    \includegraphics[width=0.6\textwidth,trim={0cm 18.5cm 0cm 0cm},clip]{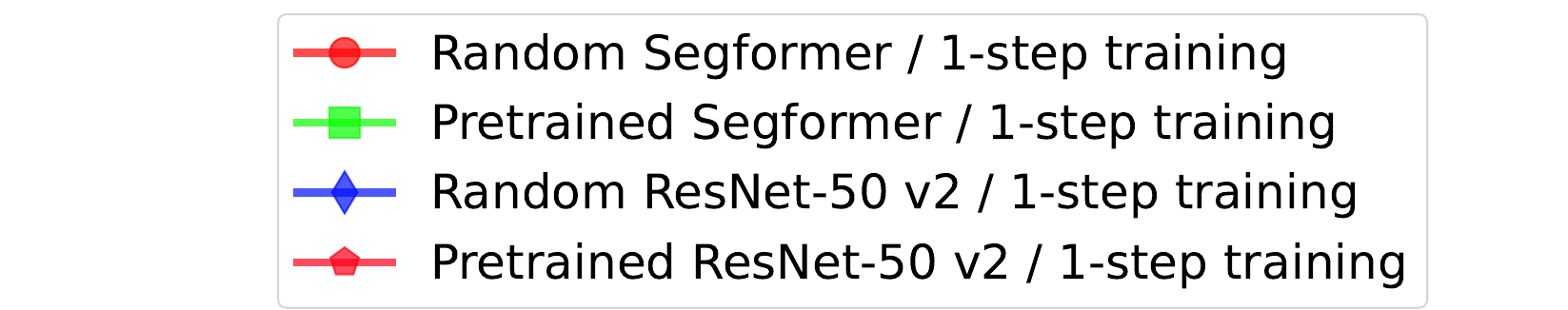}

    \includegraphics[width=0.495\textwidth,trim={0cm 0cm 0cm 5.8cm},clip]{results_v2/img_pretraining/img_pretraining_ood_num_steps_test_loss.geometric_acc.mean.pdf}
    \includegraphics[width=0.485\textwidth,trim={0cm 0cm 0cm 5.8cm},clip]{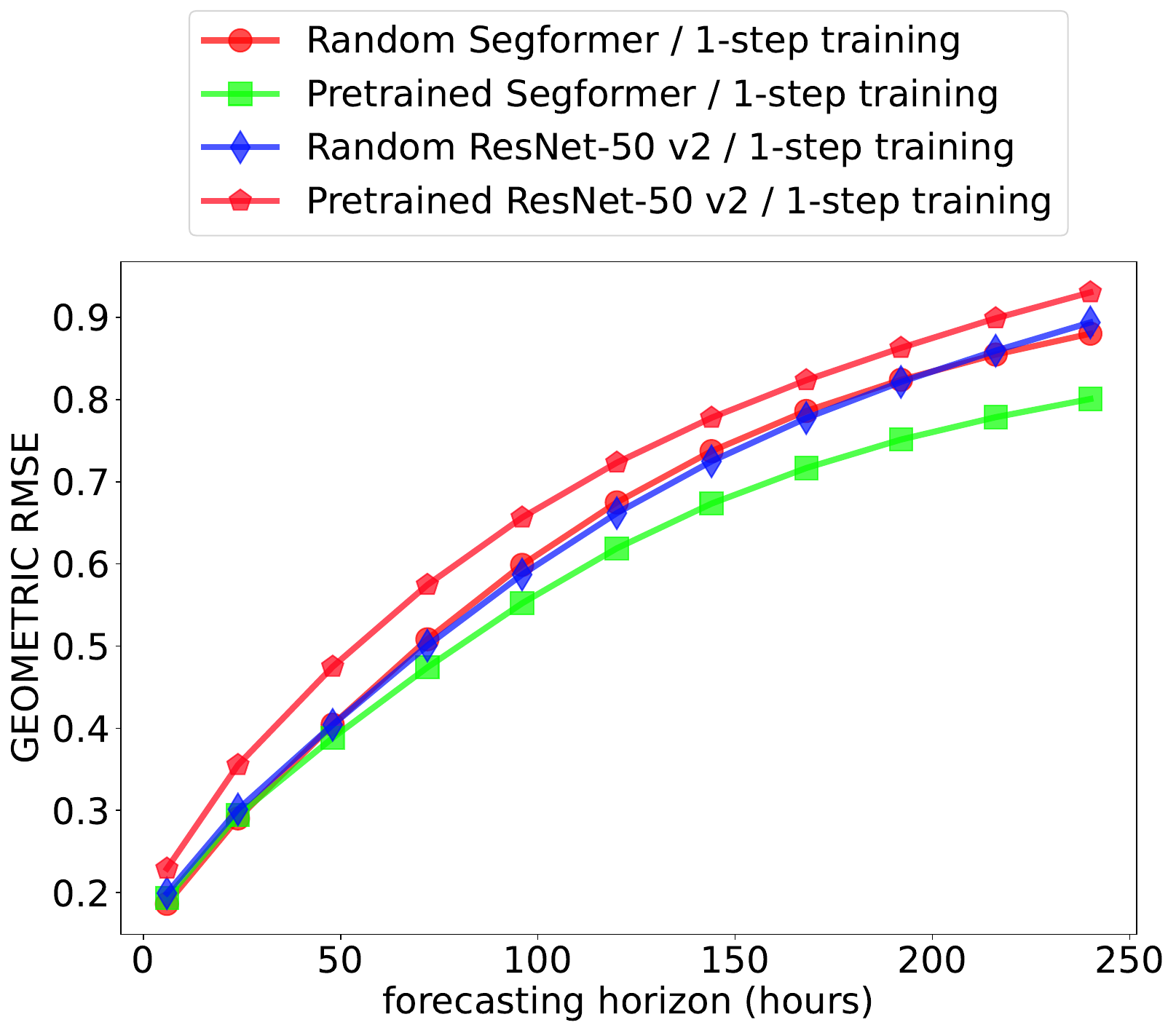}

    \caption{\textbf{Geometric ACC (left) and RMSE (right) when using image-based pretrained models.} The figure highlights that there are non-trivial gains for Segformer when using a pretrained model which was directly trained for segmentation as compared to ResNet-50 which was trained for classification (results in tabular form are presented in Table~\ref{img_pretraining_ood_num_steps_test_loss.mean}).}
    \label{fig:img_pretraining}
\end{figure}

\input{results_v2/img_pretraining/img_pretraining_ood_num_steps_test_loss.mean}

\begin{figure}[t]
    \centering

    \includegraphics[width=0.375\textwidth,trim={0cm 0cm 17.5cm 0cm},clip]{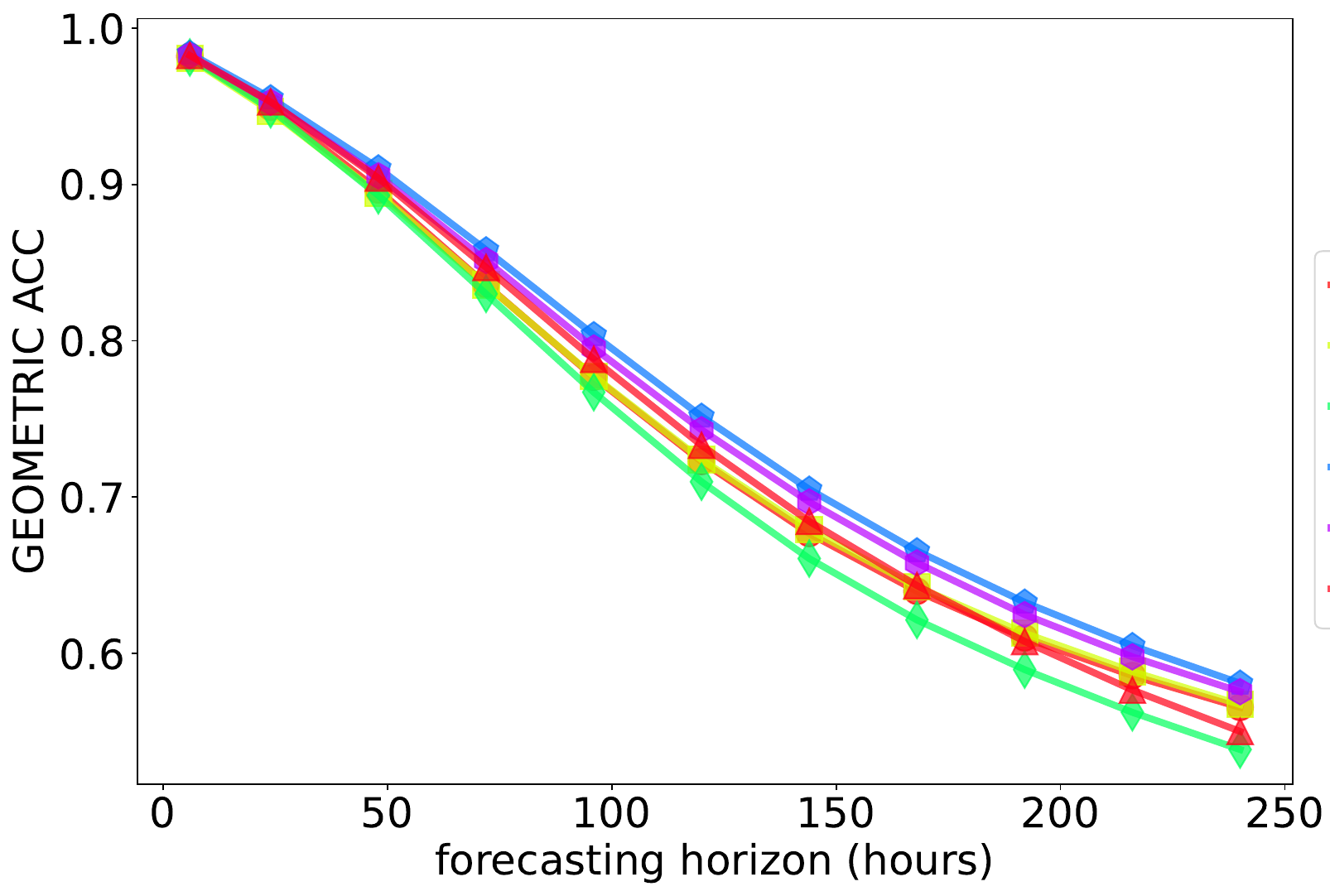}
    \includegraphics[width=0.615\textwidth,trim={0cm 0cm 0cm 0cm},clip]{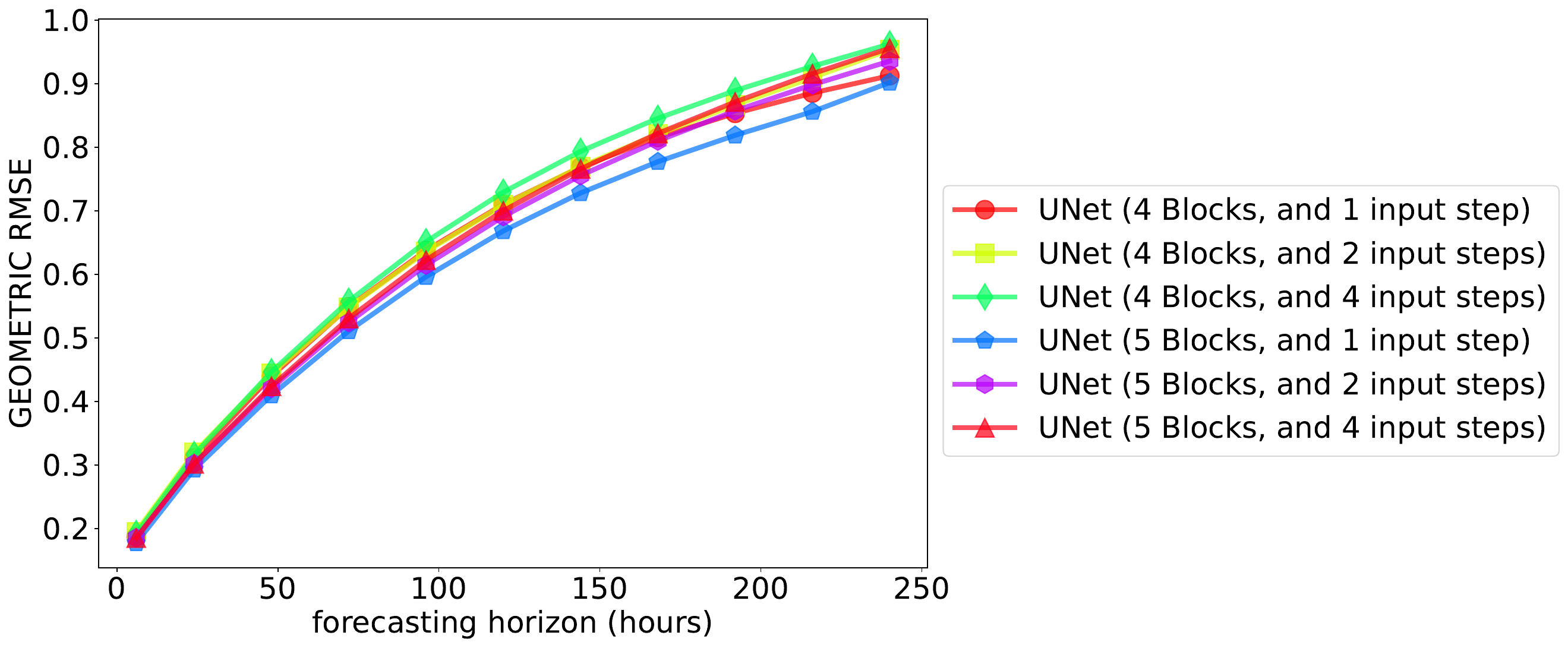}

    \caption{\textbf{Geometric ACC (left) and RMSE (right) with multi-step inputs.} Given the fixed number of training steps, we observe a negligible difference between 1 and 2 input steps starting with the next step prediction. However, performance significantly degrades when using 4 input steps despite achieving a similar starting performance to the 2 input step model for the next step i.e., 6h prediction (results in tabular form are presented in Table~\ref{multistep_inputs_ood_num_steps_test_loss.mean}).}
    \label{fig:multi_step_inputs}
\end{figure}

\input{results_v2/multistep_inputs/multistep_inputs_ood_num_steps_test_loss.mean}

\begin{figure}[t]
    \centering

    \includegraphics[width=0.6\textwidth,trim={0cm 18.5cm 0cm 0cm},clip]{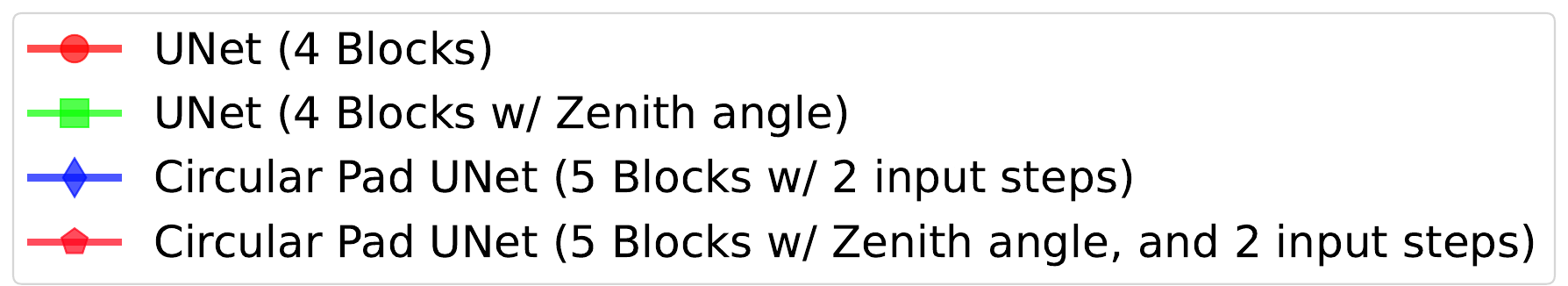}
    \vspace{2mm}

    \includegraphics[width=0.495\textwidth,trim={0.5cm 0cm 2cm 6cm},clip]{results_v2/zenith_angle/zenith_angle_ood_num_steps_test_loss.geometric_acc.mean.pdf}
    \includegraphics[width=0.495\textwidth,trim={0.5cm 0cm 2cm 6cm},clip]{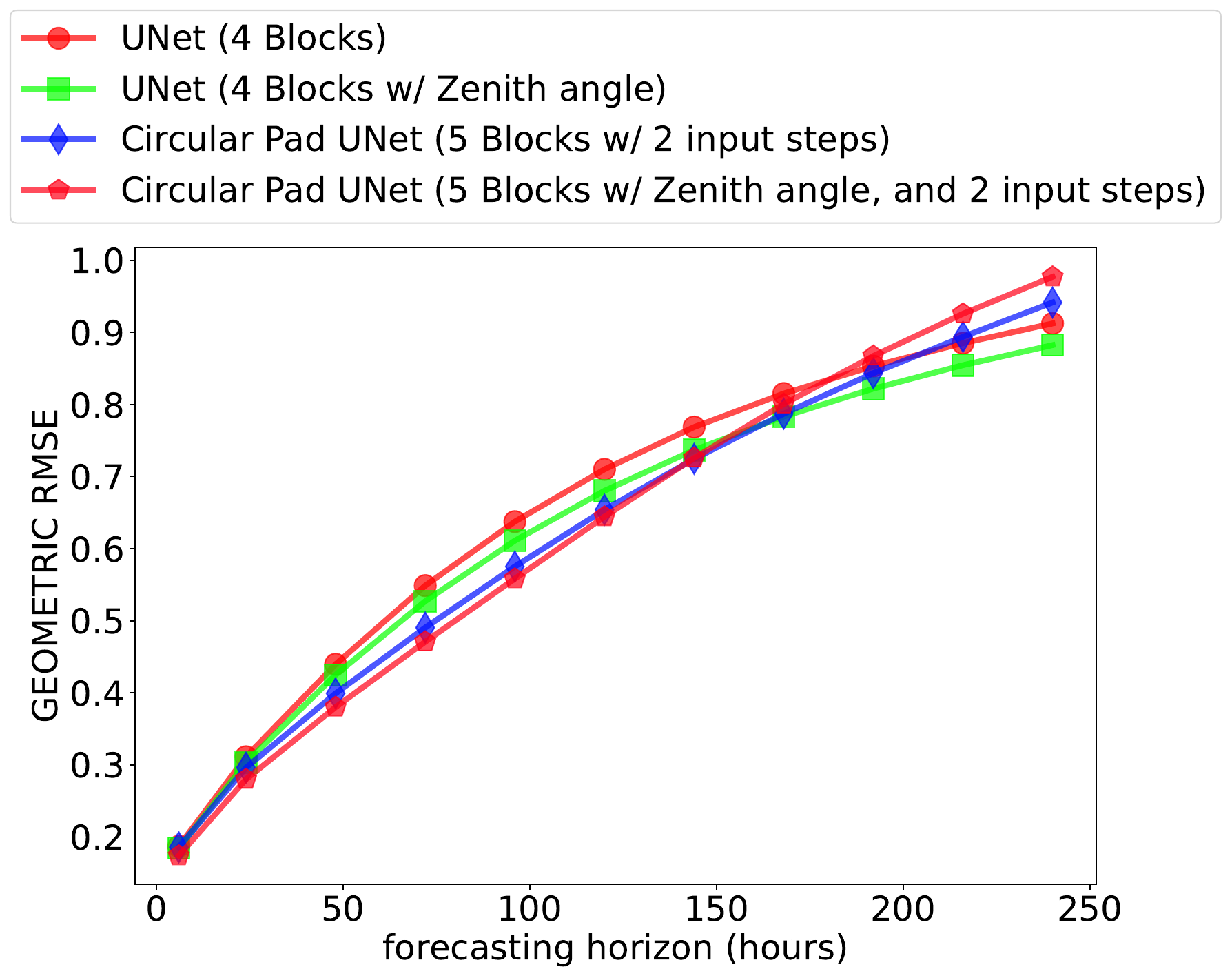}

    \caption{\textbf{Geometric ACC (left) and RMSE (right) with the inclusion of zenith angle.} The figure highlights that the Zenith angle provides consistent gains in predictive performance at both short-horizon and long-horizon forecasts (results in tabular form are presented in Table~\ref{zenith_angle_ood_num_steps_test_loss.mean}).}
    \label{fig:results_zenith}
\end{figure}

\input{results_v2/zenith_angle/zenith_angle_ood_num_steps_test_loss.mean}

\begin{figure}[t]
    \centering
    \includegraphics[width=0.7\textwidth,trim={0cm 18.25cm 0cm 0cm},clip]{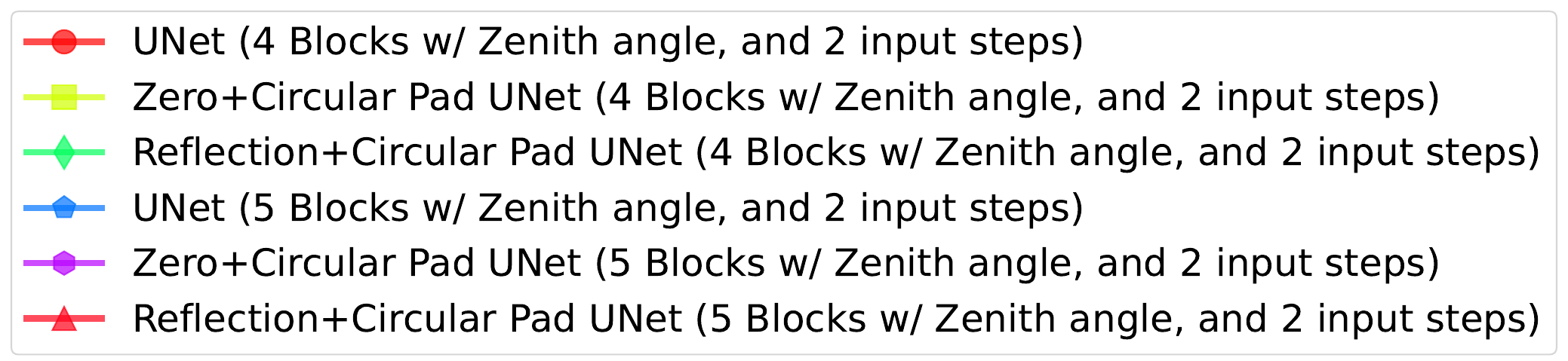}
    \vspace{1mm}

    \includegraphics[width=0.495\textwidth,trim={3cm 0cm 5cm 8.25cm},clip]{results_v2/padding_models/padding_models_ood_num_steps_test_loss.geometric_acc.mean.pdf}
    \includegraphics[width=0.495\textwidth,trim={3cm 0cm 5cm 8.25cm},clip]{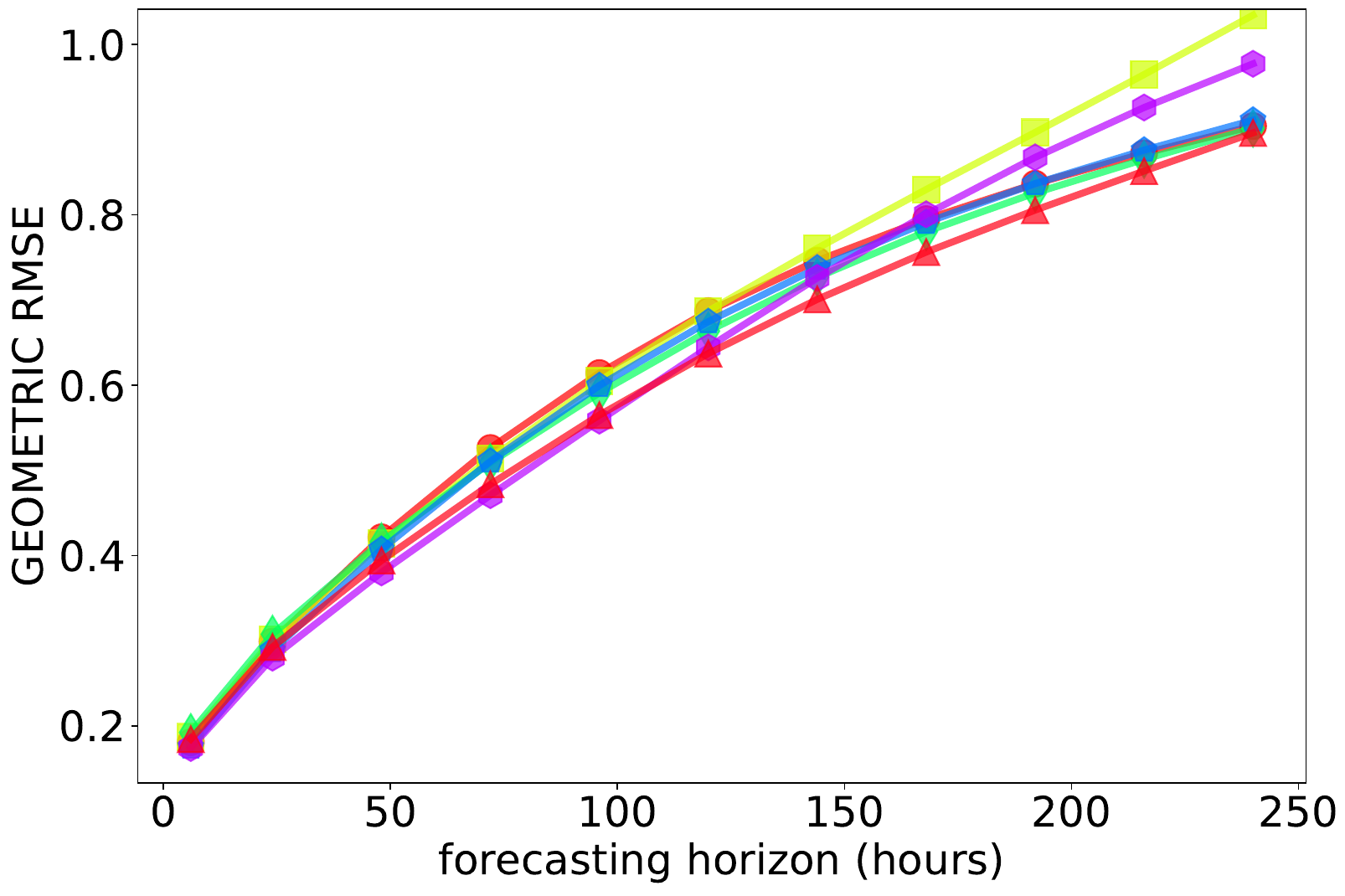}

    \caption{\textbf{Geometric ACC (a) and RMSE (b) with different padding schemes for the convolutional layers in the UNet architecture.} The figure highlights that using circular padding along the longitudinal direction and either zero or reflection padding along the latitudinal direction provides consistent gains in performance (results in tabular form are presented in Table~\ref{padding_models_ood_num_steps_test_loss.mean}).}
    \label{fig:results_padding_scheme}
\end{figure}

\input{results_v2/padding_models/padding_models_ood_num_steps_test_loss.mean}

\begin{figure}[t]
    \centering
    \includegraphics[width=0.7\textwidth,trim={0cm 18.5cm 0cm 0cm},clip]{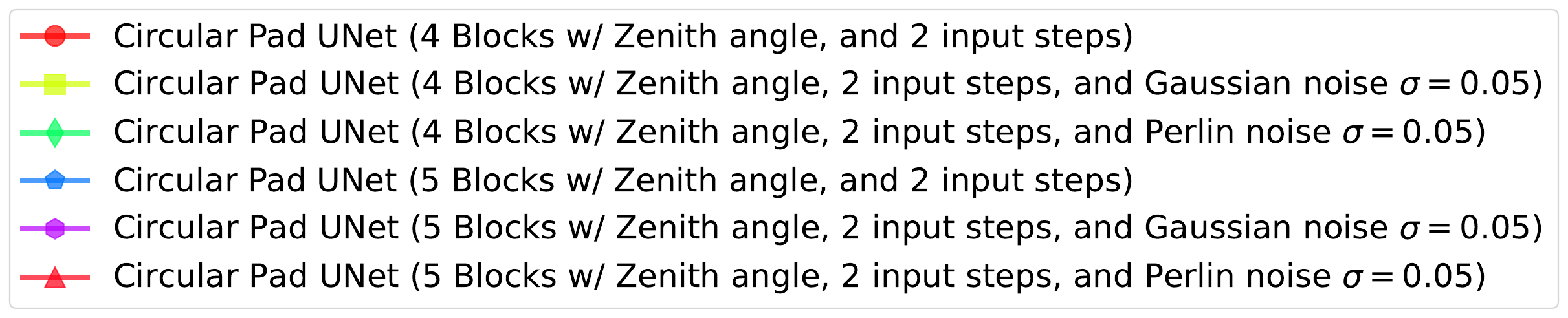}

    \includegraphics[width=0.495\textwidth,trim={6cm 0cm 8cm 8.25cm},clip]{results_v2/noise_models/noise_models_ood_num_steps_test_loss.geometric_acc.mean.pdf}
    \includegraphics[width=0.495\textwidth,trim={6cm 0cm 8cm 8.25cm},clip]{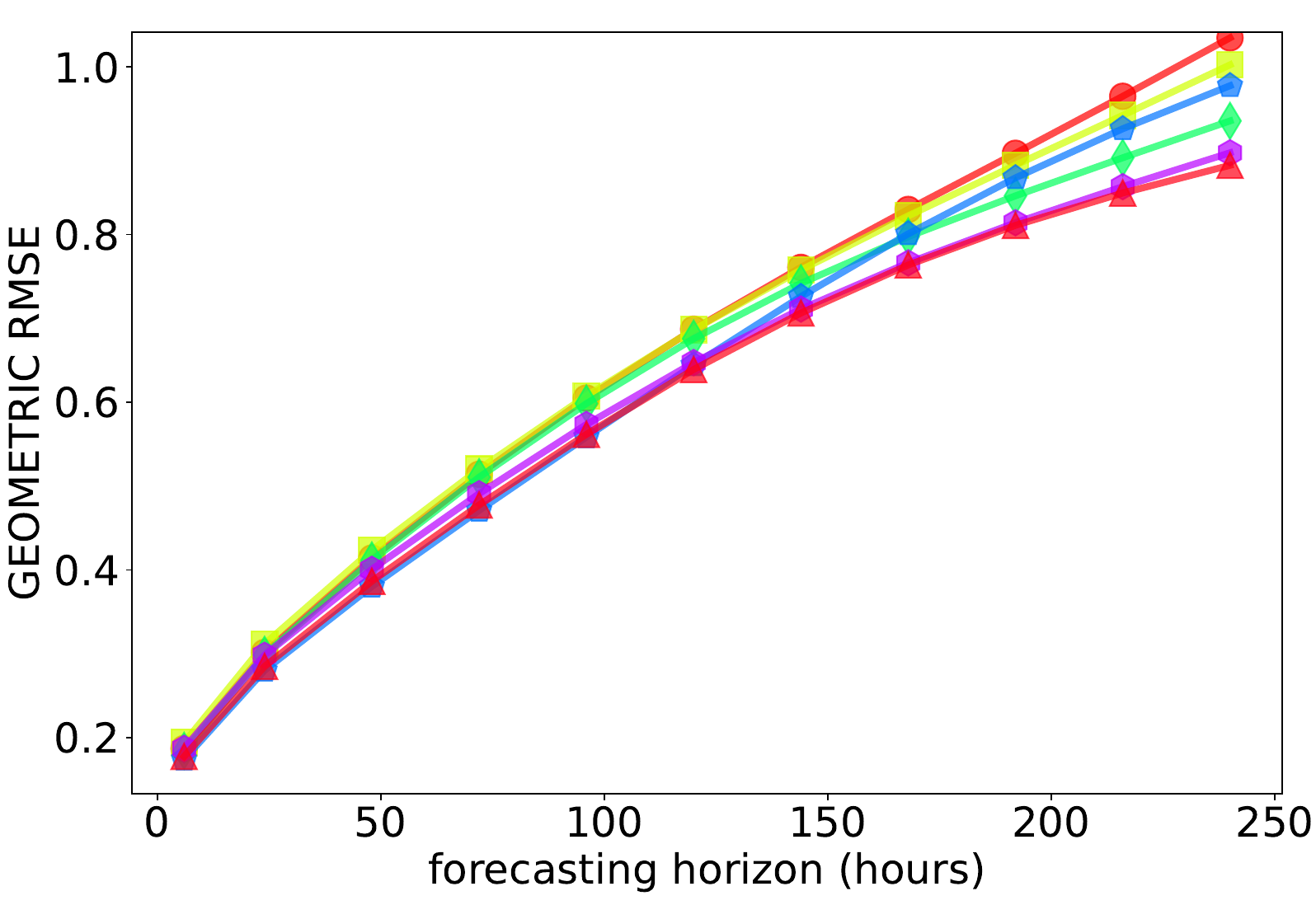}

    \caption{\textbf{Geometric ACC (left) and RMSE (right) with different noise addition schemes.} The figure highlights that adding some forms of noise (particularly Perlin noise as used in prior work~\citep{bi2022pangu,chen2023fuxi}) improves predictive performance for the long-horizon case with only minor detrimental effects on the short-horizon forecast (results in tabular form are presented in Table~\ref{noise_models_ood_num_steps_test_loss.mean}).}
    \label{fig:results_noise_add}
\end{figure}

\input{results_v2/noise_models/noise_models_ood_num_steps_test_loss.mean}

Our evaluation focuses on analyzing different aspects of the design space, starting from evaluating the impact of different problem formulations (Section~\ref{subsec:results_formulation}), model architecture (Section~\ref{subsec:results_arch}), self-supervised pretraining objectives (Section~\ref{subsec:results_pretraining_schemes}), image-based pretraining (Section~\ref{subsec:results_image_pretraining}), multi-step inputs (Section~\ref{subsec:results_multi_step_inputs}), zenith angle as an additional input channel (Section~\ref{subsec:results_zenith}), padding schemes for the convolutional layers (Section~\ref{subsec:results_padding_scheme}), noise addition to the input states (Section~\ref{subsec:results_noise_addition}), auxiliary information regarding the 3D coordinates on the sphere (Section~\ref{subsec:results_input_loc}), loss functions (Section~\ref{subsec:results_loss_fn}), additional auxiliary static channels such as soil type, land-sea mask, and topography (Section~\ref{subsec:results_constant_masks}), wider models by increasing the hidden dimension (Section~\ref{subsec:results_wider_models}), multi-step fine-tuning including models operating at a larger prediction horizon (Section~\ref{subsec:results_multistep_ft}), comparison of different models with multi-step fine-tuning (Section~\ref{subsec:multistep_ft_compare_models}), and training on a larger hourly dataset (Section~\ref{subsec:results_hourly_training}).

We always train the model using Mean-Squared Error (MSE) loss unless specified otherwise.
The number of parameters in the different models tested is presented in Table~\ref{tab:num_params} for reference.
Note that the number of parameters is not directly proportional to the memory footprint for that model.
A notable example is GraphCast~\citep{lam2022graphcast}, which has only a small number of parameters, but a very significant memory footprint.

As these sections are relatively independent, one can look at the sections independently without going through the others.

\subsection{Impact of formulation}
\label{subsec:results_formulation}

Direct prediction refers to the case where we directly generate the next weather state conditioned on the previous state(s). More concretely, $\mathbf{s_{t}} = \Phi(\mathbf{s_{t-1}})$ where $\Phi$ represents the deep-learning-based weather forecasting system. This has been a common choice for prior high-resolution models~\citep{pathak2022fourcastnet,bonev2023sfno,bi2022pangu,lam2022graphcast}.

Delta prediction refers to the case where we predict the change to be made to the previous state considering the weather forecasting to be discretization of a continuous-time process~\citep{chen2018neuralode}. More concretely, $\mathbf{s_{t}} = \mathbf{s_{t-1}} + \Phi(\mathbf{s_{t-1}})$ where $\Phi$ represents the deep-learning-based weather forecasting system. 
Note that this residual prediction is distinct from residual layers used within the model architecture.
Residual prediction for weather forecasting has also been explored in the recent past~\citep{nguyen2023stormer}.
However, we attempt to disentangle the precise impact of residual prediction instead of directly using it as our proposed approach.

The geometric ACC and RMSE for direct vs. delta prediction on selected architectures are presented in Table~\ref{direct_vs_delta_pred_ood_num_steps_test_loss.mean} (Fig.~\ref{fig:direct_vs_delta_pred}).
We see a clear trend in this case, with significantly better performance when using delta prediction compared to direct prediction. Fixed-grid models including UNet~\citep{ronneberger2015unet}, ResNet-50 with fully-convolutional decoder~\citep{long2015fcn}, SEgmentation TRansformer (SETR)~\citep{zheng2021setr}, and Segformer~\citep{xie2021segformer} perform better than other alternate architectures with only minor differences in performance between them.
One notable exception is the Spherical Fourier Neural Operator (SFNO)~\citep{bonev2023sfno} which achieves the lowest geometric RMSE, potentially due to correct modeling at the poles as it integrates geometric properties of the sphere\footnote{Note that we use the official SFNO repository to report SFNO results, therefore, while we attempted to minimize differences, the significantly lower RMSE might have been the result of differences in implementation. For this reason, we avoid making any claims about this result, and exclude it from discussion.}.

Using delta prediction essentially diminishes the performance difference between residual and non-residual architectures.
These results establish the utility of delta prediction compared to direct prediction (even for residual architectures) given the slow evolution of weather states at small prediction horizons (we used a prediction horizon of just 6 hours to train these models).
We expect diminishing improvements with delta prediction for larger prediction horizons due to significant changes in the weather state,
which we consider to be an interesting direction to investigate further in the future.

Given the utility of delta prediction in our case, all subsequent experiments use delta prediction as the formulation unless mentioned otherwise.
We will explore more architectures for this task in the next sub-section.

\subsection{Evaluation of different model architectures}
\label{subsec:results_arch}

Prior work on high-resolution weather forecasting explored different architecture, ranging from Adaptive Fourier Neural Operators~\citep{pathak2022fourcastnet}, Spherical Fourier Neural Operators~\citep{bonev2023sfno}, Vision Transformer~\citep{bi2022pangu,nguyen2023climax}, and message-passing GNNs~\citep{lam2022graphcast,price2023gencast}.
Note that the large input size used for high-resolution weather forecasting makes some architectures more applicable to this problem due to memory constraints.
Therefore, we further expand on the list of architectures explored in the last section.
Given the established utility of delta prediction, we only focus on delta prediction in this case.

In particular, we consider three different model families. The first type of models considered are regular image-to-image models prevalent in the computer-vision literature, including variants of convolutional networks~\citep{lecun1989convnet,ronneberger2015unet} such as UNet~\citep{ronneberger2015unet} and ResNet-50 with convolutional decoder~\citep{long2015fcn}, and variants of transformer architecture~\citep{vaswani2017attention,dosovitskiy2020vit} such as SEgementation TRansformer (SETR)~\citep{zheng2021setr}, Swin transformer~\citep{liu2021swin}, and Segformer~\citep{xie2021segformer}.

The second type that we consider are architectures that directly operate on point clouds, such as graph neural networks~\citep{gilmer2017mpgnn}, or point-transformers~\citep{wu2023pointtranformerv3}.
In particular, we evaluate many variants of these architectures including point transformer v3~\citep{wu2023pointtranformerv3}, Octformer~\citep{wang2023octformer}, and message-passing GNNs~\citep{gilmer2017mpgnn} in the form of GraphCast~\citep{lam2022graphcast}.
To fit the GraphCast model fully within a single GPU, we used 256 hidden dimensions (the original implementation used 512 as the hidden dimension) and mixed precision for training in contrast to all other models that were trained in full precision i.e., FP32.

The third type focuses on operator-based models, primarily designed to learn solution operators of differential equations that are resolution invariant by construction~\citep{li2020fno,ashiqur2022uno}.
FourCastNet~\citep{pathak2022fourcastnet} is based on the (Adaptive) Fourier Neural Operator (FNO)~\citep{li2020fno,guibas2021afno} model.
We further test two other operator models including Spherical Fourier Neural Operator (SFNO)~\citep{bonev2023sfno}, and U-shaped Neural Operator (UNO)~\citep{ashiqur2022uno}.

The geometric ACC and RMSE for residual prediction using a range of different architectures are presented in Table~\ref{delta_pred_ood_num_steps_test_loss.mean} (Fig.~\ref{fig:delta_pred}).
The results primarily highlight that fixed-grid models are superior in terms of performance as compared to graph-based models. In particular, we see the dominant performance of UNet~\citep{ronneberger2015unet}, Segformer~\citep{xie2021segformer}, SETR~\citep{zheng2021setr}, and ResNet-50 with a fully convolutional decoder~\citep{long2015fcn} at different stages.
Note that these results are based on just single-step training for 10 epochs.
Although one would prefer to train every model till convergence (or for a large number of epochs as commonly done in prior work~\citep{pathak2022fourcastnet,bi2022pangu}), this is computationally infeasible.
Therefore, our exploration which is compute-bound implicitly prioritizes more efficient learning architectures.

Due to the simplicity and prevalence of the UNet architecture, our future exploration focuses on UNet with residual prediction.
As UNet loses the ability to be grid-invariant~\citep{liu2024neural}, we also explore the coupling of UNet with a 2-layered graph neural operator layers as graph encoder and decoder (using $k=4$ for graph integration)~\citep{li2020neural} with the UNet model sandwiched in between, which we call `Graph UNet` (see Fig.~\ref{fig:graph_unet} for a visual illustration). Graph UNet retains the properties of being grid invariant via the graph encoder and decoder layers, while still retaining the performance of the UNet model as seen by the model's performance. Therefore, following the graph kernel operator learning~\citep{li2020neural} and 
adaptation of convolution neural networks to neural operators~\citep{liu2024neural}, our model qualities as a neural operator. 
This result shows that hybrid architectures that combine the flexibility of grid-invariant models with the performance of fixed-grid models can be a lucrative direction for designing future weather forecasting systems.
Based on these findings, we focus on the fixed-resolution models in this work, and train models without the graph encoder and decoder unless mentioned otherwise.

Note that since we do not use official implementation for all model classes, and use fixed hyperparameters, this may undermine the true performance of the model.
We explore the efficacy of the multi-step fine-tuning as well as several other techniques that we introduce in subsequent sections for ResNet-50 with fully-convolutional decoder~\citep{long2015fcn} and Segformer~\citep{xie2021segformer} towards the end in Section~\ref{subsec:multistep_ft_compare_models}.

\subsection{Impact of self-supervised pretraining}
\label{subsec:results_pretraining_schemes}

A wide range of different pretraining objectives has been used for pretraining deep-learning models in the past~\citep{he2022masked,brempong2022denoising,el2024scalable}.
\citet{man2023wmae} applied mask-autoencoding objective~\citep{he2022masked} for the pretraining of weather-forecasting systems.
However, their analysis confounded the impact of the pretraining objective, model architecture, and the training budget that our analysis attempts to disentangle.
Similarly, ClimaX~\citep{nguyen2023climax} trained using heterogeneous sources, evading the possibility of understanding the marginal contribution of the pretraining objective itself.
We explore the use of four simple pretraining techniques in this work:

\begin{itemize}
    \item Pretraining directly on the forecasting task. Note that fine-tuning using this task is equivalent to training the model for longer with a restart in the learning rate schedule.
    \item Auto-encoder pretraining where the model is trained to reconstruct the input. In this formulation, we remove the skip connections as it introduces a trivial solution, and initialize these skip connections randomly in the fine-tuning phase.
    \item Masked auto-encoder~\citep{he2022masked} where we mask complete channels of the input, and apply the loss to reconstruct only the masked channels, following the masked auto-encoder recipe~\citep{he2022masked}.
    \item Denoising auto-encoder~\citep{bengio2013denoising} where we add noise to the input, and ask the model to generate a denoised version of the input. In this case, since there is no trivial solution, we retain the skip connections in the model.
\end{itemize}

We always use the delta prediction task during fine-tuning, which may introduce a mismatch from the pretraining phase such as for the auto-encoder models.
However, we assume that pretraining might still provide a useful initialization point for the model.
We used a masking ratio of 50\% for masked auto-encoder training and a standard deviation of 0.1 for noise injection for the training of denoising auto-encoder.

\subsubsection{UNet (4 Blocks)}

The geometric ACC and RMSE for the different pretraining objectives on 4-block UNet are presented in Table~\ref{pretraining_4layers_ood_num_steps_test_loss.mean} (Fig.~\ref{fig:pretraining_4layers}).
The results highlight that pretraining using the supervised objective (pretraining and fine-tuning tasks are the same in this case) achieves better performance in comparison to the evaluated self-supervised objectives.
Despite not observing gains in performance with self-supervised objectives in our case, it is likely that improvements in the objectives used would result in significant improvement in forecasting capabilities in the future.
Furthermore, we only evaluate a single set of hyperparameters. It is possible that a slightly different combination of these hyperparameters can introduce significant differences in these results.
We leave the evaluation of other pretraining techniques such as generative pretraining~\citep{chen2020generativeimgpretraining}, as well as a more thorough search over hyperparameters as promising directions to explore in the future.

\subsubsection{UNet (5 Blocks)}

The geometric ACC and RMSE for the different pretraining objectives on 5-block UNet are presented in Table~\ref{pretraining_5layers_ood_num_steps_test_loss.mean} (Fig.~\ref{fig:pretraining_5layers}).
These results are consistent with the results on the 4-block UNet, where we observe supervised pretraining to outperform self-supervised objectives.

\begin{figure}[t]
    \centering
    \includegraphics[width=0.6\textwidth,trim={0cm 18.5cm 0cm 0cm},clip]{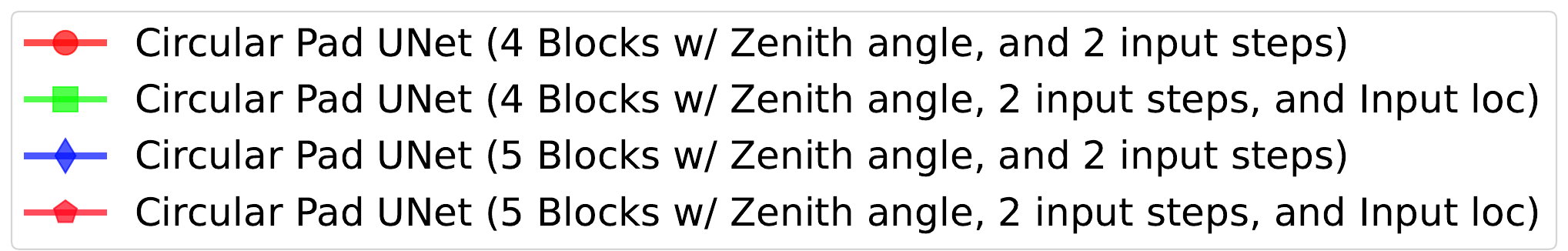}
    \vspace{1mm}

    \includegraphics[width=0.495\textwidth,trim={2.5cm 0cm 4.5cm 6cm},clip]{results_v2/input_loc/input_loc_ood_num_steps_test_loss.geometric_acc.mean.pdf}
    \includegraphics[width=0.495\textwidth,trim={2.5cm 0cm 4.5cm 6cm},clip]{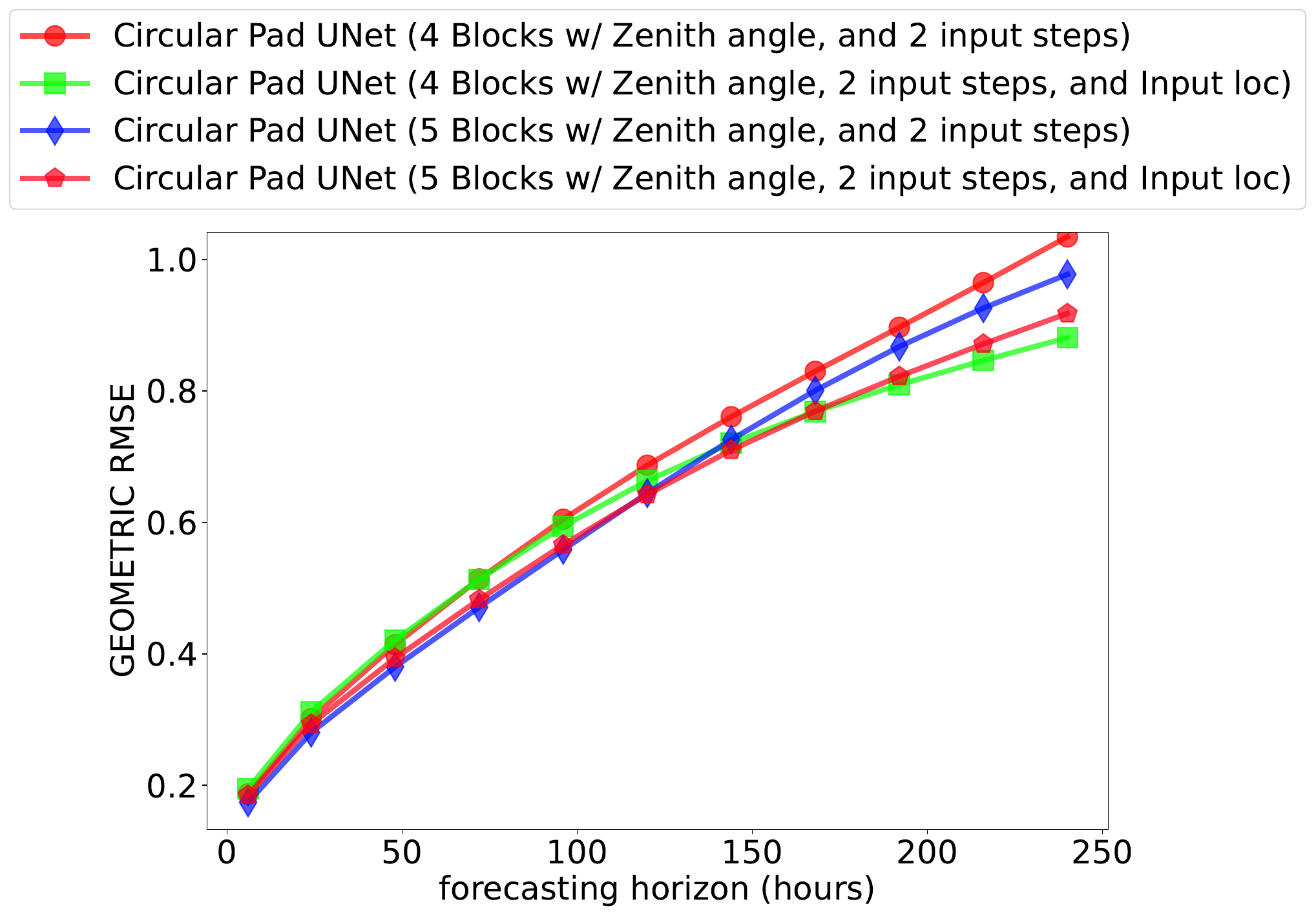}

    \caption{\textbf{Geometric ACC (left) and RMSE (right) with the addition of 3D coordinates on the sphere as static masks.} The figure highlights that providing the model with this auxiliary information regarding the proximity of different points on the globe results in more stable long-horizon forecasts (results in tabular form are presented in Table~\ref{input_loc_ood_num_steps_test_loss.mean}).}
    \label{fig:results_input_loc}
\end{figure}

\input{results_v2/input_loc/input_loc_ood_num_steps_test_loss.mean}

\begin{figure}[t]
    \centering

    \includegraphics[width=0.7\textwidth,trim={0cm 18.5cm 0cm 0cm},clip]{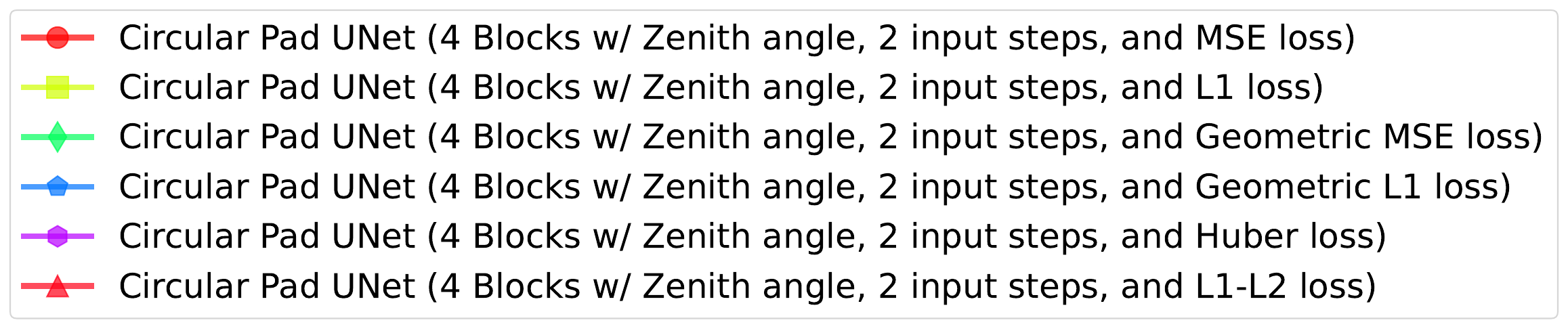}

    \includegraphics[width=0.495\textwidth,trim={4cm 0cm 7cm 8.5cm},clip]{results_v2/loss_models_4layers/loss_models_4layers_ood_num_steps_test_loss.geometric_acc.mean.pdf}
    \includegraphics[width=0.495\textwidth,trim={4cm 0cm 7cm 8.5cm},clip]{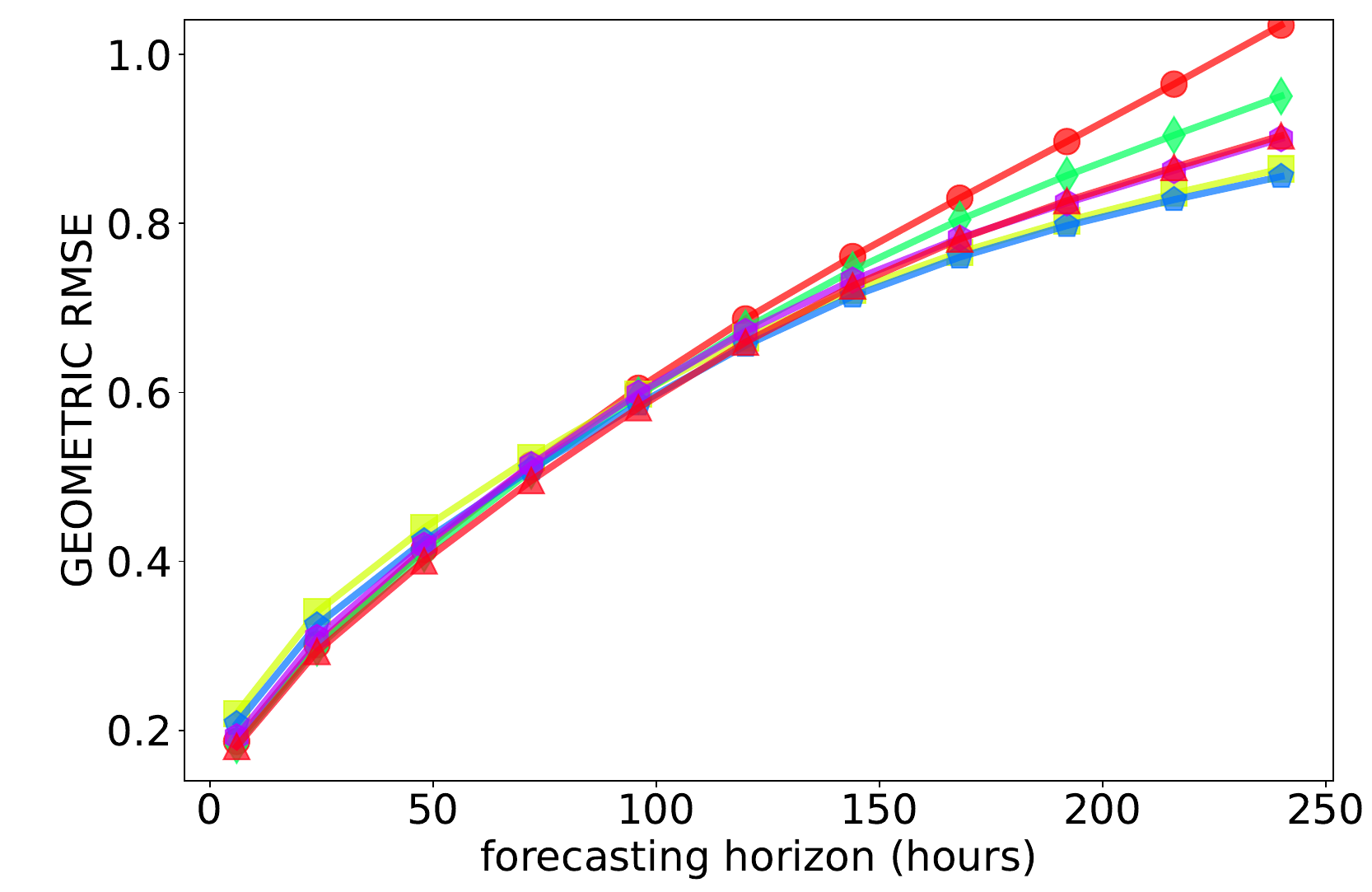}

    \caption{\textbf{Geometric ACC (left) and RMSE (right) with the use of different loss functions on 4-block UNet.} The figure highlights that variants of $L_1$ loss, including $L_1$, geometric $L_1$, $L_1$-$L_2$, and Huber loss, provide consistent gains in predictive performance in contrast to MSE, potentially due to their robustness against outliers (results in tabular form are presented in Table~\ref{loss_models_4layers_ood_num_steps_test_loss.mean}).}
    \label{fig:results_loss_fn_4layers}
\end{figure}

\input{results_v2/loss_models_4layers/loss_models_4layers_ood_num_steps_test_loss.mean}

\begin{figure}[t]
    \centering
    \includegraphics[width=0.7\textwidth,trim={0cm 18.5cm 0cm 0cm},clip]{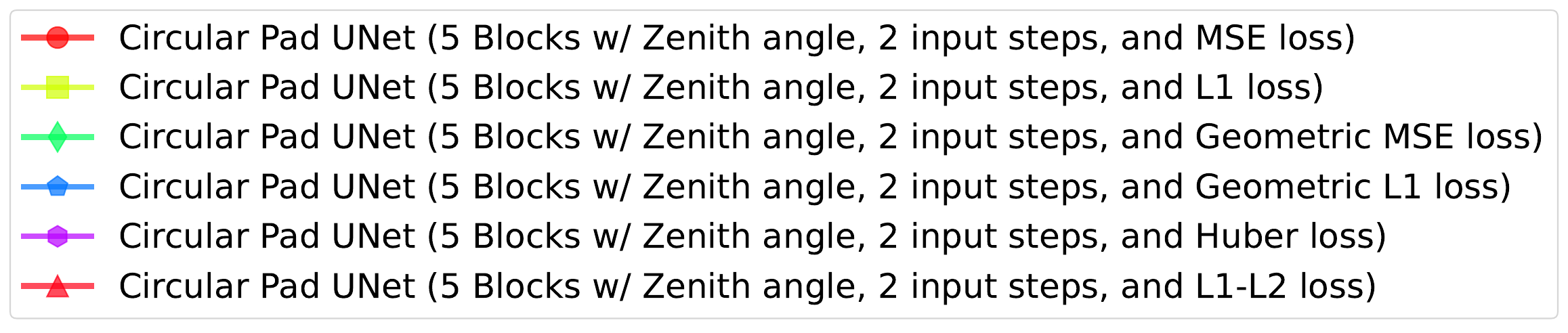}

    \includegraphics[width=0.495\textwidth,trim={4cm 0cm 7cm 8.5cm},clip]{results_v2/loss_models_5layers/loss_models_5layers_ood_num_steps_test_loss.geometric_acc.mean.pdf}
    \includegraphics[width=0.495\textwidth,trim={4cm 0cm 7cm 8.5cm},clip]{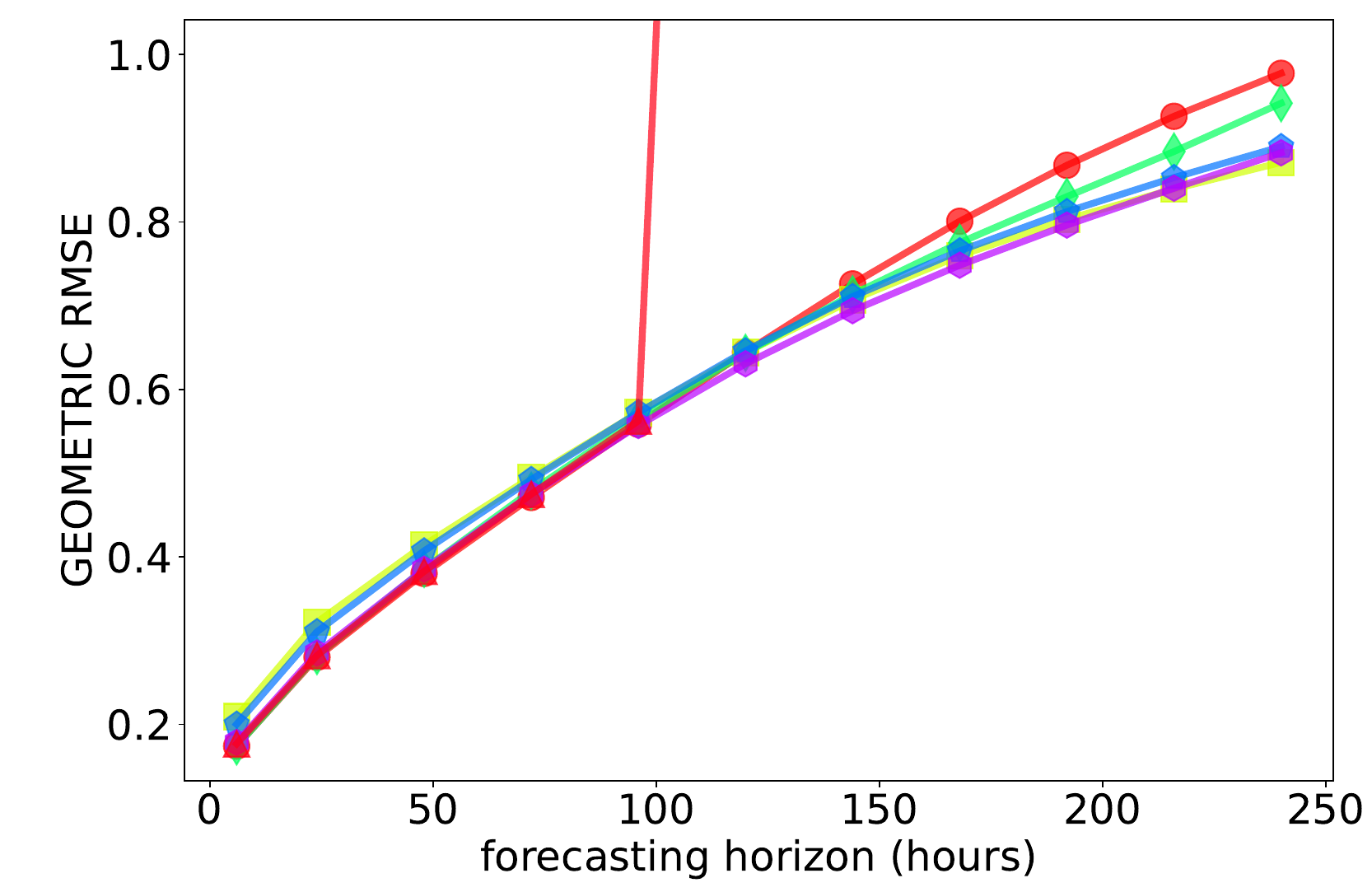}

    \caption{\textbf{Geometric ACC (left) and RMSE (right) with the use of different loss functions on 5-block UNet.} In contrast to the 4-block UNet, MSE dominates in the short-horizon prediction, while variants of $L_{1}$ loss dominate in the long-horizon prediction (results in tabular form are presented in Table~\ref{loss_models_5layers_ood_num_steps_test_loss.mean}).}
    \label{fig:results_loss_fn_5layers}
\end{figure}

\input{results_v2/loss_models_5layers/loss_models_5layers_ood_num_steps_test_loss.mean}

\subsection{Impact of using image-based pretraining}
\label{subsec:results_image_pretraining}

Weather models process different resolutions and channels in contrast to their image-based counterparts.
Despite this, image-based pertaining has been looked at as a form of careful initialization that is useful even in the presence of a task mismatch~\citep{kolesnikov2020bigtransfer,bommasani2021foundationmodels}.

Therefore, we aim to evaluate the impact of using pretrained weights obtained after image-based pretraining and finetune them for the weather forecasting task by reinitializing the first and the last layer to match the input and output channels.
Note that not all parameters are loaded i.e., some parameters are initialized randomly such as the decoder head that we attach to the original model.
This includes the input and output features as the image models are trained with just 3 input channels, while we have a significantly larger number of channels in our case.
In particular, we use the Segformer B5\footnote{\url{https://download.openmmlab.com/mmsegmentation/v0.5/pretrain/segformer/mit_b5_20220624-658746d9.pth}} pretrained backbone from MMSeg as our Segformer pretrained model and use the torchvision pretrained ResNet-50\footnote{\url{https://download.pytorch.org/models/resnet50-11ad3fa6.pth}}~\citep{he2016deep} as our ResNet-50 with conv-decoder pretrained model (which is only trained for classification).

The geometric ACC and RMSE for image-based pretraining are presented in Table~\ref{img_pretraining_ood_num_steps_test_loss.mean} (Fig.~\ref{fig:img_pretraining}).
There is a conflicting trend where we see performance improvement for the Segformer model while we observe performance degradation for ResNet-50.
We consider this to be primarily an artifact of the loaded model as only the base residual modules were successfully loaded for ResNet-50, leaving the decoder as well as the input stem to be initialized randomly.
Furthermore, the model is only trained for classification as compared to Segformer, which is trained directly for segmentation.
It would be interesting as a future step to train perfectly aligned models (except the input and the output layers which would be different) and redo this evaluation to get a clear sense of the transfer possible with image-based pretrained models.

\subsection{Impact of multi-step inputs}
\label{subsec:results_multi_step_inputs}

All our prior comparisons just took a single step as input.
However, it is possible to condition the model prediction on multiple previous input steps as commonly done in the past~\citep{lam2022graphcast,price2023gencast,chen2023fuxi}. This is particularly useful when doing long auto-regressive rollouts where the model can look at the previous steps to decide the right delta to be used for the next step. The problem formulation in this case is $\mathbf{s_{t}} = \mathbf{s_{t-1}} + \Phi(\mathbf{\{s_{t-n}, ..., s_{t-2}, s_{t-1}\}})$ where $n$ specifies the number of input time-steps.

The geometric ACC and RMSE for multi-step inputs are presented in Table~\ref{multistep_inputs_ood_num_steps_test_loss.mean} (Fig.~\ref{fig:multi_step_inputs}).
These results highlight that model performance is nearly maintained for 1 or 2 input steps but degrades significantly when using 4 input steps.
We consider the difference in performance between the 1 and 2 steps an artifact of the performance difference at the first step as evident by the constant difference between the model outputs.
Note that we are using delta prediction formulation, where we add the predicted delta to the last input step, which might reduce the utility of the inclusion of previous steps in contrast to direct prediction.
Furthermore, multiple input steps when combined with other techniques such as a longer training schedule might outperform the single-step performance.
As the performance degrades significantly with 4 input steps, we use two input steps for all subsequent experiments unless mentioned otherwise.

\subsection{Impact of zenith angle}
\label{subsec:results_zenith}

Solar zenith angle specifies the angle of the Sun to the vertical, providing a notion of time to the model.
Therefore, this has been commonly used in prior models~\citep{pathak2022fourcastnet}.
We also evaluate the impact of adding zenith angle as an additional input channel to the model.
For the auto-regressive rollout, we append the next zenith angle to the model's prediction before feeding it back to the model.
We only add the zenith angle for the last input step, even when using multi-step inputs as the delta is computed w.r.t. the last input step.

The geometric ACC and RMSE for zenith angle comparison are presented in Table~\ref{zenith_angle_ood_num_steps_test_loss.mean} (Fig.~\ref{fig:results_zenith}).
It is clear from the results that the zenith angle always improves performance for both short-horizon and long-horizon predictions by equipping the model with a notion of time.
Hence, all our subsequent models use the zenith angle unless mentioned otherwise.

\subsection{Impact of padding scheme}
\label{subsec:results_padding_scheme}

We evaluated the impact of different padding schemes. The default UNet padding scheme is zero padding on both sides.
As we have continuity along the longitudinal direction (x-axis), we also evaluate the impact of using a circular padding scheme in the longitudinal direction.
Furthermore, we evaluate the impact of adding the reflection padding scheme instead of the zero padding scheme for the latitudinal direction (y-axis) to account for the Neumann boundary condition.
\citet{mccabe2023spheretotorus} explored sphere-to-torus transform as a way to stabilize auto-regressive rollouts for operator-based models~\citep{pathak2022fourcastnet}.
Recently, \citet{cheon2024karina} also explored using a circular padding scheme.
However, they do not disentangle the marginal contribution of the padding scheme itself, which is a focus of our work.

The geometric ACC and RMSE for different padding schemes are presented in Table~\ref{padding_models_ood_num_steps_test_loss.mean} (Fig.~\ref{fig:results_padding_scheme}).
We do see a clear improvement when considering longer prediction horizons with circular padding along the longitudinal direction, which encodes the information regarding the geometry of the sphere.
We also see significant differences in performance when replacing zero padding with reflection padding along the latitudinal direction.
We consider simple ways to encode geometric information about the sphere within the model to be an exciting direction for further research.
We use circular padding along the longitudinal direction and zero padding along the latitudinal direction for all further experiments unless mentioned otherwise.

\begin{figure}[t]
    \centering

    \includegraphics[width=0.7\textwidth,trim={0cm 18.5cm 0cm 0cm},clip]{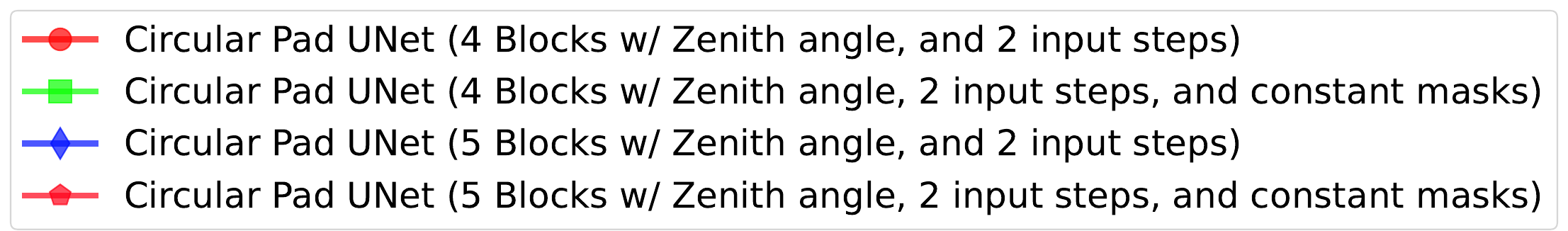}
    \vspace{1mm}

    \includegraphics[width=0.495\textwidth,trim={4cm 0cm 6cm 6cm},clip]{results_v2/constant_masks/constant_masks_ood_num_steps_test_loss.geometric_acc.mean.pdf}
    \includegraphics[width=0.495\textwidth,trim={4cm 0cm 6cm 6cm},clip]{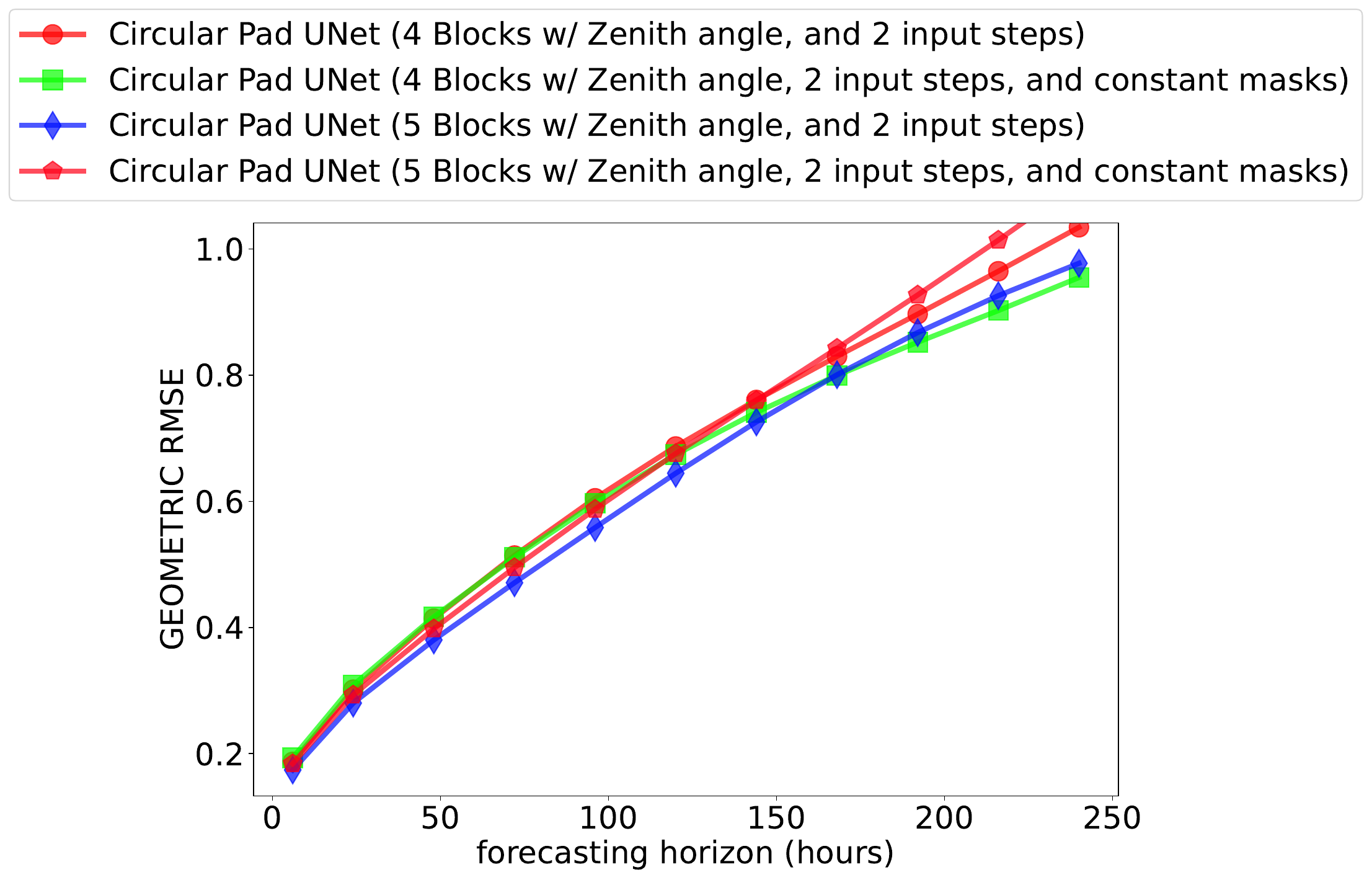}

    \caption{\textbf{Geometric ACC (left) and RMSE (right) with the addition of static constant masks including topography, soil-type, and land-sea mask following~\citep{bi2022pangu}.} The figure highlights an interesting trend where the performance is mostly hampered by the inclusion of constant masks when training for a small and fixed number of training steps as in our case (results in tabular form are presented in Table~\ref{constant_masks_ood_num_steps_test_loss.mean}).}
    \label{fig:results_constant_masks}
\end{figure}

\input{results_v2/constant_masks/constant_masks_ood_num_steps_test_loss.mean}

\begin{figure}
    \centering

    \includegraphics[width=0.7\textwidth,trim={0cm 18.5cm 0cm 0cm},clip]{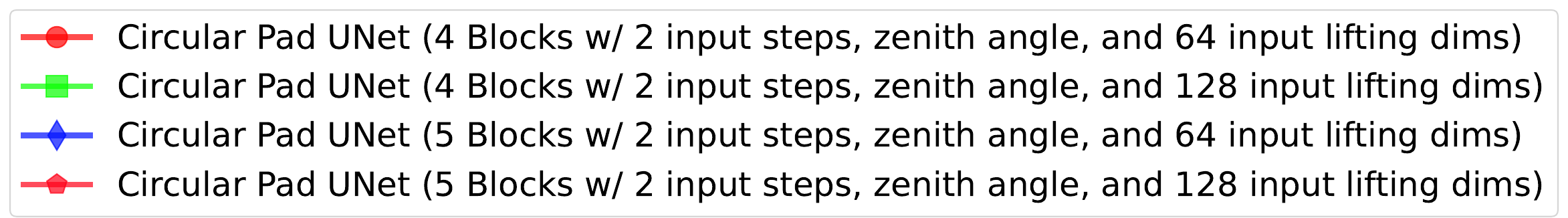}
    \vspace{1mm}

    \includegraphics[width=0.495\textwidth,trim={5cm 0cm 7cm 6cm},clip]{results_v2/wide/wide_ood_num_steps_test_loss.geometric_acc.mean.pdf}
    \includegraphics[width=0.495\textwidth,trim={5cm 0cm 7cm 6cm},clip]{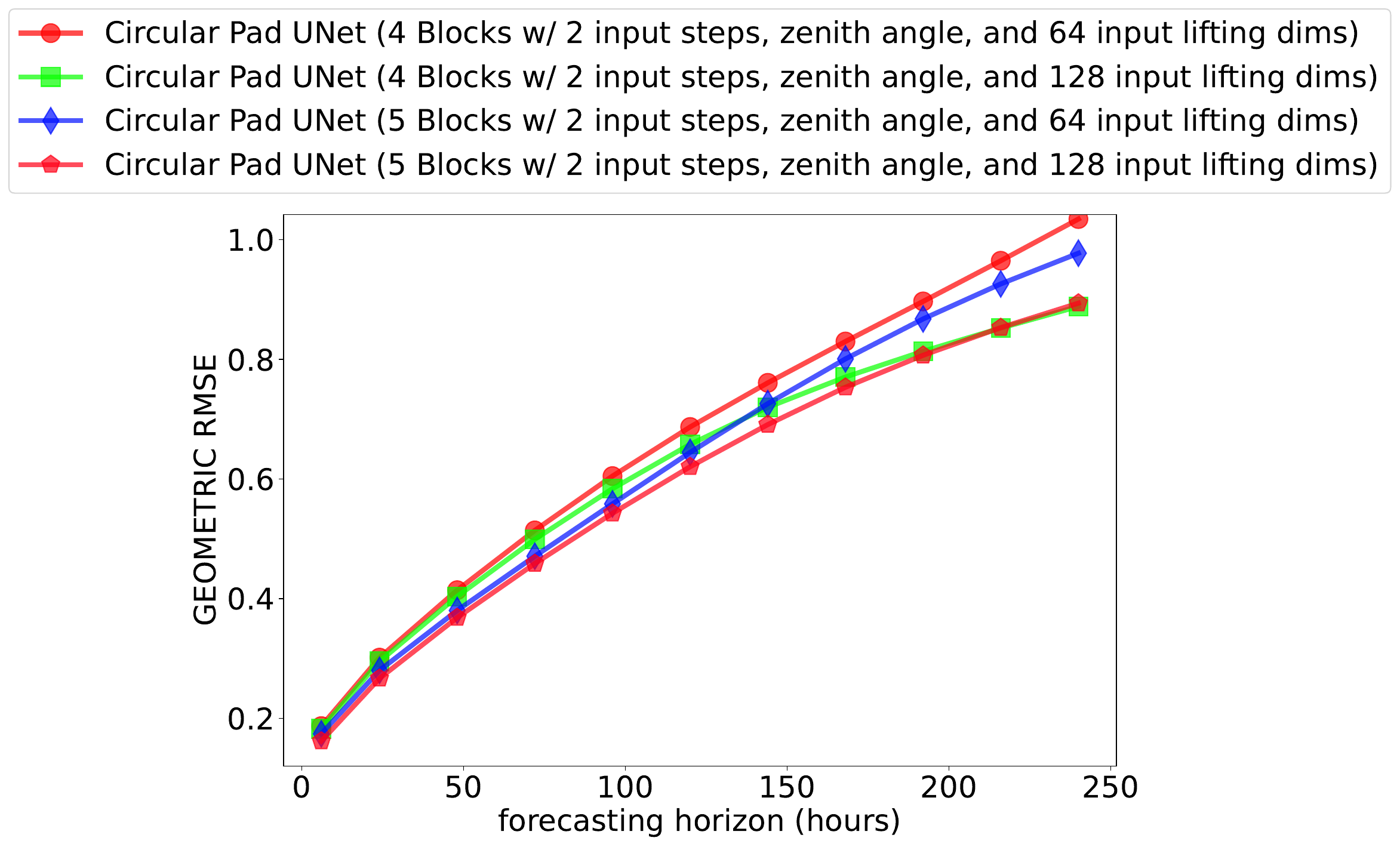}

    \caption{\textbf{Geometric ACC (left) and RMSE (right) with wider models i.e., larger hidden dimension.} We observe consistent gains with wider networks i.e., with an increased number of parameters for both 4-block and 5-block UNet variants (results in tabular form are presented in Table~\ref{wide_ood_num_steps_test_loss.mean}).}
    \label{fig:results_model_width}
\end{figure}

\input{results_v2/wide/wide_ood_num_steps_test_loss.mean}

\begin{figure}[t]
    \centering

    \includegraphics[width=0.8\textwidth,trim={0cm 18.25cm 0cm 0cm},clip]{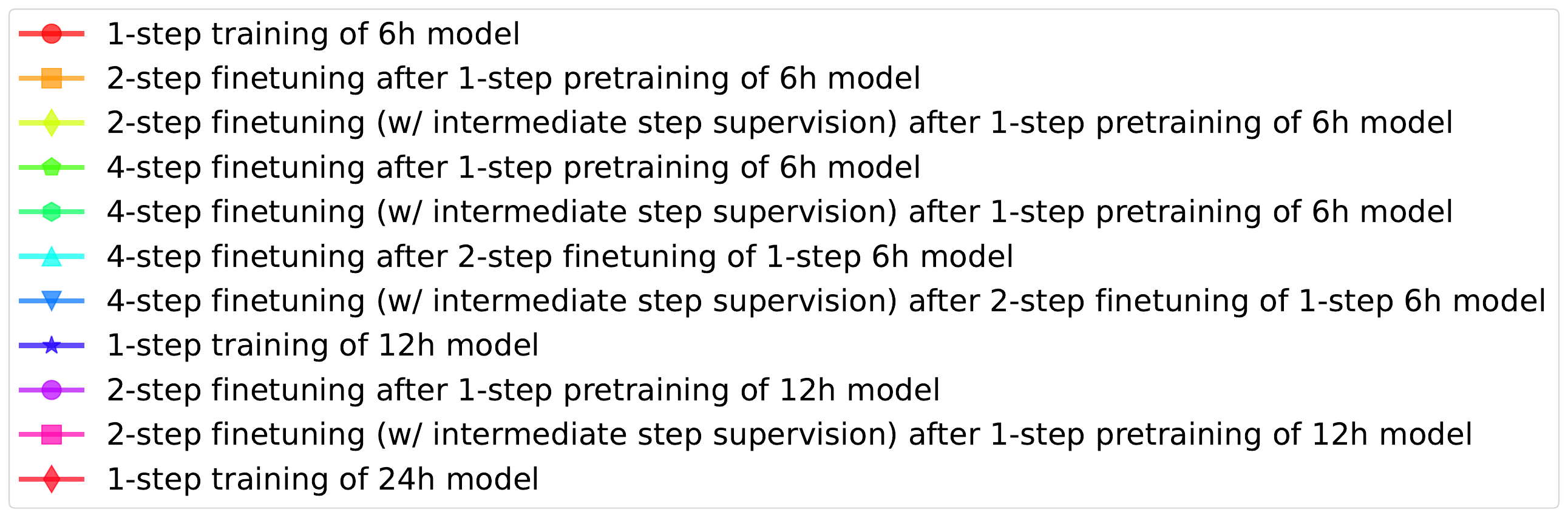}
    \vspace{1mm}

    \includegraphics[width=0.495\textwidth,trim={7cm 0cm 9cm 14.5cm},clip]{results_v2/ft_4layers/ft_4layers_ood_num_steps_test_loss.geometric_acc.mean.pdf}
    \includegraphics[width=0.495\textwidth,trim={7cm 0cm 9cm 14.5cm},clip]{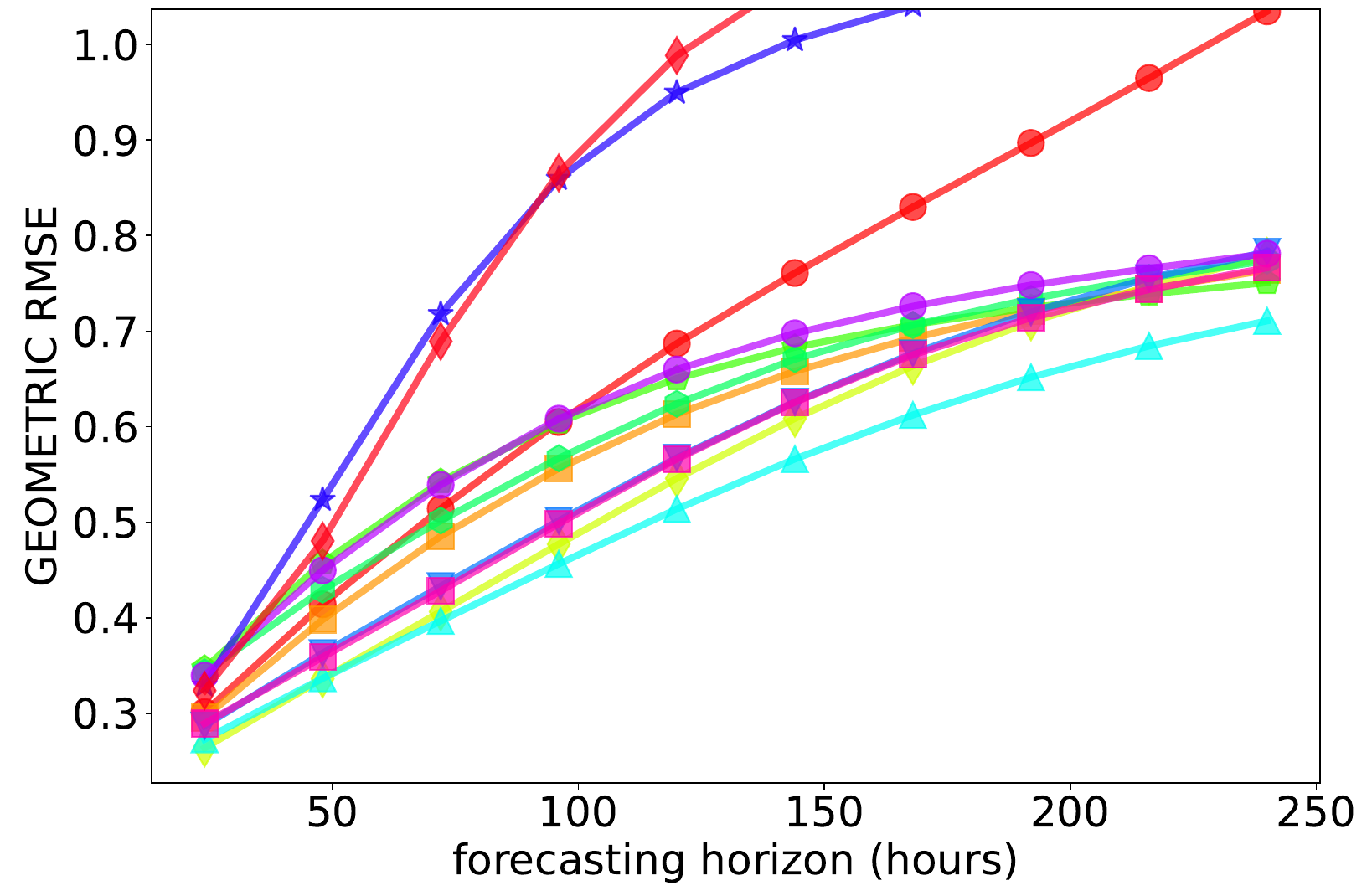}

    \caption{\textbf{Geometric ACC (left) and RMSE (right) with multi-step fine-tuning for 4-block UNet.} The figure highlights that training the model with the smallest stride considered i.e., 6h coupled with sequential fine-tuning provides the best performance, which has been the most common choice in the past (results in tabular form are presented in Table~\ref{ft_4layers_ood_num_steps_test_loss.mean}).}
    \label{fig:results_multistep_ft_4blocks}
\end{figure}

\input{results_v2/ft_4layers/ft_4layers_ood_num_steps_test_loss.mean}

\begin{figure}[t]
    \centering

    \includegraphics[width=0.8\textwidth,trim={0cm 18.25cm 0cm 0cm},clip]{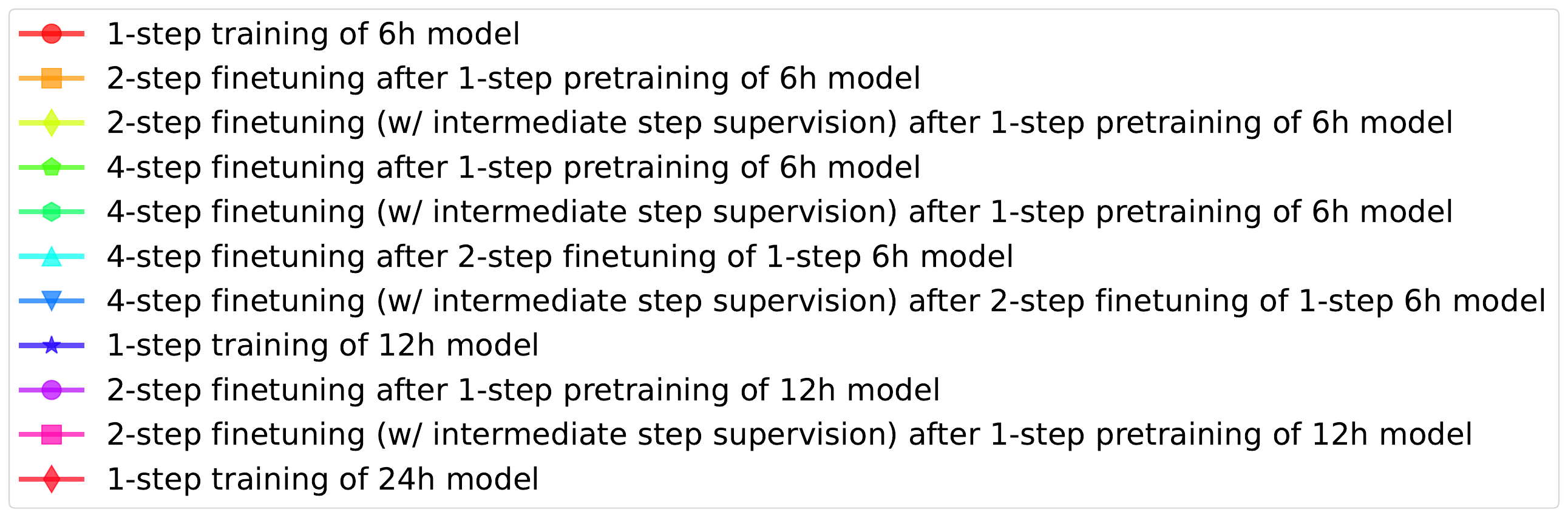}
    \vspace{1mm}

    \includegraphics[width=0.495\textwidth,trim={7cm 0cm 9cm 14.5cm},clip]{results_v2/ft_5layers/ft_5layers_ood_num_steps_test_loss.geometric_acc.mean.pdf}
    \includegraphics[width=0.495\textwidth,trim={7cm 0cm 9cm 14.5cm},clip]{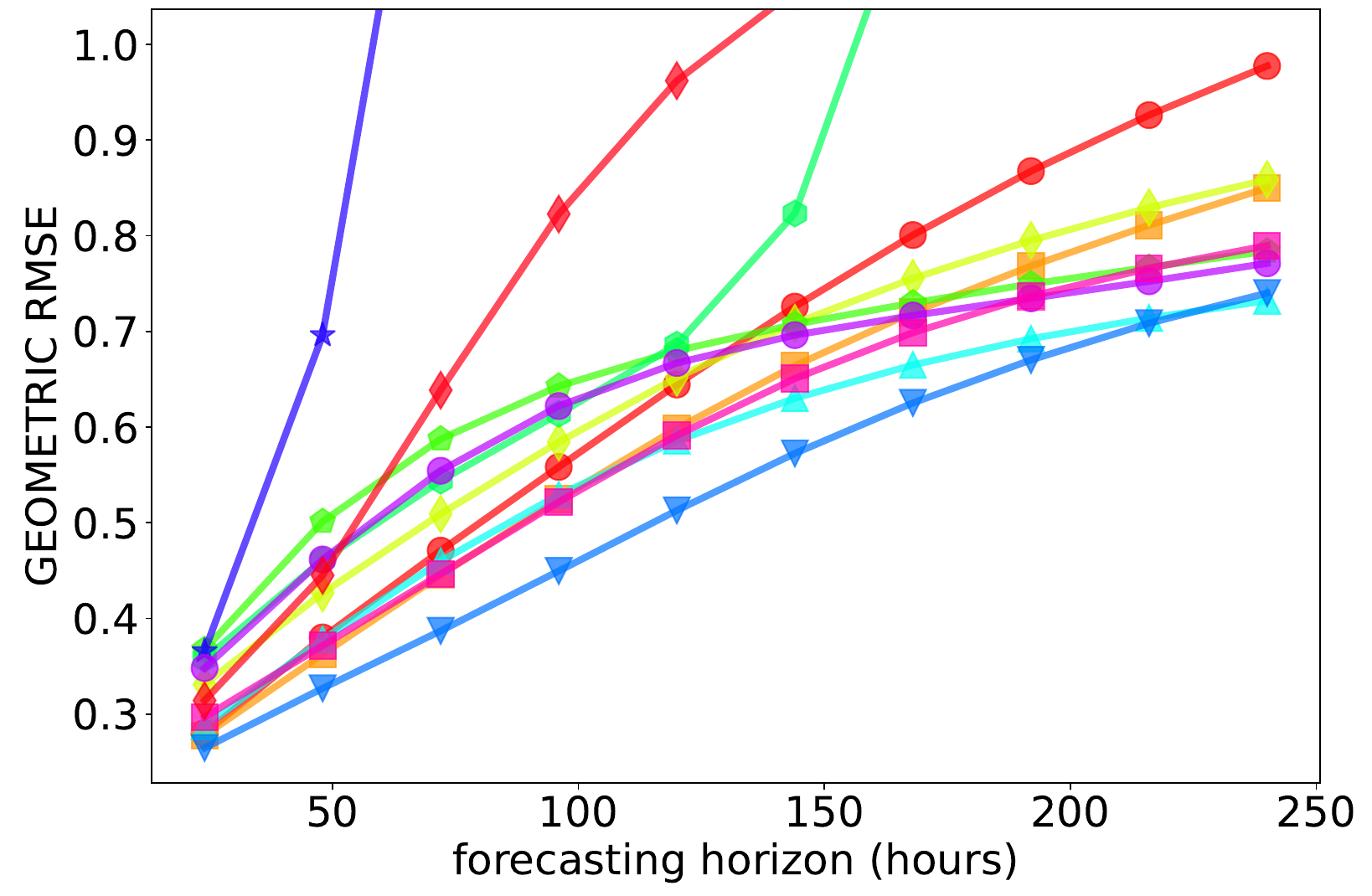}

    \caption{\textbf{Geometric ACC (left) and RMSE (right) with multi-step fine-tuning for 5-block UNet.} In contrast to 4-block UNet, we see that 5-block UNet benefits from intermediate step supervision for sequential fine-tuning (results in tabular form are presented in Table~\ref{ft_5layers_ood_num_steps_test_loss.mean}).}
    \label{fig:results_multistep_ft_5blocks}
\end{figure}

\input{results_v2/ft_5layers/ft_5layers_ood_num_steps_test_loss.mean}

\subsection{Impact of noise addition}
\label{subsec:results_noise_addition}

Noise addition has been used as an effective strategy for ensemble generation in the past~\citep{pathak2022fourcastnet,bi2022pangu}, where FourCastNet~\citep{pathak2022fourcastnet} added Gaussian noise while Pangu-weather~\citep{bi2022pangu} and FuXi~\citep{chen2023fuxi} applied Perlin noise~\citep{perlin2002noise} to the initial states.
Given these evaluations, it is useful to understand the impact of noise addition on model performance during training as a regularization technique~\citep{lopes2019patchgaussian}.
Therefore, we evaluate the impact of additive Gaussian and Perlin noise during model training.

The geometric ACC and RMSE for noise addition are presented in Table~\ref{noise_models_ood_num_steps_test_loss.mean} (Fig.~\ref{fig:results_noise_add}).
The figure highlights that adding noise improves model performance in the long-horizon prediction while having a negligible detrimental effect on the short-horizon forecast.
In particular, Perlin noise~\citep{perlin2002noise}, being structured, outperforms unstructured additive Gaussian noise.
This gain in performance, however, is very likely to impact the ability to generate prediction ensembles via input state perturbation as the model learns to be invariant to this additive noise.

Note that we only evaluate on a fixed noise level, and skip a detailed ablation over different noise levels.
We consider a more thorough analysis regarding the impact of noise during model training, not only for deterministic forecasts but also when trying to generate prediction ensembles as the model is directly trained with noise as interesting directions to explore further.

\subsection{Impact of appending 3D coordinates}
\label{subsec:results_input_loc}

As a fixed-sized grid is not natural for weather forecasting due to the underlying spherical structure of the globe, and limits querying the model on any arbitrary grid point, resolution invariant models that use the query location coordinates as input to generate predictions have been a major focus~\citep{pathak2022fourcastnet,bonev2023sfno,lam2022graphcast}.

This 3D coordinate information is useful for the model to understand the proximity in 3D space (specifically close to the poles due to oversampling when projecting to a fixed-sized lat-long grid). Latitude-weighted loss functions have been used to circumvent this issue partially due to this oversampling.
In our architecture search, we also used several 3D methods that directly use this 3D coordinate information.
This coordinate information can also potentially be useful for even image-based models as used in prior models such as CoordConv~\citep{liu2018coordconv}.

Therefore, we experiment with appending 3D coordinates converted from the lat-long grid as three additional input channels representing the x, y, and z coordinates on the sphere respectively.
As this is input-independent, we add a single set of x, y, and z coordinates even for multi-step models.

The geometric ACC and RMSE with the inclusion of 3D location coordinates as additional static channels are presented in Table~\ref{input_loc_ood_num_steps_test_loss.mean} (Fig.~\ref{fig:results_input_loc}).
The results highlight the positive impact of appending the 3D coordinates, specifically for the long-horizon prediction task.
There is however a disparity between the two model sizes where 5-block UNet seems to be benefitting more from this auxiliary information.
This highlights the fact that the utility of each of these approaches can change significantly given the exact model in question.
This is a recurring theme for many of our subsequent results.

\subsection{Impact of loss functions}
\label{subsec:results_loss_fn}

Most prior works experimented with different loss functions, such as Geometric MSE~\citep{pathak2022fourcastnet,bonev2023sfno,lam2022graphcast,nguyen2023stormer} or geometric $L_1$~\citep{chen2023fuxi}.
We attempt to evaluate the impact of different loss functions on the predictive performance of the resulting model, which includes regular MSE which is equivalent to $L_2^2$ (the default loss function that we use to train all our models), geometric MSE, $L_1$ loss, geometric $L_1$, Huber (smooth $L_1$), and $L_1$-$L_2$ loss (with a weight of 0.05 assigned to $L_1$ loss and 0.95 assigned to $L_2^2$ loss).
Note that we did not explore the impact of weighing the variables or the pressure levels differently as explored in some other works~\citep{lam2022graphcast,nguyen2023stormer}.

\subsubsection{UNet (4 Blocks)}

The geometric ACC and RMSE for different loss functions on the 4-block UNet model are presented in Table~\ref{loss_models_4layers_ood_num_steps_test_loss.mean} (Fig.~\ref{fig:results_loss_fn_4layers}).
The figure highlights the superiority of $L_1$-$L_2$ loss when considering short-horizon predictions.
However, when considering long-horizon predictions, geometric $L_1$ loss dominates.
In general, we observe better performance with $L_1$ variants in contrast to $L_2$ variants potentially due to its capacity to ignore outliers, which is particularly helpful during long auto-regressive rollouts.

\subsubsection{UNet (5 Blocks)}

The geometric ACC and RMSE for different loss functions on the 5-block UNet model are presented in Table~\ref{loss_models_5layers_ood_num_steps_test_loss.mean} (Fig.~\ref{fig:results_loss_fn_5layers}).
In contrast to the 4-block UNet, $L_2^2$ (MSE) variants dominate for the short-horizon forecast, while variants of $L_1$ loss dominate in the long-horizon prediction.
Interestingly, $L_1$-$L_2$ loss diverges in this case, highlighting a potential optimization instability with $L_1$ variants.
As seen previously, this highlights that different models might be best suited to different loss functions and hence, require a model-specific evaluation.

\subsection{Impact of constant masks}
\label{subsec:results_constant_masks}

Constant masks are static masks that provide additional information to the model that might be useful for prediction.
Following~\citet{bi2022pangu}, we consider three constant masks including topography, soil type, and land-sea mask.
As these masks are static, we only append them once even during the auto-regressive rollout.
Since all our input channels are standardized based on the dataset mean and standard deviation, we also standardize the constant masks based on the computed mean and standard deviation (note that this is again a constant as we are using static masks).

The geometric ACC and RMSE for constant mask comparison are presented in Table~\ref{constant_masks_ood_num_steps_test_loss.mean} (Fig.~\ref{fig:results_constant_masks}).
We see a less pronounced effect in comparison to the zenith angle which always helps.
We again see a model-specific impact where performance improves for the 4-block variant, but degrades for the 5-block variant of UNet.
This is surprising as we would expect larger models to make better use of additional information. We speculate this to be an artifact of overfitting on these auxiliary static masks.

\begin{figure}[t]
    \centering
    \includegraphics[width=0.55\textwidth,trim={0cm 18.25cm 0cm 0cm},clip]{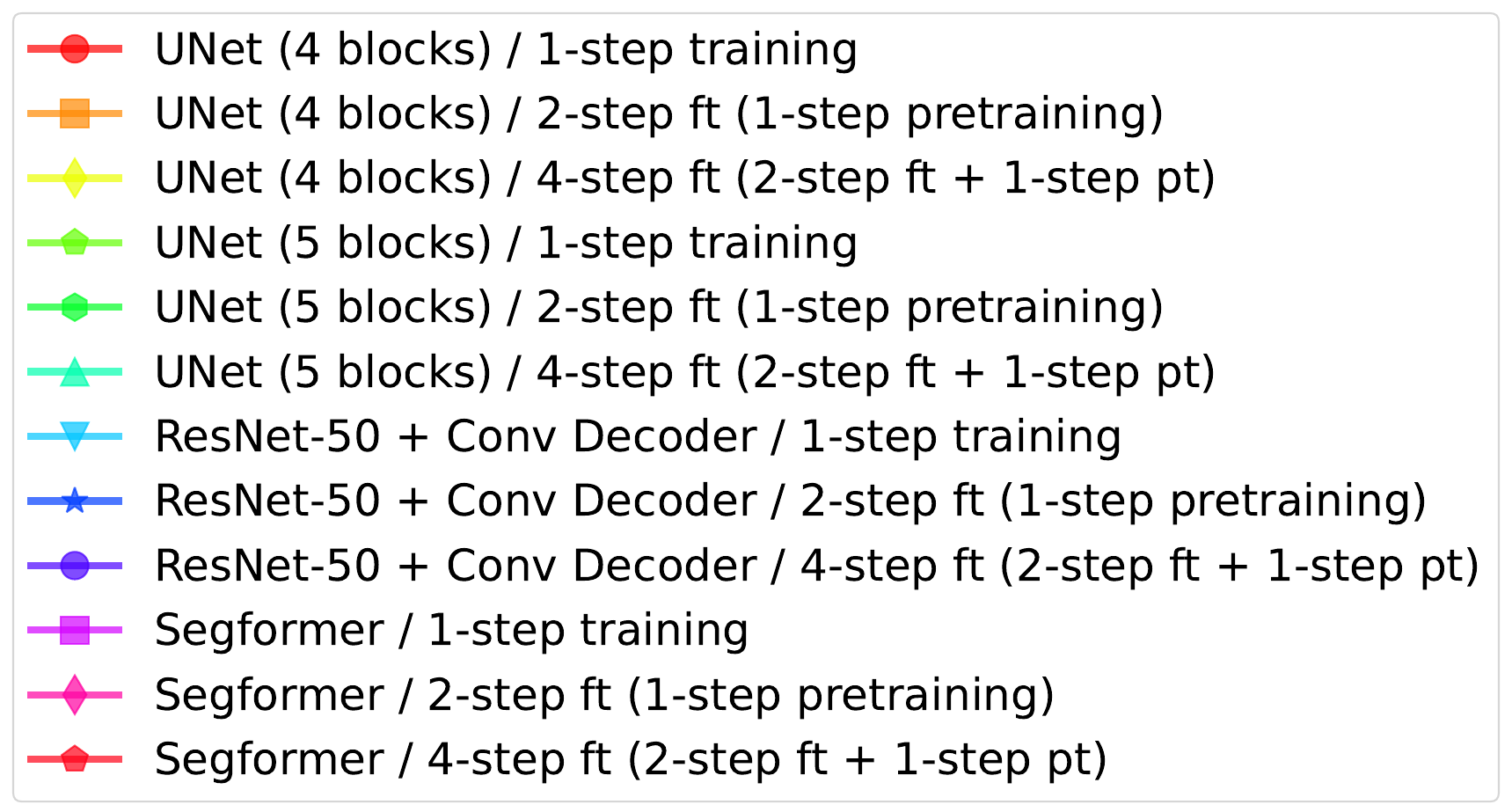}

    \includegraphics[width=0.495\textwidth,trim={0.5cm 0cm 0.5cm 15.5cm},clip]{results_v2/compare_multistep_ft_models/compare_multistep_ft_models_ood_num_steps_test_loss.geometric_acc.mean.pdf}
    \includegraphics[width=0.495\textwidth,trim={0.5cm 0cm 0.5cm 15.5cm},clip]{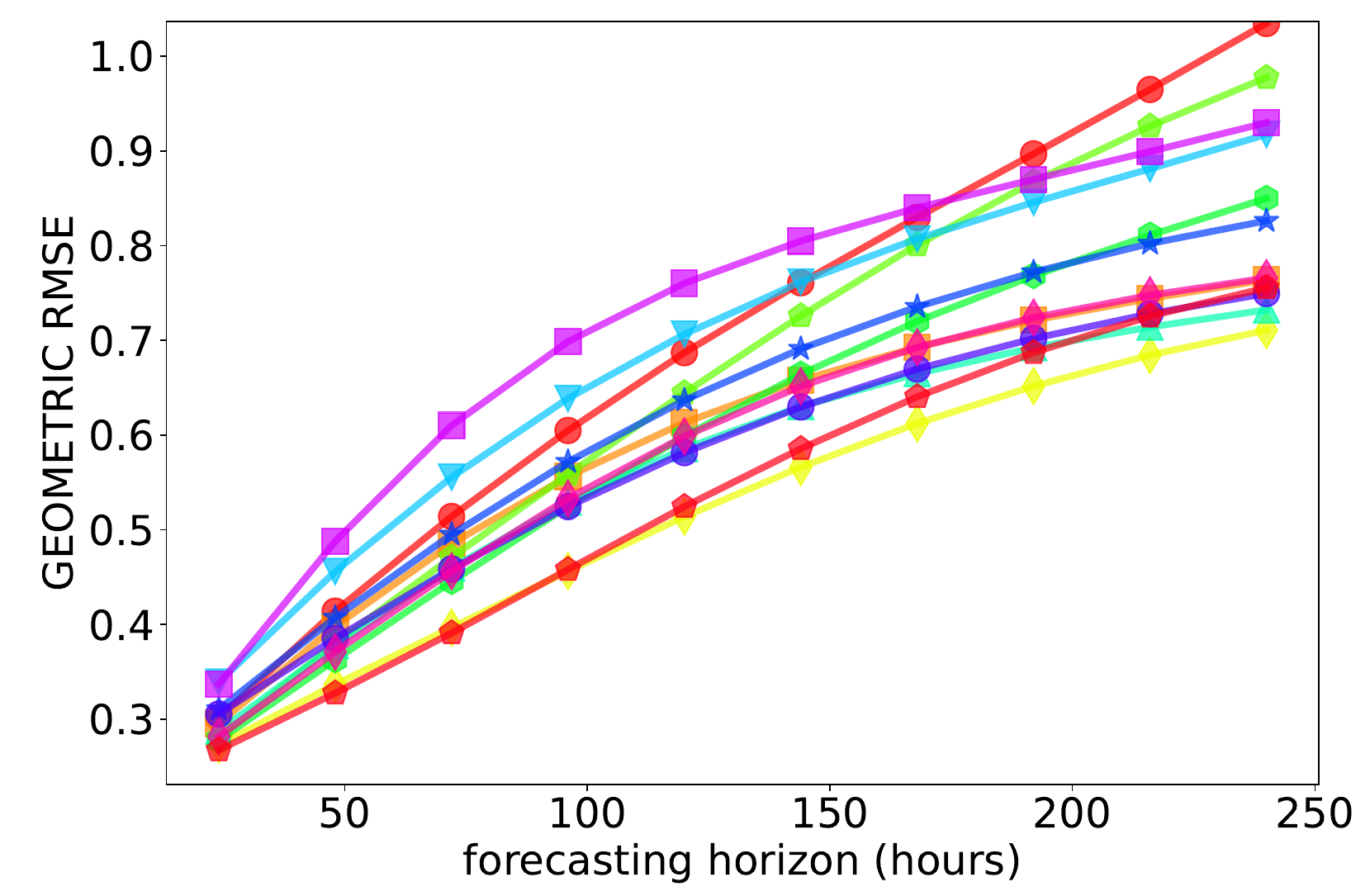}

    \caption{\textbf{Geometric ACC (left) and RMSE (right) with multi-step fine-tuning for comparing other promising models alongside UNet including ResNet-50 w/ convolutional decoder and Segformer.} All models are trained with the zenith angle. We see that multi-step fine-tuning improves performance significantly for all models evaluated (results in tabular form are presented in Table~\ref{compare_multistep_ft_models_ood_num_steps_test_loss.mean}).}
    \label{fig:results_multistep_ft}
\end{figure}

\input{results_v2/compare_multistep_ft_models/compare_multistep_ft_models_ood_num_steps_test_loss.mean}

\begin{figure}[t]
    \centering

    \includegraphics[width=0.6\textwidth,trim={0cm 18.25cm 0cm 0cm},clip]{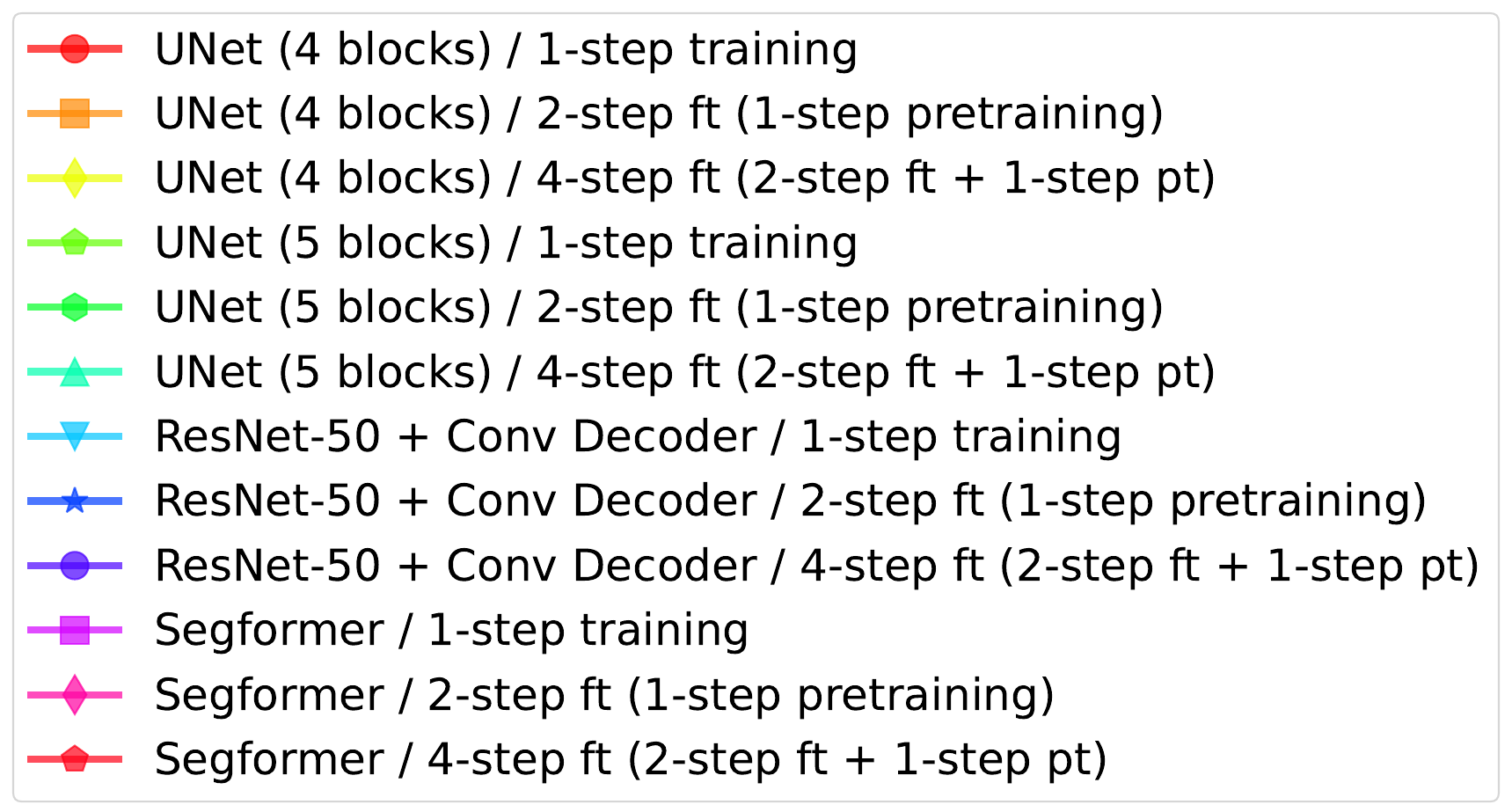}

    \includegraphics[width=0.495\textwidth,trim={0.5cm 0cm 0.5cm 15.5cm},clip]{results_v2/short_term_compare_multistep_ft_models/short_term_compare_multistep_ft_models_ood_num_steps_test_loss.geometric_acc.mean.pdf}
    \includegraphics[width=0.495\textwidth,trim={0.5cm 0cm 0.5cm 15.5cm},clip]{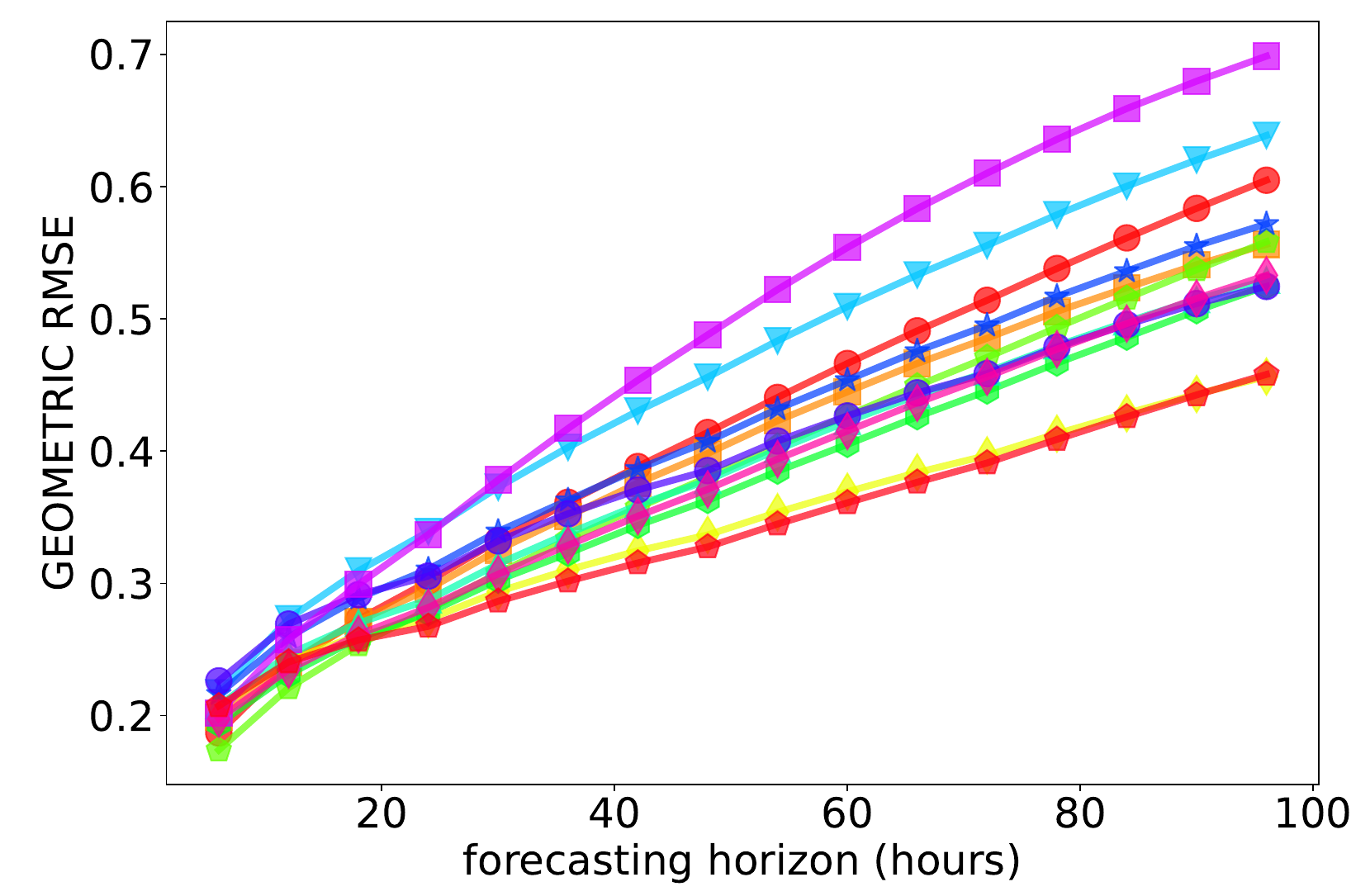}

    \caption{\textbf{Geometric ACC (left) and RMSE (right) when focusing on short-horizon forecast with multi-step fine-tuning for comparing other promising models alongside UNet including ResNet-50 w/ convolutional decoder and Segformer.} All models are trained with the zenith angle. We see that multi-step fine-tuning improves performance significantly for all models evaluated (see Fig.~\ref{fig:results_multistep_ft} for results over longer horizon -- results in tabular form are presented in Table~\ref{short_term_compare_multistep_ft_models_ood_num_steps_test_loss.mean}).}
    \label{fig:results_short_term_multistep_ft}
\end{figure}

\input{results_v2/short_term_compare_multistep_ft_models/short_term_compare_multistep_ft_models_ood_num_steps_test_loss.mean}

\begin{figure}[t]
    \centering

    \includegraphics[width=0.7\textwidth,trim={0cm 18.25cm 0cm 0cm},clip]{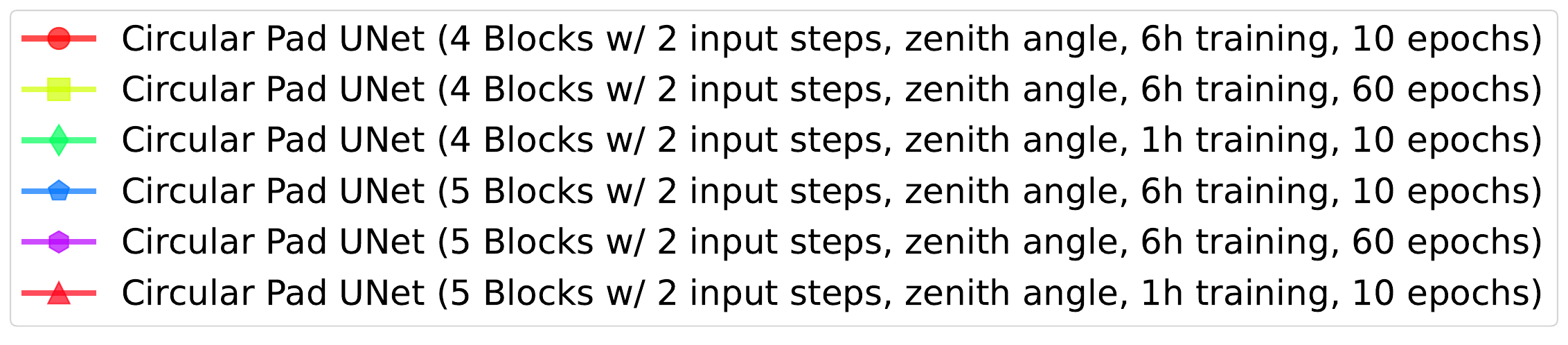}

    \includegraphics[width=0.495\textwidth,trim={4cm 0cm 6.5cm 8.5cm},clip]{results_v2/hourly/hourly_ood_num_steps_test_loss.geometric_acc.mean.pdf}
    \includegraphics[width=0.495\textwidth,trim={4cm 0cm 6.5cm 8.5cm},clip]{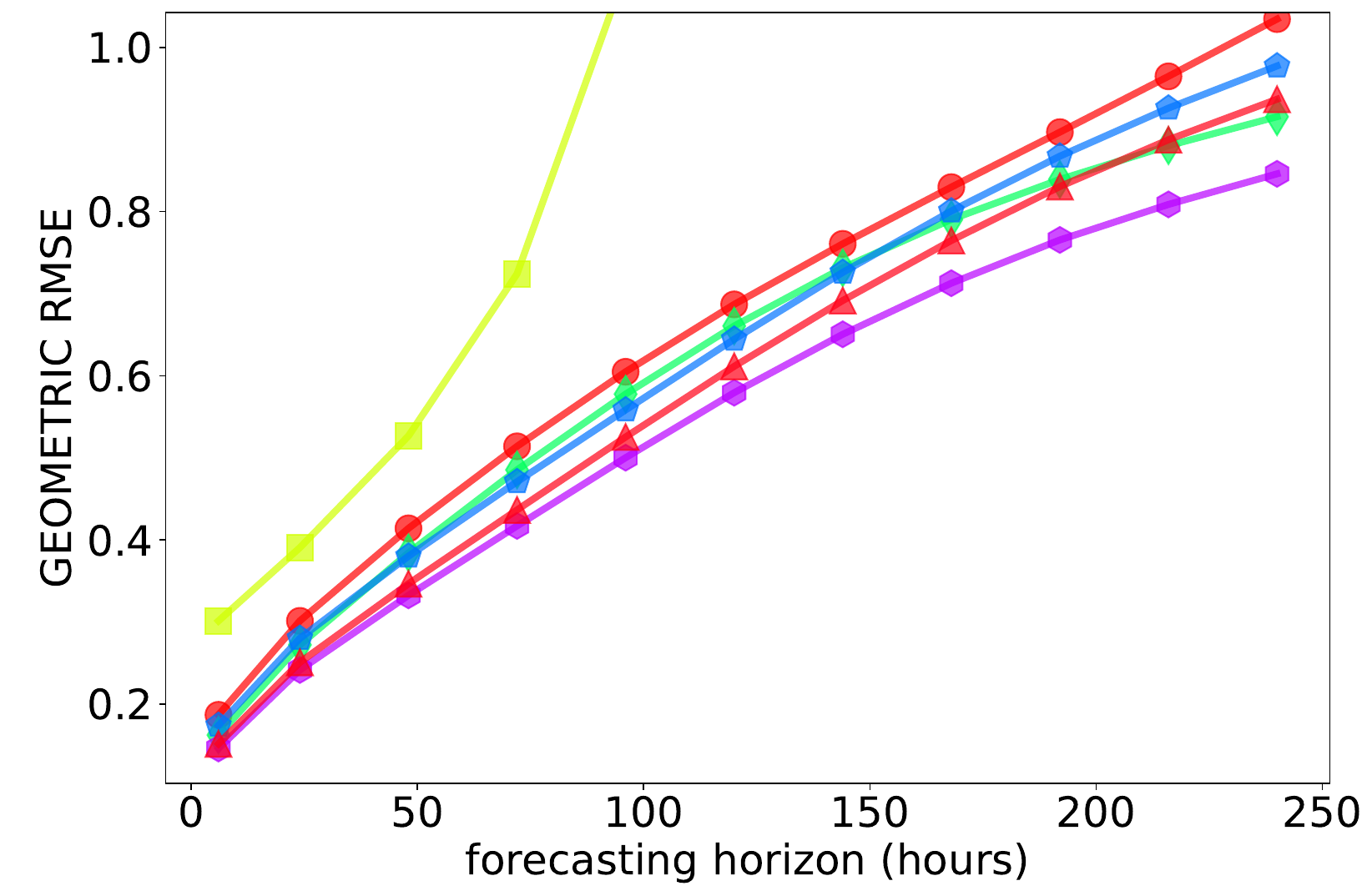}

    \caption{\textbf{Geometric ACC (left) and RMSE (right) when training on a larger dataset.} The figure highlights that training on the smaller dataset for longer is detrimental to the performance of the 4-block UNet which shows signs of overfitting, but training on the larger dataset improves performance. On the other hand, 5-block UNet effectively achieves similar performance in both cases, highlighting that different models might benefit differently from increased training data (results in tabular form are presented in Table~\ref{hourly_ood_num_steps_test_loss.mean}).}
    \label{fig:results_hourly}
\end{figure}

\input{results_v2/hourly/hourly_ood_num_steps_test_loss.mean}

\subsection{Impact of wider models}
\label{subsec:results_wider_models}

The interplay between the depth and width of the model has been an interesting point of discussion in the past~\citep{zagoruyko2016wideresnets}. 
In order to understand the gain in performance with model width, we experiment with increasing the starting embedding dimension of our model from 64. Note that we have an hourglass architecture, which results in doubling the number of channels after every spatial downsampling stage following the original UNet architecture~\citep{ronneberger2015unet}. In this case, we increase the number of channels from 64 which was the default for all our prior experiments to 128.

The geometric ACC and RMSE with an increased model width are presented in Table~\ref{wide_ood_num_steps_test_loss.mean} (Fig.~\ref{fig:results_model_width}).
The results highlight that increasing the hidden dimension of the model from 64 to 128 boosts model performance for both the 4-block and 5-block variants.
However, the gain is more pronounced for the 5-block UNet, despite having an extremely large number of parameters in contrast to all other models (i.e., 1.6 billion parameters as highlighted in Table~\ref{tab:num_params}).
These findings are consistent with prior results in the computer vision literature, where they found a prominent effect of model width on performance~\citep{zagoruyko2016wideresnets}.

\subsection{Impact of multi-step fine-tuning}
\label{subsec:results_multistep_ft}

With the notable exception of \citet{bi2022pangu} where they used multiple models operating at different temporal resolutions, and \citet{price2023gencast} where a diffusion model was employed, all prior works rely on fine-tuning the base model for multi-step forecasting to enable the model to generate smooth long-horizon predictions~\citep{pathak2022fourcastnet,lam2022graphcast,bonev2023sfno}.
There are different choices involved in this case, where a complicated combination of temporal resolution, number of auto-regressive steps, and appropriate fine-tuning supervision are potential choices that one encounters.

We therefore attempt to understand the marginal impact of each of these complex design choices.
We evaluate two variants of fine-tuning supervision in this case i.e.,  intermediate-step supervision and only last-step supervision.
When using intermediate-step supervision, we use a discount factor of 0.9 for all the subsequent prediction steps starting with a weight of 1.0, in order to represent higher uncertainty as commonly used in reinforcement learning~\citep{sutton2018reinforcement}.
Even in the case of just supervising the model on the last step, the gradient is computed through the auto-regressive rollout trajectory.
Note that the initial input is the same in all cases i.e., two input steps with a stride of 6h are concatenated regardless of the prediction temporal resolution of the model for consistency between results.

We only consider delta prediction formulation for our experiments.
However, delta prediction might not be well-suited for larger stride models beyond the 6h horizon considered here.
Therefore, there is a chance that these experiments greatly undermine the actual potential of larger stride models.
We consider resolving this discrepancy and doing a more detailed analysis as an important future direction.

\subsubsection{UNet (4 Blocks)}

The results in terms of geometric ACC and RMSE for the 4 block UNet are presented Table~\ref{ft_4layers_ood_num_steps_test_loss.mean} (Fig.~\ref{fig:results_multistep_ft_4blocks}).
The results show a clear trend where fine-tuning the model on subsequently larger strides while only supervising the last step provides the best performance, surprisingly even better than supervising the intermediate steps of the model.
This reaffirms the common practice of sequentially fine-tuning a small stride model (such as 6h) on increasingly larger prediction horizons~\citep{pathak2022fourcastnet,bonev2023sfno,lam2022graphcast}.

Note that there is also a confounding variable in this case in the form of compute time.
As each of our runs has a fixed budget of 10 epochs, sequential fine-tuning results in a direct increase in the allocated computational budget.
The obtained conclusions might differ if each of these runs is trained till convergence.

\subsubsection{UNet (5 Blocks)}

The results for the 5-block UNet are presented in Table~\ref{ft_5layers_ood_num_steps_test_loss.mean} (Fig.~\ref{fig:results_multistep_ft_5blocks}).
Similar to the 4-block UNet, we see that the model benefits from sequential fine-tuning.
However, in contrast to the 4-block model, the model benefits from intermediate-step supervision.
This again highlights that the exact fine-tuning recipe to be used might differ based on the model in question.

\subsection{Comparing best models with multi-step fine-tuning}
\label{subsec:multistep_ft_compare_models}

We initially picked one model i.e., UNet~\citep{ronneberger2015unet} out of multiple other promising models including ResNet-50 with fully-convolutional decoder~\citep{long2015fcn}, Segformer~\citep{xie2021segformer}, and SETR~\citep{zheng2021setr} from our initial architecture search in Section~\ref{subsec:results_arch}.
To understand the full potential of these models, we evaluate them again with the inclusion of the uncovered important design choices i.e., using 2-input steps, zenith angle, and multi-step fine-tuning.

We compare the performance between these models in Table~\ref{compare_multistep_ft_models_ood_num_steps_test_loss.mean} (Fig.~\ref{fig:results_multistep_ft}).
We adopt the sequential fine-tuning scheme in this case, supervising only the last step.
Since this is ideal for the 4-block UNet, but not the 5-block variant, we see that the performance of our 4-block model outcompetes the 5-block variant in this case.
Furthermore, Segformer achieves performance comparable to our 4-block UNet, while ResNet-50 with convolutional decoder follows the performance of our 5-block UNet.
Note that Segformer without fine-tuning is the worst model in the plot with a large margin, again highlighting the importance of multi-step fine-tuning for these models to work well.

Note that we excluded SETR for comparison here despite it being a competitive model due to memory constraints.
We assume the performance for SETR would be similar to that of Segformer as both are based on the transformer architecture~\citep{vaswani2017attention}.

\subsubsection{Focusing on short-horizon forecasting}

As our previous results were plotted with a larger stride of a day, we now focus on having a deeper look at the short-term forecast by plotting the results at the lowest granularity i.e., 6h prediction.
The results are presented in Table~\ref{short_term_compare_multistep_ft_models_ood_num_steps_test_loss.mean} (Fig.~\ref{fig:results_short_term_multistep_ft}) respectively.
These results are fully consistent with previous results due to smoothness where we see best performance from Segformer and 4-block UNet variants.

\subsection{Impact of training on a larger dataset}
\label{subsec:results_hourly_training}

All our experiments prior focused on training the model on a 6h dataset variant sampled at 0000, 0600, 1200, and 1800 UTC~\citep{pathak2022fourcastnet,lam2022graphcast,bonev2023sfno}.
However, ERA5 naturally provides an hourly snapshot.
Therefore, we further evaluate the impact of training the model on this hourly snapshot while retaining the 6h prediction horizon, resulting in 6x more training examples.
Training the model on this hourly dataset for 10 epochs is equivalent to training the model on the 6h dataset for 60 epochs in terms of the number of training steps.
Therefore, we add the 60 epochs baseline on the 6h dataset as a comparison which matches the computational budget for training on the 1h dataset for 10 epochs.
For a fair comparison, we evaluate our models on the same 6h out-of-sample set that we used for all our prior evaluations.

The results are presented in Table~\ref{hourly_ood_num_steps_test_loss.mean} (Fig.~\ref{fig:results_hourly}).
Similar to some of our prior results, we observe differences between the 4-block and 5-block UNet variants in this case, where training on the smaller dataset for longer degrades performance.
This can be attributed to the overfitting of the model on the training set.
However, the performance improves when training on the larger dataset.
Our 5-block UNet on the other hand achieves similar performance in both cases i.e., training for longer on the smaller dataset and training for shorter on the larger dataset.

As the performance is always better or on par with the performance achieved by training on the smaller dataset, this indicates that using more data should be preferred wherever possible instead of training for longer on the smaller dataset.

%% file: sections/num_params_table.tex
\begin{table*}[t]
\centering
\begin{tabular}{l r}
\toprule

\textbf{Model Name} & \textbf{\# of Parameters} \\
\midrule

FNO~\citep{li2020fno} & 233,039,081 \\
FourCastNet~\citep{pathak2022fourcastnet} & 79,900,416 \\
SFNO~\citep{bonev2023sfno} & 289,392,337 \\
UNO~\citep{ashiqur2022uno} & 145,418,729 \\

ISNet~\citep{qin2022isnet} & 45,114,917 \\
ResNet-50 + Fully Convolutional Decoder~\citep{long2015fcn} & 47,202,505 \\
Segformer~\citep{xie2021segformer} & 82,208,137 \\
Segmentation Transformer (SETR)~\citep{zheng2021setr} & 343,751,497 \\
Swin Transformer~\citep{liu2021swin} & 59,086,083 \\
Efficient ViT L2~\citep{cai2022efficientvit} & 51,403,689 \\
PanguWeather~\citep{bi2022pangu} & 64,247,821 \\
UNet (4 Layers)~\citep{ronneberger2015unet} & 47,152,969 \\
UNet (5 Blocks)~\citep{ronneberger2015unet} & 399,563,977 \\

UNet (4 Layers - 128-dim initial projection)~\citep{ronneberger2015unet} & 188,520,649 \\
UNet (5 Blocks - 128-dim initial projection)~\citep{ronneberger2015unet} & 1,598,034,889 \\

GraphCast~\citep{lam2022graphcast} & 8,866,377 \\
Point Transformer v3~\citep{wu2023pointtranformerv3} & 46,439,017 \\
Octformer~\citep{wang2023octformer} & 40,497,325 \\
Graph UNet (4 Blocks)~\citep{ronneberger2015unet} & 47,377,417 \\
Graph UNet (5 Blocks)~\citep{ronneberger2015unet} & 399,788,425 \\
\bottomrule
\end{tabular}
\caption{\textbf{Total number of parameters in different models evaluated in our work.} Note that the number of parameters is not directly proportional to the memory footprint.
}
\label{tab:num_params}
\end{table*}

%% file: results_v3/direct_vs_delta_pred/direct_vs_delta_pred_ood_num_steps_test_loss.mean.tex
\begin{table*}[t]
\centering

\begin{subtable}{\textwidth}
\begin{adjustbox}{width=\linewidth,center}
\begin{tabular}{lccccccccccc}
\toprule
\multirow{2}{*}{Model} & \multicolumn{11}{c}{Forecasting horizon (hours)} \\
\cline{2-12}
 & 6 & 24 & 48 & 72 & 96 & 120 & 144 & 168 & 192 & 216 & 240 \\
\midrule
FNO (direct) & 0.9650 & 0.9231 & 0.7446 & 0.3922 & 0.2812 & 0.2580 & 0.2469 & 0.2205 & 0.2260 & 0.2306 & 0.2142 \\
FNO (delta) & 0.9722 & 0.9244 & 0.8571 & 0.7805 & 0.7010 & 0.6255 & 0.5573 & 0.4944 & 0.4334 & 0.3721 & 0.3121 \\
SFNO (direct) & 0.9743 & 0.9255 & 0.8478 & 0.7512 & 0.6432 & 0.5358 & 0.4396 & 0.3593 & 0.2939 & 0.2395 & 0.1930 \\
SFNO (delta) & 0.9801 & 0.9422 & 0.8879 & 0.8227 & 0.7482 & 0.6709 & 0.5988 & 0.5383 & 0.4888 & 0.4497 & 0.4189 \\
UNO (direct) & 0.9571 & 0.9012 & 0.7755 & 0.5427 & 0.2938 & 0.1734 & 0.1319 & 0.1128 & 0.0984 & 0.0841 & 0.0697 \\
UNO (delta) & 0.9701 & 0.9083 & 0.8126 & 0.6909 & 0.5337 & 0.3371 & 0.1625 & 0.0705 & 0.0435 & 0.0378 & 0.0370 \\
ResNet-50 + Conv Decoder (direct) & 0.9757 & 0.9506 & 0.9034 & 0.8465 & 0.7881 & 0.7353 & 0.6913 & 0.6565 & 0.6278 & 0.6053 & 0.5880 \\
ResNet-50 + Conv Decoder (delta) & 0.9801 & 0.9539 & 0.9150 & 0.8679 & 0.8177 & 0.7692 & 0.7257 & 0.6884 & 0.6572 & 0.6313 & 0.6094 \\
Segformer (direct) & 0.9780 & 0.9454 & 0.8894 & 0.8241 & 0.7582 & 0.6984 & 0.6481 & 0.6064 & 0.5691 & 0.5334 & 0.4961 \\
Segformer (delta) & \textbf{0.9825} & \textbf{0.9568} & 0.9149 & 0.8650 & 0.8132 & 0.7645 & 0.7215 & 0.6869 & \textbf{0.6604} & \textbf{0.6398} & \textbf{0.6236} \\
SETR (direct) & 0.9789 & 0.9448 & 0.8543 & 0.7349 & 0.6240 & 0.5375 & 0.4653 & 0.3996 & 0.3361 & 0.2710 & 0.2071 \\
SETR (delta) & 0.9789 & 0.9559 & \textbf{0.9211} & \textbf{0.8756} & \textbf{0.8248} & \textbf{0.7748} & \textbf{0.7302} & \textbf{0.6917} & 0.6598 & 0.6327 & 0.6102 \\
UNet (4 Blocks - direct) & 0.9760 & 0.9408 & 0.8725 & 0.7948 & 0.7215 & 0.6588 & 0.6090 & 0.5720 & 0.5449 & 0.5239 & 0.5064 \\
UNet (4 Blocks - delta) & 0.9820 & 0.9495 & 0.8965 & 0.8360 & 0.7768 & 0.7228 & 0.6767 & 0.6398 & 0.6100 & 0.5858 & 0.5655 \\
\bottomrule
\end{tabular}
\end{adjustbox}
\caption{Geometric ACC}
\end{subtable}

\begin{subtable}{\textwidth}
\begin{adjustbox}{width=\linewidth,center}
\begin{tabular}{lccccccccccc}
\toprule
\multirow{2}{*}{Model} & \multicolumn{11}{c}{Forecasting horizon (hours)} \\
\cline{2-12}
 & 6 & 24 & 48 & 72 & 96 & 120 & 144 & 168 & 192 & 216 & 240 \\
\midrule
FNO (direct) & 0.2602 & 0.3820 & 0.7381 & 1.7854 & 3.4580 & 5.7151 & 8.1361 & 12.5979 & 19.2537 & 26.5092 & 38.7152 \\
FNO (delta) & 0.2346 & 0.3867 & 0.5322 & 0.6768 & 0.8314 & 1.0052 & 1.2096 & 1.4664 & 1.8029 & 2.2588 & 2.8828 \\
SFNO (direct) & 0.1575 & 0.2694 & 0.3844 & 0.4922 & 0.5900 & 0.6728 & 0.7373 & 0.7862 & 0.8273 & 0.8675 & 0.9094 \\
SFNO (delta) & \textbf{0.1488} & \textbf{0.2506} & \textbf{0.3448} & \textbf{0.4295} & \textbf{0.5070} & \textbf{0.5751} & \textbf{0.6318} & \textbf{0.6755} & \textbf{0.7091} & \textbf{0.7348} & \textbf{0.7546} \\
UNO (direct) & 0.2860 & 0.4266 & 0.6987 & 1.4600 & 2.9173 & 4.8864 & 7.6792 & 12.3996 & 21.8269 & 43.2852 & 96.2574 \\
UNO (delta) & 0.2429 & 0.4231 & 0.6223 & 0.9225 & 1.8544 & 7.8484 & 65.0581 & 663.3524 & 7056.8694 & 71029.8833 & 683521.7333 \\
ResNet-50 + Conv Decoder (direct) & 0.2162 & 0.3058 & 0.4203 & 0.5234 & 0.6101 & 0.6789 & 0.7311 & 0.7703 & 0.8016 & 0.8257 & 0.8441 \\
ResNet-50 + Conv Decoder (delta) & 0.1992 & 0.3012 & 0.4041 & 0.5006 & 0.5871 & 0.6618 & 0.7248 & 0.7774 & 0.8214 & 0.8594 & 0.8938 \\
Segformer (direct) & 0.2077 & 0.3219 & 0.4579 & 0.5780 & 0.6784 & 0.7584 & 0.8204 & 0.8698 & 0.9137 & 0.9564 & 1.0010 \\
Segformer (delta) & 0.1863 & 0.2902 & 0.4044 & 0.5082 & 0.5988 & 0.6748 & 0.7372 & 0.7861 & 0.8240 & 0.8548 & 0.8802 \\
SETR (direct) & 0.2061 & 0.3324 & 0.5108 & 0.6576 & 0.7716 & 0.8703 & 0.9763 & 1.1057 & 1.2576 & 1.4070 & 1.5149 \\
SETR (delta) & 0.2079 & 0.2989 & 0.3945 & 0.4916 & 0.5815 & 0.6598 & 0.7247 & 0.7784 & 0.8228 & 0.8615 & 0.8951 \\
UNet (4 Blocks - direct) & 0.2151 & 0.3333 & 0.4771 & 0.5942 & 0.6848 & 0.7550 & 0.8087 & 0.8484 & 0.8781 & 0.9015 & 0.9217 \\
UNet (4 Blocks - delta) & 0.1871 & 0.3113 & 0.4397 & 0.5488 & 0.6376 & 0.7103 & 0.7689 & 0.8154 & 0.8535 & 0.8853 & 0.9127 \\
\bottomrule
\end{tabular}
\end{adjustbox}
\caption{Geometric RMSE}
\end{subtable}

\caption{\textbf{Geometric ACC (a) and RMSE (b) for comparison between direct and delta prediction formulations.} The table highlights that delta prediction is almost always superior in terms of performance as compared to direct prediction at the 6h prediction horizon (results in graphical form are presented in Fig.~\ref{fig:direct_vs_delta_pred}).}
\label{direct_vs_delta_pred_ood_num_steps_test_loss.mean}
\end{table*}

%% file: results_v3/delta_pred/delta_pred_ood_num_steps_test_loss.mean.tex
\begin{table*}[htbp]
\centering

\begin{subtable}{\textwidth}
\begin{adjustbox}{width=\linewidth,center}
\begin{tabular}{lccccccccccc}
\toprule
\multirow{2}{*}{Model} & \multicolumn{11}{c}{Forecasting horizon (hours)} \\
\cline{2-12}
 & 6 & 24 & 48 & 72 & 96 & 120 & 144 & 168 & 192 & 216 & 240 \\
\midrule
FNO & 0.9722 & 0.9244 & 0.8571 & 0.7805 & 0.7010 & 0.6255 & 0.5573 & 0.4944 & 0.4334 & 0.3721 & 0.3121 \\
FourCastNet & 0.9581 & 0.8040 & 0.2314 & 0.0420 & 0.0107 & 0.0032 & 0.0013 & 0.0008 & 0.0006 & 0.0008 & 0.0010 \\
SFNO & 0.9801 & 0.9422 & 0.8879 & 0.8227 & 0.7482 & 0.6709 & 0.5988 & 0.5383 & 0.4888 & 0.4497 & 0.4189 \\
UNO & 0.9701 & 0.9083 & 0.8126 & 0.6909 & 0.5337 & 0.3371 & 0.1625 & 0.0705 & 0.0435 & 0.0378 & 0.0370 \\
ISNet & 0.9806 & 0.9471 & 0.8957 & 0.8362 & 0.7745 & 0.7147 & 0.6610 & 0.6149 & 0.5753 & 0.5416 & 0.5127 \\
ResNet-50 + Conv Decoder & 0.9801 & 0.9539 & 0.9150 & 0.8679 & 0.8177 & 0.7692 & 0.7257 & 0.6884 & 0.6572 & 0.6313 & 0.6094 \\
Segformer & 0.9825 & \textbf{0.9568} & 0.9149 & 0.8650 & 0.8132 & 0.7645 & 0.7215 & 0.6869 & 0.6604 & 0.6398 & 0.6236 \\
SETR & 0.9789 & 0.9559 & \textbf{0.9211} & \textbf{0.8756} & 0.8248 & 0.7748 & 0.7302 & 0.6917 & 0.6598 & 0.6327 & 0.6102 \\
Swin Transformer & 0.9099 & 0.6219 & 0.4138 & 0.3061 & 0.2415 & 0.1988 & 0.1682 & 0.1453 & 0.1276 & 0.1139 & 0.1028 \\
Efficient ViT L2 & 0.9669 & 0.6210 & 0.2311 & 0.0836 & nan & nan & nan & nan & nan & nan & nan \\
PanguWeather & 0.9785 & 0.9191 & 0.8082 & 0.6861 & 0.5794 & 0.4841 & 0.3939 & 0.3065 & 0.2237 & 0.1503 & 0.0897 \\
UNet (4 Blocks) & 0.9820 & 0.9495 & 0.8965 & 0.8360 & 0.7768 & 0.7228 & 0.6767 & 0.6398 & 0.6100 & 0.5858 & 0.5655 \\
UNet (5 Blocks) & \textbf{0.9838} & 0.9554 & 0.9105 & 0.8581 & 0.8038 & 0.7517 & 0.7050 & 0.6656 & 0.6327 & 0.6049 & 0.5809 \\
GraphCast & 0.9672 & 0.8843 & 0.7801 & 0.7044 & 0.6575 & 0.6282 & 0.6070 & 0.5888 & 0.5723 & 0.5561 & 0.5397 \\
Point Transformer v3 & 0.9764 & 0.9404 & 0.8861 & 0.8243 & 0.7606 & 0.7000 & 0.6460 & 0.6007 & 0.5637 & 0.5340 & 0.5086 \\
Octformer & 0.9737 & 0.9340 & 0.8823 & 0.8294 & 0.7773 & 0.7299 & 0.6903 & 0.6586 & 0.6331 & 0.6115 & 0.5943 \\
Graph UNet (4 Blocks) & 0.9781 & 0.9412 & 0.8827 & 0.8185 & 0.7571 & 0.7040 & 0.6611 & 0.6278 & 0.6020 & 0.5814 & 0.5646 \\
Graph UNet (5 Blocks) & 0.9807 & 0.9492 & 0.9029 & 0.8508 & 0.7966 & 0.7461 & 0.7018 & 0.6649 & 0.6353 & 0.6110 & 0.5913 \\
\bottomrule
\end{tabular}
\end{adjustbox}
\caption{Geometric ACC}
\end{subtable}

\begin{subtable}{\textwidth}
\begin{adjustbox}{width=\linewidth,center}
\begin{tabular}{lccccccccccc}
\toprule
\multirow{2}{*}{Model} & \multicolumn{11}{c}{Forecasting horizon (hours)} \\
\cline{2-12}
 & 6 & 24 & 48 & 72 & 96 & 120 & 144 & 168 & 192 & 216 & 240 \\
\midrule
FNO & 0.2346 & 0.3867 & 0.5322 & 0.6768 & 0.8314 & 1.0052 & 1.2096 & 1.4664 & 1.8029 & 2.2588 & 2.8828 \\
FourCastNet & 0.2822 & 0.5932 & 2.8587 & 15.3645 & 54.4597 & 172.6959 & 531.8129 & 1542.8602 & 4298.7569 & 11823.4250 & 32522.8528 \\
SFNO & \textbf{0.1488} & \textbf{0.2506} & \textbf{0.3448} & \textbf{0.4295} & \textbf{0.5070} & \textbf{0.5751} & \textbf{0.6318} & \textbf{0.6755} & \textbf{0.7091} & \textbf{0.7348} & \textbf{0.7546} \\
UNO & 0.2429 & 0.4231 & 0.6223 & 0.9225 & 1.8544 & 7.8484 & 65.0581 & 663.3524 & 7056.8694 & 71029.8833 & 683521.7333 \\
ISNet & 0.1952 & 0.3186 & 0.4421 & 0.5511 & 0.6469 & 0.7319 & 0.8072 & 0.8745 & 0.9365 & 0.9941 & 1.0481 \\
ResNet-50 + Conv Decoder & 0.1992 & 0.3012 & 0.4041 & 0.5006 & 0.5871 & 0.6618 & 0.7248 & 0.7774 & 0.8214 & 0.8594 & 0.8938 \\
Segformer & 0.1863 & 0.2902 & 0.4044 & 0.5082 & 0.5988 & 0.6748 & 0.7372 & 0.7861 & 0.8240 & 0.8548 & 0.8802 \\
SETR & 0.2079 & 0.2989 & 0.3945 & 0.4916 & 0.5815 & 0.6598 & 0.7247 & 0.7784 & 0.8228 & 0.8615 & 0.8951 \\
Swin Transformer & 0.4481 & 1.3812 & 2.5692 & 3.7529 & 4.9411 & 6.1339 & 7.3301 & 8.5284 & 9.7282 & 10.9290 & 12.1305 \\
Efficient ViT L2 & 0.2496 & 3.1187 & 4112.0438 & 9155085.9130 & nan & nan & nan & nan & nan & nan & nan \\
PanguWeather & 0.2036 & 0.3854 & 0.5772 & 0.7563 & 0.9313 & 1.1134 & 1.3104 & 1.5293 & 1.8021 & 2.2216 & 3.0362 \\
UNet (4 Blocks) & 0.1871 & 0.3113 & 0.4397 & 0.5488 & 0.6376 & 0.7103 & 0.7689 & 0.8154 & 0.8535 & 0.8853 & 0.9127 \\
UNet (5 Blocks) & 0.1774 & 0.2933 & 0.4099 & 0.5109 & 0.5964 & 0.6683 & 0.7281 & 0.7773 & 0.8188 & 0.8561 & 0.9019 \\
GraphCast & 0.2507 & 0.4536 & 0.6064 & 0.6971 & 0.7545 & 0.7958 & 0.8298 & 0.8601 & 0.8870 & 0.9113 & 0.9338 \\
Point Transformer v3 & 0.2141 & 0.3378 & 0.4610 & 0.5696 & 0.6653 & 0.7481 & 0.8190 & 0.8792 & 0.9314 & 0.9783 & 1.0243 \\
Octformer & 0.2262 & 0.3574 & 0.4720 & 0.5652 & 0.6453 & 0.7131 & 0.7689 & 0.8140 & 0.8528 & 0.8880 & 0.9200 \\
Graph UNet (4 Blocks) & 0.2051 & 0.3353 & 0.4679 & 0.5789 & 0.6690 & 0.7401 & 0.7955 & 0.8386 & 0.8742 & 0.9042 & 0.9312 \\
Graph UNet (5 Blocks) & 0.1937 & 0.3129 & 0.4271 & 0.5250 & 0.6103 & 0.6809 & 0.7381 & 0.7836 & 0.8195 & 0.8481 & 0.8722 \\
\bottomrule
\end{tabular}
\end{adjustbox}
\caption{Geometric RSME}
\end{subtable}

\caption{\textbf{Geometric ACC (a) and RMSE (b) for comparison of different architectures when using the delta prediction formulation.} The table highlights that some architectures such as Segformer, SETR, ResNet-50 w/ convolutional decoder, and UNet are more efficient and effective for weather forecasting given a small and fixed number of training steps that we focused on in this work (results in graphical form are presented in Fig.~\ref{fig:delta_pred}).}
\label{delta_pred_ood_num_steps_test_loss.mean}
\end{table*}

%% file: results_v2/pretraining_4layers/pretraining_4layers_ood_num_steps_test_loss.mean.tex
\begin{table*}[t]
\centering

\begin{subtable}{\textwidth}
\begin{adjustbox}{width=\linewidth,center}
\begin{tabular}{lccccccccccc}
\toprule
\multirow{2}{*}{Model} & \multicolumn{11}{c}{Forecasting horizon (hours)} \\
\cline{2-12}
 & 6 & 24 & 48 & 72 & 96 & 120 & 144 & 168 & 192 & 216 & 240 \\
\midrule
UNet (4 Blocks) / 1-step finetuning after 1-step pretraining & \textbf{0.9839} & \textbf{0.9538} & 0.9032 & 0.8440 & 0.7850 & 0.7308 & 0.6839 & 0.6449 & 0.6131 & 0.5869 & 0.5655 \\
UNet (4 Blocks) / 2-step finetuning after 1-step pretraining & 0.9779 & 0.9515 & \textbf{0.9108} & \textbf{0.8654} & \textbf{0.8195} & \textbf{0.7771} & \textbf{0.7400} & \textbf{0.7090} & \textbf{0.6835} & \textbf{0.6636} & \textbf{0.6474} \\
UNet (4 Blocks) / 1-step finetuning after auto-encoder pretraining & 0.9758 & 0.9359 & 0.8779 & 0.8171 & 0.7602 & 0.7100 & 0.6674 & 0.6325 & 0.6041 & 0.5808 & 0.5615 \\
UNet (4 Blocks) / 2-step finetuning after auto-encoder pretraining & 0.9352 & 0.7085 & 0.5383 & 0.4534 & 0.3993 & 0.3574 & 0.3211 & 0.2885 & 0.2587 & 0.2317 & 0.2069 \\
UNet (4 Blocks) / 1-step finetuning after denoising auto-encoder pretraining & 0.9723 & 0.9305 & 0.8753 & 0.8161 & 0.7605 & 0.7119 & 0.6720 & 0.6401 & 0.6149 & 0.5945 & 0.5772 \\
UNet (4 Blocks) / 2-step finetuning after denoising auto-encoder pretraining & 0.9718 & 0.9357 & 0.8862 & 0.8350 & 0.7863 & 0.7433 & 0.7077 & 0.6803 & 0.6600 & 0.6437 & 0.6300 \\
UNet (4 Blocks) / 1-step finetuning after masked auto-encoder pretraining & 0.9764 & 0.9386 & 0.8829 & 0.8236 & 0.7675 & 0.7189 & 0.6787 & 0.6465 & 0.6211 & 0.5995 & 0.5815 \\
UNet (4 Blocks) / 2-step finetuning after masked auto-encoder pretraining & 0.9723 & 0.9387 & 0.8931 & 0.8450 & 0.7974 & 0.7531 & 0.7135 & 0.6792 & 0.6493 & 0.6230 & 0.6000 \\
\bottomrule
\end{tabular}
\end{adjustbox}
\caption{Geometric ACC}
\end{subtable}

\begin{subtable}{\textwidth}
\begin{adjustbox}{width=\linewidth,center}
\begin{tabular}{lccccccccccc}
\toprule
\multirow{2}{*}{Model} & \multicolumn{11}{c}{Forecasting horizon (hours)} \\
\cline{2-12}
 & 6 & 24 & 48 & 72 & 96 & 120 & 144 & 168 & 192 & 216 & 240 \\
\midrule
UNet (4 Blocks) / 1-step finetuning after 1-step pretraining & \textbf{0.1763} & \textbf{0.2975} & 0.4262 & 0.5371 & 0.6281 & 0.7023 & 0.7620 & 0.8098 & 0.8483 & 0.8802 & 0.9068 \\
UNet (4 Blocks) / 2-step finetuning after 1-step pretraining & 0.2064 & 0.3044 & \textbf{0.4060} & \textbf{0.4924} & \textbf{0.5646} & \textbf{0.6225} & \textbf{0.6687} & \textbf{0.7048} & \textbf{0.7332} & \textbf{0.7550} & \textbf{0.7728} \\
UNet (4 Blocks) / 1-step finetuning after auto-encoder pretraining & 0.2169 & 0.3497 & 0.4743 & 0.5747 & 0.6545 & 0.7183 & 0.7696 & 0.8111 & 0.8460 & 0.8761 & 0.9021 \\
UNet (4 Blocks) / 2-step finetuning after auto-encoder pretraining & 0.3539 & 0.6921 & 0.8613 & 0.9442 & 1.0011 & 1.0495 & 1.0951 & 1.1395 & 1.1835 & 1.2271 & 1.2704 \\
UNet (4 Blocks) / 1-step finetuning after denoising auto-encoder pretraining & 0.2310 & 0.3696 & 0.4865 & 0.5808 & 0.6559 & 0.7151 & 0.7608 & 0.7961 & 0.8246 & 0.8485 & 0.8700 \\
UNet (4 Blocks) / 2-step finetuning after denoising auto-encoder pretraining & 0.2338 & 0.3499 & 0.4571 & 0.5439 & 0.6144 & 0.6702 & 0.7130 & 0.7443 & 0.7672 & 0.7858 & 0.8020 \\
UNet (4 Blocks) / 1-step finetuning after masked auto-encoder pretraining & 0.2141 & 0.3438 & 0.4693 & 0.5726 & 0.6557 & 0.7215 & 0.7732 & 0.8141 & 0.8474 & 0.8769 & 0.9027 \\
UNet (4 Blocks) / 2-step finetuning after masked auto-encoder pretraining & 0.2321 & 0.3432 & 0.4449 & 0.5295 & 0.6015 & 0.6633 & 0.7172 & 0.7653 & 0.8109 & 0.8564 & 0.9018 \\
\bottomrule
\end{tabular}
\end{adjustbox}
\caption{Geometric RMSE}
\end{subtable}

\caption{\textbf{Geometric ACC (a) and RMSE (b) for comparison of different pretraining objectives on UNet with 4 blocks.} The table highlights that supervised pretraining achieves better performance in comparison to self-supervised pretraining using different objectives evaluated in our case (results in graphical form are presented in Fig.~\ref{fig:pretraining_4layers}).}
\label{pretraining_4layers_ood_num_steps_test_loss.mean}
\end{table*}

%% file: results_v2/pretraining_5layers/pretraining_5layers_ood_num_steps_test_loss.mean.tex
\begin{table*}[t]
\centering

\begin{subtable}{\textwidth}
\begin{adjustbox}{width=\linewidth,center}
\begin{tabular}{lccccccccccc}
\toprule
\multirow{2}{*}{Model} & \multicolumn{11}{c}{Forecasting horizon (hours)} \\
\cline{2-12}
 & 6 & 24 & 48 & 72 & 96 & 120 & 144 & 168 & 192 & 216 & 240 \\
\midrule
UNet (5 Blocks) / 1-step finetuning after 1-step pretraining & \textbf{0.9855} & 0.9593 & 0.9180 & 0.8696 & 0.8192 & 0.7696 & 0.7252 & 0.6869 & 0.6557 & 0.6305 & 0.6099 \\
UNet (5 Blocks) / 2-step finetuning after 1-step pretraining & 0.9827 & \textbf{0.9625} & \textbf{0.9286} & \textbf{0.8879} & \textbf{0.8453} & \textbf{0.8048} & \textbf{0.7685} & \textbf{0.7370} & \textbf{0.7104} & \textbf{0.6890} & \textbf{0.6717} \\
UNet (5 Blocks) / 1-step finetuning after auto-encoder pretraining & 0.9787 & 0.9440 & 0.8919 & 0.8328 & 0.7742 & 0.7211 & 0.6764 & 0.6410 & 0.6139 & 0.5929 & 0.5754 \\
UNet (5 Blocks) / 2-step finetuning after auto-encoder pretraining & 0.9745 & 0.9459 & 0.9075 & 0.8640 & 0.8186 & 0.7756 & 0.7380 & 0.7057 & 0.6794 & 0.6584 & 0.6413 \\
UNet (5 Blocks) / 1-step finetuning after denoising auto-encoder pretraining & 0.9798 & 0.9469 & 0.8971 & 0.8392 & 0.7809 & 0.7270 & 0.6807 & 0.6421 & 0.6109 & 0.5855 & 0.5637 \\
UNet (5 Blocks) / 2-step finetuning after denoising auto-encoder pretraining & 0.9724 & 0.9389 & 0.8927 & 0.8430 & 0.7929 & 0.7462 & 0.7050 & 0.6703 & 0.6414 & 0.6174 & 0.5969 \\
UNet (5 Blocks) / 1-step finetuning after masked auto-encoder pretraining & 0.9793 & 0.9458 & 0.8947 & 0.8367 & 0.7792 & 0.7266 & 0.6815 & 0.6447 & 0.6148 & 0.5904 & 0.5692 \\
UNet (5 Blocks) / 2-step finetuning after masked auto-encoder pretraining & 0.9750 & 0.9467 & 0.9082 & 0.8649 & 0.8199 & 0.7770 & 0.7382 & 0.7049 & 0.6768 & 0.6539 & 0.6348 \\
\bottomrule
\end{tabular}
\end{adjustbox}
\caption{Geometric ACC}
\end{subtable}

\begin{subtable}{\textwidth}
\begin{adjustbox}{width=\linewidth,center}
\begin{tabular}{lccccccccccc}
\toprule
\multirow{2}{*}{Model} & \multicolumn{11}{c}{Forecasting horizon (hours)} \\
\cline{2-12}
 & 6 & 24 & 48 & 72 & 96 & 120 & 144 & 168 & 192 & 216 & 240 \\
\midrule
UNet (5 Blocks) / 1-step finetuning after 1-step pretraining & \textbf{0.1681} & 0.2806 & 0.3946 & 0.4929 & 0.5759 & 0.6466 & 0.7043 & 0.7516 & 0.7897 & 0.8209 & 0.8472 \\
UNet (5 Blocks) / 2-step finetuning after 1-step pretraining & 0.1825 & \textbf{0.2693} & \textbf{0.3644} & \textbf{0.4480} & \textbf{0.5181} & \textbf{0.5752} & \textbf{0.6212} & \textbf{0.6582} & \textbf{0.6880} & \textbf{0.7113} & \textbf{0.7299} \\
UNet (5 Blocks) / 1-step finetuning after auto-encoder pretraining & 0.2036 & 0.3272 & 0.4476 & 0.5513 & 0.6373 & 0.7074 & 0.7635 & 0.8076 & 0.8428 & 0.8717 & 0.8970 \\
UNet (5 Blocks) / 2-step finetuning after auto-encoder pretraining & 0.2229 & 0.3226 & 0.4138 & 0.4942 & 0.5641 & 0.6218 & 0.6676 & 0.7044 & 0.7330 & 0.7549 & 0.7725 \\
UNet (5 Blocks) / 1-step finetuning after denoising auto-encoder pretraining & 0.1984 & 0.3189 & 0.4401 & 0.5486 & 0.6414 & 0.7201 & 0.7866 & 0.8436 & 0.8935 & 0.9393 & 0.9834 \\
UNet (5 Blocks) / 2-step finetuning after denoising auto-encoder pretraining & 0.2317 & 0.3421 & 0.4449 & 0.5310 & 0.6038 & 0.6646 & 0.7147 & 0.7557 & 0.7896 & 0.8186 & 0.8449 \\
UNet (5 Blocks) / 1-step finetuning after masked auto-encoder pretraining & 0.2014 & 0.3228 & 0.4435 & 0.5466 & 0.6311 & 0.6998 & 0.7547 & 0.7983 & 0.8338 & 0.8638 & 0.8910 \\
UNet (5 Blocks) / 2-step finetuning after masked auto-encoder pretraining & 0.2208 & 0.3204 & 0.4135 & 0.4953 & 0.5664 & 0.6263 & 0.6760 & 0.7163 & 0.7493 & 0.7758 & 0.7981 \\
\bottomrule
\end{tabular}
\end{adjustbox}
\caption{Geometric RMSE}
\end{subtable}

\caption{\textbf{Geometric ACC (a) and RMSE (b) for comparison of different pretraining objectives on UNet with 5 blocks.} The table highlights that supervised pretraining achieves better performance in comparison to self-supervised pretraining using different objectives evaluated in this work (results in graphical form are presented in Fig.~\ref{fig:pretraining_5layers}).}
\label{pretraining_5layers_ood_num_steps_test_loss.mean}
\end{table*}

%% file: results_v2/img_pretraining/img_pretraining_ood_num_steps_test_loss.mean.tex
\begin{table*}[t]
\centering

\begin{subtable}{\textwidth}
\begin{adjustbox}{width=\linewidth,center}
\begin{tabular}{lccccccccccc}
\toprule
\multirow{2}{*}{Model} & \multicolumn{11}{c}{Forecasting horizon (hours)} \\
\cline{2-12}
 & 6 & 24 & 48 & 72 & 96 & 120 & 144 & 168 & 192 & 216 & 240 \\
\midrule
Random Segformer / 1-step training & \textbf{0.9825} & \textbf{0.9568} & 0.9149 & 0.8650 & 0.8132 & 0.7645 & 0.7215 & 0.6869 & 0.6604 & 0.6398 & 0.6236 \\
Pretrained Segformer / 1-step training & 0.9811 & 0.9555 & \textbf{0.9209} & \textbf{0.8800} & \textbf{0.8352} & \textbf{0.7911} & \textbf{0.7517} & \textbf{0.7183} & \textbf{0.6904} & \textbf{0.6680} & \textbf{0.6496} \\
Random ResNet-50 v2 / 1-step training & 0.9801 & 0.9539 & 0.9150 & 0.8679 & 0.8177 & 0.7692 & 0.7257 & 0.6884 & 0.6572 & 0.6313 & 0.6094 \\
Pretrained ResNet-50 v2 / 1-step training & 0.9733 & 0.9353 & 0.8818 & 0.8253 & 0.7715 & 0.7241 & 0.6843 & 0.6517 & 0.6249 & 0.6014 & 0.5817 \\
\bottomrule
\end{tabular}
\end{adjustbox}
\caption{Geometric ACC}
\end{subtable}

\begin{subtable}{\textwidth}
\begin{adjustbox}{width=\linewidth,center}
\begin{tabular}{lccccccccccc}
\toprule
\multirow{2}{*}{Model} & \multicolumn{11}{c}{Forecasting horizon (hours)} \\
\cline{2-12}
 & 6 & 24 & 48 & 72 & 96 & 120 & 144 & 168 & 192 & 216 & 240 \\
\midrule
Random Segformer / 1-step training & \textbf{0.1863} & \textbf{0.2902} & 0.4044 & 0.5082 & 0.5988 & 0.6748 & 0.7372 & 0.7861 & 0.8240 & 0.8548 & 0.8802 \\
Pretrained Segformer / 1-step training & 0.1935 & 0.2942 & \textbf{0.3882} & \textbf{0.4740} & \textbf{0.5520} & \textbf{0.6189} & \textbf{0.6734} & \textbf{0.7166} & \textbf{0.7513} & \textbf{0.7785} & \textbf{0.8007} \\
Random ResNet-50 v2 / 1-step training & 0.1992 & 0.3012 & 0.4041 & 0.5006 & 0.5871 & 0.6618 & 0.7248 & 0.7774 & 0.8214 & 0.8594 & 0.8938 \\
Pretrained ResNet-50 v2 / 1-step training & 0.2292 & 0.3552 & 0.4748 & 0.5744 & 0.6564 & 0.7232 & 0.7779 & 0.8235 & 0.8629 & 0.8989 & 0.9304 \\
\bottomrule
\end{tabular}
\end{adjustbox}
\caption{Geometric RMSE}
\end{subtable}

\caption{\textbf{Geometric ACC (a) and RMSE (b) when using image-based pretrained models.} The table highlights that there are non-trivial gains for Segformer when using a pretrained model which was directly trained for segmentation as compared to ResNet-50 which was trained for classification (results in graphical form are presented in Fig.~\ref{fig:img_pretraining}).}
\label{img_pretraining_ood_num_steps_test_loss.mean}
\end{table*}

%% file: results_v2/multistep_inputs/multistep_inputs_ood_num_steps_test_loss.mean.tex
\begin{table*}[t]
\centering

\begin{subtable}{\textwidth}
\begin{adjustbox}{width=\linewidth,center}
\begin{tabular}{lccccccccccc}
\toprule
\multirow{2}{*}{Model} & \multicolumn{11}{c}{Forecasting horizon (hours)} \\
\cline{2-12}
 & 6 & 24 & 48 & 72 & 96 & 120 & 144 & 168 & 192 & 216 & 240 \\
\midrule
UNet (4 Blocks, and 1 input step) & 0.9820 & 0.9495 & 0.8965 & 0.8360 & 0.7768 & 0.7228 & 0.6767 & 0.6398 & 0.6100 & 0.5858 & 0.5655 \\
UNet (4 Blocks, and 2 input steps) & 0.9806 & 0.9468 & 0.8942 & 0.8357 & 0.7773 & 0.7240 & 0.6791 & 0.6426 & 0.6129 & 0.5884 & 0.5673 \\
UNet (4 Blocks, and 4 input steps) & 0.9812 & 0.9481 & 0.8930 & 0.8300 & 0.7671 & 0.7098 & 0.6607 & 0.6214 & 0.5896 & 0.5623 & 0.5385 \\
UNet (5 Blocks, and 1 input step) & \textbf{0.9838} & \textbf{0.9554} & \textbf{0.9105} & \textbf{0.8581} & \textbf{0.8038} & \textbf{0.7517} & \textbf{0.7050} & \textbf{0.6656} & \textbf{0.6327} & \textbf{0.6049} & \textbf{0.5809} \\
UNet (5 Blocks, and 2 input steps) & 0.9825 & 0.9526 & 0.9055 & 0.8513 & 0.7954 & 0.7431 & 0.6969 & 0.6579 & 0.6248 & 0.5978 & 0.5755 \\
UNet (5 Blocks, and 4 input steps) & 0.9825 & 0.9528 & 0.9041 & 0.8470 & 0.7882 & 0.7332 & 0.6845 & 0.6434 & 0.6076 & 0.5766 & 0.5501 \\
\bottomrule
\end{tabular}
\end{adjustbox}
\caption{Geometric ACC}
\end{subtable}

\begin{subtable}{\textwidth}
\begin{adjustbox}{width=\linewidth,center}
\begin{tabular}{lccccccccccc}
\toprule
\multirow{2}{*}{Model} & \multicolumn{11}{c}{Forecasting horizon (hours)} \\
\cline{2-12}
 & 6 & 24 & 48 & 72 & 96 & 120 & 144 & 168 & 192 & 216 & 240 \\
\midrule
UNet (4 Blocks, and 1 input step) & 0.1871 & 0.3113 & 0.4397 & 0.5488 & 0.6376 & 0.7103 & 0.7689 & 0.8154 & 0.8535 & 0.8853 & 0.9127 \\
UNet (4 Blocks, and 2 input steps) & 0.1951 & 0.3196 & 0.4438 & 0.5480 & 0.6358 & 0.7092 & 0.7696 & 0.8208 & 0.8665 & 0.9096 & 0.9533 \\
UNet (4 Blocks, and 4 input steps) & 0.1922 & 0.3165 & 0.4467 & 0.5578 & 0.6515 & 0.7294 & 0.7936 & 0.8456 & 0.8892 & 0.9275 & 0.9625 \\
UNet (5 Blocks, and 1 input step) & \textbf{0.1774} & \textbf{0.2933} & \textbf{0.4099} & \textbf{0.5109} & \textbf{0.5964} & \textbf{0.6683} & \textbf{0.7281} & \textbf{0.7773} & \textbf{0.8188} & \textbf{0.8561} & \textbf{0.9019} \\
UNet (5 Blocks, and 2 input steps) & 0.1854 & 0.3020 & 0.4209 & 0.5244 & 0.6147 & 0.6911 & 0.7555 & 0.8100 & 0.8574 & 0.8984 & 0.9356 \\
UNet (5 Blocks, and 4 input steps) & 0.1851 & 0.3019 & 0.4240 & 0.5306 & 0.6224 & 0.7001 & 0.7660 & 0.8218 & 0.8713 & 0.9158 & 0.9559 \\
\bottomrule
\end{tabular}
\end{adjustbox}
\caption{Geometric RMSE}
\end{subtable}

\caption{\textbf{Geometric ACC (a) and RMSE (b) with multi-step inputs.} Given the fixed number of training steps, we observe a negligible difference between 1 and 2 input steps starting with the next step prediction. However, performance significantly degrades when using 4 input steps despite achieving a similar starting performance to the 2 input step model for the next step i.e., 6h prediction (results in graphical form are presented in Fig.~\ref{fig:multi_step_inputs}).}
\label{multistep_inputs_ood_num_steps_test_loss.mean}
\end{table*}

%% file: results_v2/zenith_angle/zenith_angle_ood_num_steps_test_loss.mean.tex
\begin{table*}[t]
\centering

\begin{subtable}{\textwidth}
\begin{adjustbox}{width=\linewidth,center}
\begin{tabular}{lccccccccccc}
\toprule
\multirow{2}{*}{Model} & \multicolumn{11}{c}{Forecasting horizon (hours)} \\
\cline{2-12}
 & 6 & 24 & 48 & 72 & 96 & 120 & 144 & 168 & 192 & 216 & 240 \\
\midrule
UNet (4 Blocks) & 0.9820 & 0.9495 & 0.8965 & 0.8360 & 0.7768 & 0.7228 & 0.6767 & 0.6398 & 0.6100 & 0.5858 & 0.5655 \\
UNet (4 Blocks w/ Zenith angle) & 0.9826 & 0.9521 & 0.9032 & 0.8484 & 0.7947 & 0.7461 & 0.7050 & 0.6715 & \textbf{0.6446} & \textbf{0.6232} & \textbf{0.6063} \\
Circular Pad UNet (5 Blocks w/ 2 input steps) & 0.9824 & 0.9547 & 0.9161 & 0.8718 & 0.8229 & 0.7717 & 0.7224 & 0.6781 & 0.6393 & 0.6063 & 0.5771 \\
Circular Pad UNet (5 Blocks w/ Zenith angle, and 2 input steps) & \textbf{0.9846} & \textbf{0.9597} & \textbf{0.9246} & \textbf{0.8831} & \textbf{0.8352} & \textbf{0.7822} & \textbf{0.7293} & \textbf{0.6806} & 0.6381 & 0.6025 & 0.5727 \\
\bottomrule
\end{tabular}
\end{adjustbox}
\caption{Geometric ACC}
\end{subtable}

\begin{subtable}{\textwidth}
\begin{adjustbox}{width=\linewidth,center}
\begin{tabular}{lccccccccccc}
\toprule
\multirow{2}{*}{Model} & \multicolumn{11}{c}{Forecasting horizon (hours)} \\
\cline{2-12}
 & 6 & 24 & 48 & 72 & 96 & 120 & 144 & 168 & 192 & 216 & 240 \\
\midrule
UNet (4 Blocks) & 0.1871 & 0.3113 & 0.4397 & 0.5488 & 0.6376 & 0.7103 & 0.7689 & 0.8154 & 0.8535 & 0.8853 & 0.9127 \\
UNet (4 Blocks w/ Zenith angle) & 0.1844 & 0.3033 & 0.4250 & 0.5272 & 0.6113 & 0.6805 & 0.7370 & \textbf{0.7835} & \textbf{0.8222} & \textbf{0.8547} & \textbf{0.8826} \\
Circular Pad UNet (5 Blocks w/ 2 input steps) & 0.1858 & 0.2962 & 0.3993 & 0.4905 & 0.5753 & 0.6541 & \textbf{0.7248} & 0.7874 & 0.8436 & 0.8942 & 0.9418 \\
Circular Pad UNet (5 Blocks w/ Zenith angle, and 2 input steps) & \textbf{0.1740} & \textbf{0.2801} & \textbf{0.3802} & \textbf{0.4709} & \textbf{0.5584} & \textbf{0.6445} & 0.7259 & 0.8007 & 0.8676 & 0.9261 & 0.9774 \\
\bottomrule
\end{tabular}
\end{adjustbox}
\caption{Geometric RMSE}
\end{subtable}

\caption{\textbf{Geometric ACC (a) and RMSE (b) with the inclusion of zenith angle.} The table highlights that the Zenith angle provides consistent gains in predictive performance at both short-horizon and long-horizon forecasts (results in graphical form are presented in Fig.~\ref{fig:results_zenith}).}
\label{zenith_angle_ood_num_steps_test_loss.mean}
\end{table*}

%% file: results_v2/padding_models/padding_models_ood_num_steps_test_loss.mean.tex
\begin{table*}[t]
\centering

\begin{subtable}{\textwidth}
\begin{adjustbox}{width=\linewidth,center}
\begin{tabular}{lccccccccccc}
\toprule
\multirow{2}{*}{Model} & \multicolumn{11}{c}{Forecasting horizon (hours)} \\
\cline{2-12}
 & 6 & 24 & 48 & 72 & 96 & 120 & 144 & 168 & 192 & 216 & 240 \\
\midrule
UNet (4 Blocks w/ Zenith angle, and 2 input steps) & 0.9834 & 0.9536 & 0.9049 & 0.8490 & 0.7929 & 0.7417 & 0.6978 & 0.6617 & 0.6322 & 0.6084 & 0.5887 \\
Zero+Circular Pad UNet (4 Blocks w/ Zenith angle, and 2 input steps) & 0.9823 & 0.9531 & 0.9099 & 0.8601 & 0.8071 & 0.7553 & 0.7089 & 0.6690 & 0.6370 & 0.6107 & 0.5880 \\
Reflection+Circular Pad UNet (4 Blocks w/ Zenith angle, and 2 input steps) & 0.9812 & 0.9511 & 0.9084 & 0.8610 & 0.8114 & 0.7621 & 0.7165 & 0.6772 & 0.6454 & 0.6196 & \textbf{0.5981} \\
UNet (5 Blocks w/ Zenith angle, and 2 input steps) & 0.9844 & 0.9569 & 0.9115 & 0.8577 & 0.8020 & 0.7496 & 0.7031 & 0.6633 & 0.6306 & 0.6039 & 0.5821 \\
Zero+Circular Pad UNet (5 Blocks w/ Zenith angle, and 2 input steps) & \textbf{0.9846} & \textbf{0.9597} & \textbf{0.9246} & \textbf{0.8831} & \textbf{0.8352} & \textbf{0.7822} & 0.7293 & 0.6806 & 0.6381 & 0.6025 & 0.5727 \\
Reflection+Circular Pad UNet (5 Blocks w/ Zenith angle, and 2 input steps) & 0.9827 & 0.9558 & 0.9180 & 0.8744 & 0.8265 & 0.7777 & \textbf{0.7309} & \textbf{0.6889} & \textbf{0.6524} & \textbf{0.6207} & 0.5923 \\
\bottomrule
\end{tabular}
\end{adjustbox}
\caption{Geometric ACC}
\end{subtable}

\begin{subtable}{\textwidth}
\begin{adjustbox}{width=\linewidth,center}
\begin{tabular}{lccccccccccc}
\toprule
\multirow{2}{*}{Model} & \multicolumn{11}{c}{Forecasting horizon (hours)} \\
\cline{2-12}
 & 6 & 24 & 48 & 72 & 96 & 120 & 144 & 168 & 192 & 216 & 240 \\
\midrule
UNet (4 Blocks w/ Zenith angle, and 2 input steps) & 0.1808 & 0.2990 & 0.4214 & 0.5263 & 0.6143 & 0.6870 & 0.7462 & 0.7950 & 0.8366 & 0.8723 & 0.9043 \\
Zero+Circular Pad UNet (4 Blocks w/ Zenith angle, and 2 input steps) & 0.1869 & 0.3014 & 0.4141 & 0.5141 & 0.6049 & 0.6870 & 0.7608 & 0.8299 & 0.8968 & 0.9648 & 1.0346 \\
Reflection+Circular Pad UNet (4 Blocks w/ Zenith angle, and 2 input steps) & 0.1922 & 0.3076 & 0.4161 & 0.5091 & 0.5910 & 0.6636 & 0.7267 & 0.7804 & 0.8258 & 0.8652 & 0.9017 \\
UNet (5 Blocks w/ Zenith angle, and 2 input steps) & 0.1751 & 0.2885 & 0.4072 & 0.5108 & 0.5996 & 0.6747 & 0.7376 & 0.7911 & 0.8367 & 0.8760 & 0.9112 \\
Zero+Circular Pad UNet (5 Blocks w/ Zenith angle, and 2 input steps) & \textbf{0.1740} & \textbf{0.2801} & \textbf{0.3802} & \textbf{0.4709} & \textbf{0.5584} & 0.6445 & 0.7259 & 0.8007 & 0.8676 & 0.9261 & 0.9774 \\
Reflection+Circular Pad UNet (5 Blocks w/ Zenith angle, and 2 input steps) & 0.1845 & 0.2927 & 0.3942 & 0.4837 & 0.5649 & \textbf{0.6372} & \textbf{0.7008} & \textbf{0.7563} & \textbf{0.8056} & \textbf{0.8515} & \textbf{0.8963} \\
\bottomrule
\end{tabular}
\end{adjustbox}
\caption{Geometric RMSE}
\end{subtable}

\caption{\textbf{Geometric ACC (a) and RMSE (b) with different padding schemes for the convolutional layers in the UNet architecture.} The table highlights that using circular padding along the longitudinal direction and either zero or reflection padding along the latitudinal direction provides consistent gains in performance (results in graphical form are presented in Fig.~\ref{fig:results_padding_scheme}).}
\label{padding_models_ood_num_steps_test_loss.mean}
\end{table*}

%% file: results_v2/noise_models/noise_models_ood_num_steps_test_loss.mean.tex
\begin{table*}[t]
\centering

\begin{subtable}{\textwidth}
\begin{adjustbox}{width=\linewidth,center}
\begin{tabular}{lccccccccccc}
\toprule
\multirow{2}{*}{Model} & \multicolumn{11}{c}{Forecasting horizon (hours)} \\
\cline{2-12}
 & 6 & 24 & 48 & 72 & 96 & 120 & 144 & 168 & 192 & 216 & 240 \\
\midrule
Circular Pad UNet (4 Blocks w/ Zenith angle, and 2 input steps) & 0.9823 & 0.9531 & 0.9099 & 0.8601 & 0.8071 & 0.7553 & 0.7089 & 0.6690 & 0.6370 & 0.6107 & 0.5880 \\
Circular Pad UNet (4 Blocks w/ Zenith angle, 2 input steps, and Gaussian noise $\sigma=0.05$) & 0.9808 & 0.9498 & 0.9054 & 0.8560 & 0.8049 & 0.7553 & 0.7108 & 0.6722 & 0.6398 & 0.6116 & 0.5858 \\
Circular Pad UNet (4 Blocks w/ Zenith angle, 2 input steps, and Perlin noise $\sigma=0.05$) & 0.9828 & 0.9540 & 0.9102 & 0.8588 & 0.8043 & 0.7505 & 0.7019 & 0.6609 & 0.6269 & 0.5979 & 0.5739 \\
Circular Pad UNet (5 Blocks w/ Zenith angle, and 2 input steps) & \textbf{0.9846} & \textbf{0.9597} & \textbf{0.9246} & \textbf{0.8831} & \textbf{0.8352} & \textbf{0.7822} & 0.7293 & 0.6806 & 0.6381 & 0.6025 & 0.5727 \\
Circular Pad UNet (5 Blocks w/ Zenith angle, 2 input steps, and Gaussian noise $\sigma=0.05$) & 0.9821 & 0.9542 & 0.9156 & 0.8710 & 0.8219 & 0.7717 & 0.7247 & 0.6829 & 0.6472 & 0.6167 & 0.5902 \\
Circular Pad UNet (5 Blocks w/ Zenith angle, 2 input steps, and Perlin noise $\sigma=0.05$) & 0.9840 & 0.9583 & 0.9219 & 0.8792 & 0.8315 & 0.7814 & \textbf{0.7340} & \textbf{0.6914} & \textbf{0.6558} & \textbf{0.6273} & \textbf{0.6038} \\
\bottomrule
\end{tabular}
\end{adjustbox}
\end{subtable}

\begin{subtable}{\textwidth}
\begin{adjustbox}{width=\linewidth,center}
\begin{tabular}{lccccccccccc}
\toprule
\multirow{2}{*}{Model} & \multicolumn{11}{c}{Forecasting horizon (hours)} \\
\cline{2-12}
 & 6 & 24 & 48 & 72 & 96 & 120 & 144 & 168 & 192 & 216 & 240 \\
\midrule
Circular Pad UNet (4 Blocks w/ Zenith angle, and 2 input steps) & 0.1869 & 0.3014 & 0.4141 & 0.5141 & 0.6049 & 0.6870 & 0.7608 & 0.8299 & 0.8968 & 0.9648 & 1.0346 \\
Circular Pad UNet (4 Blocks w/ Zenith angle, 2 input steps, and Gaussian noise $\sigma=0.05$) & 0.1939 & 0.3116 & 0.4233 & 0.5202 & 0.6074 & 0.6866 & 0.7576 & 0.8222 & 0.8827 & 0.9421 & 1.0023 \\
Circular Pad UNet (4 Blocks w/ Zenith angle, 2 input steps, and Perlin noise $\sigma=0.05$) & 0.1841 & 0.2986 & 0.4111 & 0.5105 & 0.5983 & 0.6757 & 0.7417 & 0.7974 & 0.8461 & 0.8915 & 0.9354 \\
Circular Pad UNet (5 Blocks w/ Zenith angle, and 2 input steps) & \textbf{0.1740} & \textbf{0.2801} & \textbf{0.3802} & \textbf{0.4709} & \textbf{0.5584} & 0.6445 & 0.7259 & 0.8007 & 0.8676 & 0.9261 & 0.9774 \\
Circular Pad UNet (5 Blocks w/ Zenith angle, 2 input steps, and Gaussian noise $\sigma=0.05$) & 0.1871 & 0.2981 & 0.4006 & 0.4908 & 0.5734 & 0.6472 & 0.7110 & 0.7661 & 0.8140 & 0.8571 & 0.8974 \\
Circular Pad UNet (5 Blocks w/ Zenith angle, 2 input steps, and Perlin noise $\sigma=0.05$) & 0.1774 & 0.2848 & 0.3863 & 0.4773 & 0.5616 & \textbf{0.6390} & \textbf{0.7063} & \textbf{0.7636} & \textbf{0.8109} & \textbf{0.8494} & \textbf{0.8825} \\
\bottomrule
\end{tabular}
\end{adjustbox}
\caption{Geometric RMSE}
\end{subtable}

\caption{\textbf{Geometric ACC (a) and RMSE (b) with different noise addition schemes.} The table highlights that adding some forms of noise (particularly Perlin noise as used in prior work~\citep{bi2022pangu,chen2023fuxi}) improves predictive performance for the long-horizon case with only minor detrimental effects on the short-horizon forecast (results in graphical form are presented in Fig.~\ref{fig:results_noise_add}).}
\label{noise_models_ood_num_steps_test_loss.mean}
\end{table*}

%% file: results_v2/input_loc/input_loc_ood_num_steps_test_loss.mean.tex
\begin{table*}[t]
\centering

\begin{subtable}{\textwidth}
\begin{adjustbox}{width=\linewidth,center}
\begin{tabular}{lccccccccccc}
\toprule
\multirow{2}{*}{Model} & \multicolumn{11}{c}{Forecasting horizon (hours)} \\
\cline{2-12}
 & 6 & 24 & 48 & 72 & 96 & 120 & 144 & 168 & 192 & 216 & 240 \\
\midrule
Circular Pad UNet (4 Blocks w/ Zenith angle, and 2 input steps) & 0.9823 & 0.9531 & 0.9099 & 0.8601 & 0.8071 & 0.7553 & 0.7089 & 0.6690 & 0.6370 & 0.6107 & 0.5880 \\
Circular Pad UNet (4 Blocks w/ Zenith angle, 2 input steps, and Input loc) & 0.9806 & 0.9497 & 0.9051 & 0.8552 & 0.8030 & 0.7522 & 0.7067 & 0.6678 & 0.6357 & 0.6085 & 0.5852 \\
Circular Pad UNet (5 Blocks w/ Zenith angle, and 2 input steps) & \textbf{0.9846} & \textbf{0.9597} & \textbf{0.9246} & \textbf{0.8831} & \textbf{0.8352} & \textbf{0.7822} & 0.7293 & 0.6806 & 0.6381 & 0.6025 & 0.5727 \\
Circular Pad UNet (5 Blocks w/ Zenith angle, 2 input steps, and Input loc) & 0.9826 & 0.9557 & 0.9184 & 0.8755 & 0.8279 & 0.7788 & \textbf{0.7325} & \textbf{0.6913} & \textbf{0.6558} & \textbf{0.6247} & \textbf{0.5977} \\
\bottomrule
\end{tabular}
\end{adjustbox}
\caption{Geometric ACC}
\end{subtable}

\begin{subtable}{\textwidth}
\begin{adjustbox}{width=\linewidth,center}
\begin{tabular}{lccccccccccc}
\toprule
\multirow{2}{*}{Model} & \multicolumn{11}{c}{Forecasting horizon (hours)} \\
\cline{2-12}
 & 6 & 24 & 48 & 72 & 96 & 120 & 144 & 168 & 192 & 216 & 240 \\
\midrule
Circular Pad UNet (4 Blocks w/ Zenith angle, and 2 input steps) & 0.1869 & 0.3014 & 0.4141 & 0.5141 & 0.6049 & 0.6870 & 0.7608 & 0.8299 & 0.8968 & 0.9648 & 1.0346 \\
Circular Pad UNet (4 Blocks w/ Zenith angle, 2 input steps, and Input loc) & 0.1948 & 0.3114 & 0.4210 & 0.5136 & 0.5941 & 0.6633 & 0.7207 & \textbf{0.7687} & \textbf{0.8096} & \textbf{0.8464} & \textbf{0.8811} \\
Circular Pad UNet (5 Blocks w/ Zenith angle, and 2 input steps) & \textbf{0.1740} & \textbf{0.2801} & \textbf{0.3802} & \textbf{0.4709} & \textbf{0.5584} & 0.6445 & 0.7259 & 0.8007 & 0.8676 & 0.9261 & 0.9774 \\
Circular Pad UNet (5 Blocks w/ Zenith angle, 2 input steps, and Input loc) & 0.1848 & 0.2931 & 0.3934 & 0.4827 & 0.5660 & \textbf{0.6425} & \textbf{0.7099} & 0.7692 & 0.8223 & 0.8715 & 0.9179 \\
\bottomrule
\end{tabular}
\end{adjustbox}
\caption{Geometric RMSE}
\end{subtable}

\caption{\textbf{Geometric ACC (a) and RMSE (b) with the addition of 3D coordinates on the sphere as static masks.} The figure table that providing the model with this auxiliary information regarding the proximity of different points on the globe results in more stable long-horizon forecasts (results in graphical form are presented in Fig.~\ref{fig:results_input_loc}).}
\label{input_loc_ood_num_steps_test_loss.mean}
\end{table*}

%% file: results_v2/loss_models_4layers/loss_models_4layers_ood_num_steps_test_loss.mean.tex
\begin{table*}[t]
\centering

\begin{subtable}{\textwidth}
\begin{adjustbox}{width=\linewidth,center}
\begin{tabular}{lccccccccccc}
\toprule
\multirow{2}{*}{Model} & \multicolumn{11}{c}{Forecasting horizon (hours)} \\
\cline{2-12}
 & 6 & 24 & 48 & 72 & 96 & 120 & 144 & 168 & 192 & 216 & 240 \\
\midrule
Circular Pad UNet (4 Blocks w/ Zenith angle, 2 input steps, and MSE loss) & 0.9823 & 0.9531 & 0.9099 & 0.8601 & 0.8071 & 0.7553 & 0.7089 & 0.6690 & 0.6370 & 0.6107 & 0.5880 \\
Circular Pad UNet (4 Blocks w/ Zenith angle, 2 input steps, and L1 loss) & 0.9761 & 0.9419 & 0.8998 & 0.8548 & 0.8076 & 0.7609 & 0.7186 & 0.6842 & 0.6562 & 0.6324 & 0.6125 \\
Circular Pad UNet (4 Blocks w/ Zenith angle, 2 input steps, and Geometric MSE loss) & 0.9830 & 0.9538 & 0.9111 & 0.8619 & 0.8093 & 0.7568 & 0.7092 & 0.6677 & 0.6330 & 0.6039 & 0.5787 \\
Circular Pad UNet (4 Blocks w/ Zenith angle, 2 input steps, and Geometric L1 loss) & 0.9787 & 0.9467 & 0.9060 & 0.8622 & 0.8157 & \textbf{0.7688} & \textbf{0.7265} & \textbf{0.6914} & \textbf{0.6631} & \textbf{0.6401} & \textbf{0.6211} \\
Circular Pad UNet (4 Blocks w/ Zenith angle, 2 input steps, and Huber loss) & 0.9812 & 0.9505 & 0.9078 & 0.8589 & 0.8072 & 0.7564 & 0.7114 & 0.6746 & 0.6460 & 0.6222 & 0.6016 \\
Circular Pad UNet (4 Blocks w/ Zenith angle, 2 input steps, and L1-L2 loss) & \textbf{0.9834} & \textbf{0.9556} & \textbf{0.9155} & \textbf{0.8690} & \textbf{0.8185} & 0.7669 & 0.7192 & 0.6789 & 0.6469 & 0.6206 & 0.5976 \\
\bottomrule
\end{tabular}
\end{adjustbox}
\caption{Geometric ACC}
\end{subtable}

\begin{subtable}{\textwidth}
\begin{adjustbox}{width=\linewidth,center}
\begin{tabular}{lccccccccccc}
\toprule
\multirow{2}{*}{Model} & \multicolumn{11}{c}{Forecasting horizon (hours)} \\
\cline{2-12}
 & 6 & 24 & 48 & 72 & 96 & 120 & 144 & 168 & 192 & 216 & 240 \\
\midrule
Circular Pad UNet (4 Blocks w/ Zenith angle, 2 input steps, and MSE loss) & 0.1869 & 0.3014 & 0.4141 & 0.5141 & 0.6049 & 0.6870 & 0.7608 & 0.8299 & 0.8968 & 0.9648 & 1.0346 \\
Circular Pad UNet (4 Blocks w/ Zenith angle, 2 input steps, and L1 loss) & 0.2198 & 0.3403 & 0.4395 & 0.5231 & 0.5980 & 0.6646 & 0.7210 & 0.7658 & 0.8030 & 0.8356 & 0.8644 \\
Circular Pad UNet (4 Blocks w/ Zenith angle, 2 input steps, and Geometric MSE loss) & 0.1828 & 0.2980 & 0.4093 & 0.5076 & 0.5962 & 0.6758 & 0.7444 & 0.8043 & 0.8566 & 0.9043 & 0.9506 \\
Circular Pad UNet (4 Blocks w/ Zenith angle, 2 input steps, and Geometric L1 loss) & 0.2078 & 0.3247 & 0.4243 & 0.5089 & 0.5860 & \textbf{0.6555} & \textbf{0.7137} & \textbf{0.7600} & \textbf{0.7973} & \textbf{0.8281} & \textbf{0.8555} \\
Circular Pad UNet (4 Blocks w/ Zenith angle, 2 input steps, and Huber loss) & 0.1924 & 0.3099 & 0.4189 & 0.5145 & 0.5989 & 0.6721 & 0.7329 & 0.7826 & 0.8243 & 0.8627 & 0.9004 \\
Circular Pad UNet (4 Blocks w/ Zenith angle, 2 input steps, and L1-L2 loss) & \textbf{0.1810} & \textbf{0.2933} & \textbf{0.4005} & \textbf{0.4956} & \textbf{0.5818} & 0.6593 & 0.7259 & 0.7815 & 0.8268 & 0.8663 & 0.9037 \\
\bottomrule
\end{tabular}
\end{adjustbox}
\caption{Geometric RMSE}
\end{subtable}

\caption{\textbf{Geometric ACC (a) and RMSE (b) with the use of different loss functions on 4-block UNet.} The table highlights that variants of $L_1$ loss, including $L_1$, geometric $L_1$, $L_1$-$L_2$, and Huber loss, provide consistent gains in predictive performance in contrast to MSE, potentially due to their robustness against outliers (results in graphical form are presented in Fig.~\ref{fig:results_loss_fn_4layers}).}
\label{loss_models_4layers_ood_num_steps_test_loss.mean}
\end{table*}

%% file: results_v2/loss_models_5layers/loss_models_5layers_ood_num_steps_test_loss.mean.tex
\begin{table*}[t]
\centering

\begin{subtable}{\textwidth}
\begin{adjustbox}{width=\linewidth,center}
\begin{tabular}{lccccccccccc}
\toprule
\multirow{2}{*}{Model} & \multicolumn{11}{c}{Forecasting horizon (hours)} \\
\cline{2-12}
 & 6 & 24 & 48 & 72 & 96 & 120 & 144 & 168 & 192 & 216 & 240 \\
\midrule
Circular Pad UNet (5 Blocks w/ Zenith angle, 2 input steps, and MSE loss) & \textbf{0.9846} & \textbf{0.9597} & \textbf{0.9246} & \textbf{0.8831} & \textbf{0.8352} & 0.7822 & 0.7293 & 0.6806 & 0.6381 & 0.6025 & 0.5727 \\
Circular Pad UNet (5 Blocks w/ Zenith angle, 2 input steps, and L1 loss) & 0.9783 & 0.9480 & 0.9118 & 0.8720 & 0.8274 & 0.7807 & 0.7363 & 0.6976 & 0.6659 & \textbf{0.6403} & \textbf{0.6189} \\
Circular Pad UNet (5 Blocks w/ Zenith angle, 2 input steps, and Geometric MSE loss) & \textbf{0.9846} & 0.9591 & 0.9223 & 0.8789 & 0.8304 & 0.7799 & 0.7319 & 0.6891 & 0.6529 & 0.6222 & 0.5959 \\
Circular Pad UNet (5 Blocks w/ Zenith angle, 2 input steps, and Geometric L1 loss) & 0.9802 & 0.9510 & 0.9138 & 0.8722 & 0.8263 & 0.7788 & 0.7338 & 0.6951 & 0.6624 & 0.6352 & 0.6120 \\
Circular Pad UNet (5 Blocks w/ Zenith angle, 2 input steps, and Huber loss) & 0.9839 & 0.9581 & 0.9224 & 0.8807 & 0.8345 & \textbf{0.7870} & \textbf{0.7422} & \textbf{0.7024} & \textbf{0.6672} & 0.6358 & 0.6079 \\
Circular Pad UNet (5 Blocks w/ Zenith angle, 2 input steps, and L1-L2 loss) & 0.9841 & 0.9591 & 0.9239 & 0.8822 & 0.8351 & 0.7847 & 0.7358 & 0.6938 & 0.6579 & 0.6279 & 0.6019 \\
\bottomrule
\end{tabular}
\end{adjustbox}
\caption{Geometric ACC}
\end{subtable}

\begin{subtable}{\textwidth}
\begin{adjustbox}{width=\linewidth,center}
\begin{tabular}{lccccccccccc}
\toprule
\multirow{2}{*}{Model} & \multicolumn{11}{c}{Forecasting horizon (hours)} \\
\cline{2-12}
 & 6 & 24 & 48 & 72 & 96 & 120 & 144 & 168 & 192 & 216 & 240 \\
\midrule
Circular Pad UNet (5 Blocks w/ Zenith angle, 2 input steps, and MSE loss) & 0.1740 & \textbf{0.2801} & \textbf{0.3802} & \textbf{0.4709} & 0.5584 & 0.6445 & 0.7259 & 0.8007 & 0.8676 & 0.9261 & 0.9774 \\
Circular Pad UNet (5 Blocks w/ Zenith angle, 2 input steps, and L1 loss) & 0.2092 & 0.3219 & 0.4141 & 0.4951 & 0.5722 & 0.6440 & 0.7069 & 0.7597 & 0.8031 & \textbf{0.8393} & \textbf{0.8710} \\
Circular Pad UNet (5 Blocks w/ Zenith angle, 2 input steps, and Geometric MSE loss) & \textbf{0.1736} & 0.2816 & 0.3856 & 0.4788 & 0.5650 & 0.6436 & 0.7135 & 0.7751 & 0.8301 & 0.8840 & 0.9417 \\
Circular Pad UNet (5 Blocks w/ Zenith angle, 2 input steps, and Geometric L1 loss) & 0.2001 & 0.3106 & 0.4070 & 0.4923 & 0.5724 & 0.6460 & 0.7108 & 0.7654 & 0.8123 & 0.8529 & 0.8899 \\
Circular Pad UNet (5 Blocks w/ Zenith angle, 2 input steps, and Huber loss) & 0.1784 & 0.2853 & 0.3853 & 0.4746 & \textbf{0.5568} & \textbf{0.6305} & \textbf{0.6940} & \textbf{0.7483} & \textbf{0.7962} & 0.8402 & 0.8825 \\
Circular Pad UNet (5 Blocks w/ Zenith angle, 2 input steps, and L1-L2 loss) & 0.1767 & 0.2821 & 0.3826 & 0.4748 & 0.5623 & 3.3412 & 457.1086 & 2498.2460 & 12190.9917 & 58497.4833 & 279300.7111 \\
\bottomrule
\end{tabular}
\end{adjustbox}
\caption{Geometric RMSE}
\end{subtable}

\caption{\textbf{Geometric ACC (a) and RMSE (b) with the use of different loss functions on 5-block UNet.} In contrast to the 4-block UNet, MSE dominates in the short-horizon prediction, while variants of $L_{1}$ loss dominates in the long-horizon prediction (results in tabular form are presented in Fig.~\ref{fig:results_loss_fn_5layers}).}
\label{loss_models_5layers_ood_num_steps_test_loss.mean}
\end{table*}

%% file: results_v2/constant_masks/constant_masks_ood_num_steps_test_loss.mean.tex
\begin{table*}[t]
\centering

\begin{subtable}{\textwidth}
\begin{adjustbox}{width=\linewidth,center}
\begin{tabular}{lccccccccccc}
\toprule
\multirow{2}{*}{Model} & \multicolumn{11}{c}{Forecasting horizon (hours)} \\
\cline{2-12}
 & 6 & 24 & 48 & 72 & 96 & 120 & 144 & 168 & 192 & 216 & 240 \\
\midrule
Circular Pad UNet (4 Blocks w/ Zenith angle, and 2 input steps) & 0.9823 & 0.9531 & 0.9099 & 0.8601 & 0.8071 & 0.7553 & 0.7089 & 0.6690 & 0.6370 & 0.6107 & 0.5880 \\
Circular Pad UNet (4 Blocks w/ Zenith angle, 2 input steps, and constant masks) & 0.9810 & 0.9506 & 0.9079 & 0.8604 & 0.8106 & 0.7616 & 0.7172 & 0.6792 & \textbf{0.6479} & \textbf{0.6216} & \textbf{0.5980} \\
Circular Pad UNet (5 Blocks w/ Zenith angle, and 2 input steps) & \textbf{0.9846} & \textbf{0.9597} & \textbf{0.9246} & \textbf{0.8831} & \textbf{0.8352} & \textbf{0.7822} & \textbf{0.7293} & \textbf{0.6806} & 0.6381 & 0.6025 & 0.5727 \\
Circular Pad UNet (5 Blocks w/ Zenith angle, 2 input steps, and constant masks) & 0.9827 & 0.9559 & 0.9170 & 0.8703 & 0.8182 & 0.7649 & 0.7153 & 0.6711 & 0.6324 & 0.5993 & 0.5701 \\
\bottomrule
\end{tabular}
\end{adjustbox}
\caption{Geometric ACC}
\end{subtable}

\begin{subtable}{\textwidth}
\begin{adjustbox}{width=\linewidth,center}
\begin{tabular}{lccccccccccc}
\toprule
\multirow{2}{*}{Model} & \multicolumn{11}{c}{Forecasting horizon (hours)} \\
\cline{2-12}
 & 6 & 24 & 48 & 72 & 96 & 120 & 144 & 168 & 192 & 216 & 240 \\
\midrule
Circular Pad UNet (4 Blocks w/ Zenith angle, and 2 input steps) & 0.1869 & 0.3014 & 0.4141 & 0.5141 & 0.6049 & 0.6870 & 0.7608 & 0.8299 & 0.8968 & 0.9648 & 1.0346 \\
Circular Pad UNet (4 Blocks w/ Zenith angle, 2 input steps, and constant masks) & 0.1934 & 0.3086 & 0.4168 & 0.5114 & 0.5969 & 0.6740 & 0.7411 & \textbf{0.7997} & \textbf{0.8522} & \textbf{0.9025} & \textbf{0.9549} \\
Circular Pad UNet (5 Blocks w/ Zenith angle, and 2 input steps) & \textbf{0.1740} & \textbf{0.2801} & \textbf{0.3802} & \textbf{0.4709} & \textbf{0.5584} & \textbf{0.6445} & \textbf{0.7259} & 0.8007 & 0.8676 & 0.9261 & 0.9774 \\
Circular Pad UNet (5 Blocks w/ Zenith angle, 2 input steps, and constant masks) & 0.1840 & 0.2929 & 0.3979 & 0.4949 & 0.5875 & 0.6759 & 0.7599 & 0.8427 & 0.9273 & 1.0141 & 1.1044 \\
\bottomrule
\end{tabular}
\end{adjustbox}
\caption{Geometric RMSE}
\end{subtable}

\caption{\textbf{Geometric ACC (a) and RMSE (b) with the addition of static constant masks including topography, soil-type, and land-sea mask following~\citep{bi2022pangu}.} The table highlights an interesting trend where the performance is mostly hampered by the inclusion of constant masks when training for a small and fixed number of training steps as in our case (results in graphical form are presented in Fig.~\ref{fig:results_constant_masks}).}
\label{constant_masks_ood_num_steps_test_loss.mean}
\end{table*}

%% file: results_v2/wide/wide_ood_num_steps_test_loss.mean.tex
\begin{table*}[t]
\centering

\begin{subtable}{\textwidth}
\begin{adjustbox}{width=\linewidth,center}
\begin{tabular}{lccccccccccc}
\toprule
\multirow{2}{*}{Model} & \multicolumn{11}{c}{Forecasting horizon (hours)} \\
\cline{2-12}
 & 6 & 24 & 48 & 72 & 96 & 120 & 144 & 168 & 192 & 216 & 240 \\
\midrule
Circular Pad UNet (4 Blocks w/ 2 input steps, zenith angle, and 64 input lifting dims) & 0.9823 & 0.9531 & 0.9099 & 0.8601 & 0.8071 & 0.7553 & 0.7089 & 0.6690 & 0.6370 & 0.6107 & 0.5880 \\
Circular Pad UNet (4 Blocks w/ 2 input steps, zenith angle, and 128 input lifting dims) & 0.9829 & 0.9548 & 0.9138 & 0.8660 & 0.8146 & 0.7630 & 0.7158 & 0.6754 & 0.6411 & 0.6106 & 0.5834 \\
Circular Pad UNet (5 Blocks w/ 2 input steps, zenith angle, and 64 input lifting dims) & 0.9846 & 0.9597 & 0.9246 & 0.8831 & 0.8352 & 0.7822 & 0.7293 & 0.6806 & 0.6381 & 0.6025 & 0.5727 \\
Circular Pad UNet (5 Blocks w/ 2 input steps, zenith angle, and 128 input lifting dims) & \textbf{0.9867} & \textbf{0.9632} & \textbf{0.9288} & \textbf{0.8881} & \textbf{0.8419} & \textbf{0.7922} & \textbf{0.7430} & \textbf{0.6971} & \textbf{0.6576} & \textbf{0.6243} & \textbf{0.5961} \\
\bottomrule
\end{tabular}
\end{adjustbox}
\caption{Geometric ACC}
\end{subtable}

\begin{subtable}{\textwidth}
\begin{adjustbox}{width=\linewidth,center}
\begin{tabular}{lccccccccccc}
\toprule
\multirow{2}{*}{Model} & \multicolumn{11}{c}{Forecasting horizon (hours)} \\
\cline{2-12}
 & 6 & 24 & 48 & 72 & 96 & 120 & 144 & 168 & 192 & 216 & 240 \\
\midrule
Circular Pad UNet (4 Blocks w/ 2 input steps, zenith angle, and 64 input lifting dims) & 0.1869 & 0.3014 & 0.4141 & 0.5141 & 0.6049 & 0.6870 & 0.7608 & 0.8299 & 0.8968 & 0.9648 & 1.0346 \\
Circular Pad UNet (4 Blocks w/ 2 input steps, zenith angle, and 128 input lifting dims) & 0.1826 & 0.2951 & 0.4037 & 0.4991 & 0.5838 & 0.6579 & 0.7198 & 0.7705 & 0.8135 & \textbf{0.8523} & \textbf{0.8883} \\
Circular Pad UNet (5 Blocks w/ 2 input steps, zenith angle, and 64 input lifting dims) & 0.1740 & 0.2801 & 0.3802 & 0.4709 & 0.5584 & 0.6445 & 0.7259 & 0.8007 & 0.8676 & 0.9261 & 0.9774 \\
Circular Pad UNet (5 Blocks w/ 2 input steps, zenith angle, and 128 input lifting dims) & \textbf{0.1618} & \textbf{0.2667} & \textbf{0.3682} & \textbf{0.4586} & \textbf{0.5426} & \textbf{0.6205} & \textbf{0.6910} & \textbf{0.7533} & \textbf{0.8068} & 0.8530 & 0.8940 \\
\bottomrule
\end{tabular}
\end{adjustbox}
\caption{Geometric RMSE}
\end{subtable}

\caption{\textbf{Geometric ACC (a) and RMSE (b) with wider models i.e., larger hidden dimension.} We observe consistent gains with wider networks i.e., with an increased number of parameters for both 4-block and 5-block UNet variants (results in graphical form are presented in Fig.~\ref{fig:results_model_width}).}
\label{wide_ood_num_steps_test_loss.mean}
\end{table*}

%% file: results_v2/ft_4layers/ft_4layers_ood_num_steps_test_loss.mean.tex
\begin{table*}[t]
\centering

\begin{subtable}{\textwidth}
\begin{adjustbox}{width=\linewidth,center}
\begin{tabular}{lcccccccccc}
\toprule
\multirow{2}{*}{Model} & \multicolumn{10}{c}{Forecasting horizon (hours)} \\
\cline{2-11}
 & 24 & 48 & 72 & 96 & 120 & 144 & 168 & 192 & 216 & 240 \\
\midrule
1-step training of 6h model & 0.9531 & 0.9099 & 0.8601 & 0.8071 & 0.7553 & 0.7089 & 0.6690 & 0.6370 & 0.6107 & 0.5880 \\
2-step finetuning after 1-step pretraining of 6h model & 0.9550 & 0.9148 & 0.8673 & 0.8183 & 0.7721 & 0.7313 & 0.6976 & 0.6698 & 0.6473 & 0.6286 \\
2-step finetuning (w/ intermediate step supervision) after 1-step pretraining of 6h model & \textbf{0.9645} & \textbf{0.9413} & 0.9129 & 0.8786 & 0.8397 & 0.8000 & 0.7626 & 0.7295 & 0.7015 & 0.6779 \\
4-step finetuning after 1-step pretraining of 6h model & 0.9376 & 0.8840 & 0.8249 & 0.7700 & 0.7232 & 0.6863 & 0.6589 & 0.6383 & 0.6221 & 0.6092 \\
4-step finetuning (w/ intermediate step supervision) after 1-step pretraining of 6h model & 0.9379 & 0.8989 & 0.8578 & 0.8143 & 0.7716 & 0.7336 & 0.7027 & 0.6784 & 0.6592 & 0.6440 \\
4-step finetuning after 2-step finetuning of 1-step 6h model & 0.9617 & 0.9402 & \textbf{0.9144} & \textbf{0.8837} & \textbf{0.8491} & \textbf{0.8134} & \textbf{0.7790} & \textbf{0.7477} & \textbf{0.7202} & \textbf{0.6972} \\
4-step finetuning (w/ intermediate step supervision) after 2-step finetuning of 1-step 6h model & 0.9572 & 0.9303 & 0.8993 & 0.8634 & 0.8241 & 0.7853 & 0.7495 & 0.7173 & 0.6900 & 0.6684 \\
1-step training of 12h model & 0.9444 & 0.8693 & 0.7755 & 0.6953 & 0.6378 & 0.5992 & 0.5714 & 0.5493 & 0.5307 & 0.5157 \\
2-step finetuning after 1-step pretraining of 12h model & 0.9405 & 0.8894 & 0.8298 & 0.7708 & 0.7191 & 0.6770 & 0.6440 & 0.6182 & 0.5983 & 0.5816 \\
2-step finetuning (w/ intermediate step supervision) after 1-step pretraining of 12h model & 0.9567 & 0.9319 & 0.9012 & 0.8640 & 0.8226 & 0.7816 & 0.7444 & 0.7133 & 0.6883 & 0.6685 \\
1-step training of 24h model & 0.9453 & 0.8835 & 0.7825 & 0.6783 & 0.5945 & 0.5348 & 0.4948 & 0.4670 & 0.4453 & 0.4272 \\
\bottomrule
\end{tabular}
\end{adjustbox}
\caption{Geometric ACC}
\end{subtable}

\begin{subtable}{\textwidth}
\begin{adjustbox}{width=\linewidth,center}
\begin{tabular}{lcccccccccc}
\toprule
\multirow{2}{*}{Model} & \multicolumn{10}{c}{Forecasting horizon (hours)} \\
\cline{2-11}
 & 24 & 48 & 72 & 96 & 120 & 144 & 168 & 192 & 216 & 240 \\
\midrule
1-step training of 6h model & 0.3014 & 0.4141 & 0.5141 & 0.6049 & 0.6870 & 0.7608 & 0.8299 & 0.8968 & 0.9648 & 1.0346 \\
2-step finetuning after 1-step pretraining of 6h model & 0.2956 & 0.3982 & 0.4854 & 0.5564 & 0.6131 & 0.6580 & 0.6929 & 0.7212 & 0.7442 & 0.7638 \\
2-step finetuning (w/ intermediate step supervision) after 1-step pretraining of 6h model & \textbf{0.2638} & \textbf{0.3362} & 0.4064 & 0.4771 & 0.5458 & 0.6087 & 0.6635 & 0.7097 & 0.7471 & 0.7781 \\
4-step finetuning after 1-step pretraining of 6h model & 0.3472 & 0.4577 & 0.5426 & 0.6048 & 0.6503 & 0.6831 & 0.7068 & 0.7247 & 0.7389 & 0.7507 \\
4-step finetuning (w/ intermediate step supervision) after 1-step pretraining of 6h model & 0.3428 & 0.4293 & 0.5018 & 0.5670 & 0.6240 & 0.6706 & 0.7063 & 0.7338 & 0.7556 & 0.7737 \\
4-step finetuning after 2-step finetuning of 1-step 6h model & 0.2729 & 0.3367 & \textbf{0.3968} & \textbf{0.4564} & \textbf{0.5138} & \textbf{0.5662} & \textbf{0.6123} & \textbf{0.6517} & \textbf{0.6844} & \textbf{0.7107} \\
4-step finetuning (w/ intermediate step supervision) after 2-step finetuning of 1-step 6h model & 0.2871 & 0.3629 & 0.4330 & 0.5015 & 0.5674 & 0.6260 & 0.6766 & 0.7198 & 0.7553 & 0.7830 \\
1-step training of 12h model & 0.3293 & 0.5236 & 0.7180 & 0.8595 & 0.9496 & 1.0046 & 1.0406 & 1.0674 & 1.0886 & 1.1062 \\
2-step finetuning after 1-step pretraining of 12h model & 0.3395 & 0.4497 & 0.5391 & 0.6084 & 0.6598 & 0.6977 & 0.7261 & 0.7482 & 0.7657 & 0.7810 \\
2-step finetuning (w/ intermediate step supervision) after 1-step pretraining of 12h model & 0.2894 & 0.3589 & 0.4281 & 0.4986 & 0.5662 & 0.6256 & 0.6751 & 0.7139 & 0.7435 & 0.7662 \\
1-step training of 24h model & 0.3238 & 0.4804 & 0.6893 & 0.8655 & 0.9884 & 1.0692 & 1.1233 & 1.1637 & 1.1973 & 1.2257 \\
\bottomrule
\end{tabular}
\end{adjustbox}
\caption{Geometric RMSE}
\end{subtable}

\caption{\textbf{Geometric ACC (a) and RMSE (b) with multi-step fine-tuning for 4-block UNet.} The table highlights that training the model with the smallest stride considered i.e., 6h coupled with sequential fine-tuning provides the best performance, which has been the most common choice in the past (results in graphical form are presented in Fig.~\ref{fig:results_multistep_ft_4blocks}).}
\label{ft_4layers_ood_num_steps_test_loss.mean}
\end{table*}

%% file: results_v2/ft_5layers/ft_5layers_ood_num_steps_test_loss.mean.tex
\begin{table*}[t]
\centering

\begin{subtable}{\textwidth}
\begin{adjustbox}{width=\linewidth,center}
\begin{tabular}{lcccccccccc}
\toprule
\multirow{2}{*}{Model} & \multicolumn{10}{c}{Forecasting horizon (hours)} \\
\cline{2-11}
 & 24 & 48 & 72 & 96 & 120 & 144 & 168 & 192 & 216 & 240 \\
\midrule
1-step training of 6h model & 0.9597 & 0.9246 & 0.8831 & 0.8352 & 0.7822 & 0.7293 & 0.6806 & 0.6381 & 0.6025 & 0.5727 \\
2-step finetuning after 1-step pretraining of 6h model & 0.9602 & 0.9310 & 0.8946 & 0.8526 & 0.8080 & 0.7646 & 0.7259 & 0.6924 & 0.6642 & 0.6407 \\
2-step finetuning (w/ intermediate step supervision) after 1-step pretraining of 6h model & 0.9436 & 0.9041 & 0.8610 & 0.8152 & 0.7693 & 0.7282 & 0.6936 & 0.6645 & 0.6400 & 0.6204 \\
4-step finetuning after 1-step pretraining of 6h model & 0.9308 & 0.8589 & 0.7896 & 0.7328 & 0.6883 & 0.6529 & 0.6226 & 0.5954 & 0.5713 & 0.5482 \\
4-step finetuning (w/ intermediate step supervision) after 1-step pretraining of 6h model & 0.9315 & 0.8811 & 0.8293 & 0.7785 & 0.7276 & 0.6701 & 0.6005 & 0.4889 & 0.3479 & 0.2312 \\
4-step finetuning after 2-step finetuning of 1-step 6h model & 0.9578 & 0.9241 & 0.8816 & 0.8354 & 0.7900 & 0.7492 & 0.7140 & 0.6850 & 0.6611 & 0.6412 \\
4-step finetuning (w/ intermediate step supervision) after 2-step finetuning of 1-step 6h model & \textbf{0.9641} & \textbf{0.9443} & \textbf{0.9205} & \textbf{0.8911} & \textbf{0.8565} & \textbf{0.8197} & \textbf{0.7840} & \textbf{0.7516} & \textbf{0.7223} & \textbf{0.6975} \\
1-step training of 12h model & 0.9307 & 0.7934 & 0.5831 & 0.3909 & 0.2488 & 0.1539 & 0.0951 & 0.0590 & 0.0364 & 0.0213 \\
2-step finetuning after 1-step pretraining of 12h model & 0.9363 & 0.8795 & 0.8131 & 0.7519 & 0.7048 & 0.6727 & 0.6515 & 0.6355 & 0.6213 & 0.6081 \\
2-step finetuning (w/ intermediate step supervision) after 1-step pretraining of 12h model & 0.9541 & 0.9261 & 0.8911 & 0.8493 & 0.8047 & 0.7623 & 0.7253 & 0.6946 & 0.6703 & 0.6515 \\
1-step training of 24h model & 0.9486 & 0.8991 & 0.8102 & 0.7105 & 0.6270 & 0.5683 & 0.5312 & 0.5066 & 0.4890 & 0.4730 \\
\bottomrule
\end{tabular}
\end{adjustbox}
\caption{Geometric ACC}
\end{subtable}

\begin{subtable}{\textwidth}
\begin{adjustbox}{width=\linewidth,center}
\begin{tabular}{lcccccccccc}
\toprule
\multirow{2}{*}{Model} & \multicolumn{10}{c}{Forecasting horizon (hours)} \\
\cline{2-11}
 & 24 & 48 & 72 & 96 & 120 & 144 & 168 & 192 & 216 & 240 \\
\midrule
1-step training of 6h model & 0.2801 & 0.3802 & 0.4709 & 0.5584 & 0.6445 & 0.7259 & 0.8007 & 0.8676 & 0.9261 & 0.9774 \\
2-step finetuning after 1-step pretraining of 6h model & 0.2778 & 0.3628 & 0.4454 & 0.5248 & 0.5986 & 0.6642 & 0.7201 & 0.7684 & 0.8110 & 0.8499 \\
2-step finetuning (w/ intermediate step supervision) after 1-step pretraining of 6h model & 0.3312 & 0.4266 & 0.5092 & 0.5844 & 0.6519 & 0.7086 & 0.7554 & 0.7952 & 0.8294 & 0.8587 \\
4-step finetuning after 1-step pretraining of 6h model & 0.3674 & 0.5013 & 0.5876 & 0.6428 & 0.6800 & 0.7074 & 0.7298 & 0.7496 & 0.7671 & 0.7838 \\
4-step finetuning (w/ intermediate step supervision) after 1-step pretraining of 6h model & 0.3579 & 0.4612 & 0.5441 & 0.6139 & 0.6862 & 0.8232 & 1.1680 & 1.9843 & 3.6692 & 6.7553 \\
4-step finetuning after 2-step finetuning of 1-step 6h model & 0.2875 & 0.3780 & 0.4598 & 0.5293 & 0.5858 & 0.6302 & 0.6650 & 0.6923 & 0.7141 & \textbf{0.7321} \\
4-step finetuning (w/ intermediate step supervision) after 2-step finetuning of 1-step 6h model & \textbf{0.2646} & \textbf{0.3270} & \textbf{0.3871} & \textbf{0.4496} & \textbf{0.5129} & \textbf{0.5725} & \textbf{0.6253} & \textbf{0.6702} & \textbf{0.7086} & 0.7401 \\
1-step training of 12h model & 0.3660 & 0.6962 & 1.4075 & 2.9213 & 5.5888 & 10.0250 & 17.3233 & 29.3110 & 49.0948 & 82.0989 \\
2-step finetuning after 1-step pretraining of 12h model & 0.3482 & 0.4617 & 0.5544 & 0.6220 & 0.6670 & 0.6961 & 0.7167 & 0.7346 & 0.7524 & 0.7712 \\
2-step finetuning (w/ intermediate step supervision) after 1-step pretraining of 12h model & 0.2969 & 0.3717 & 0.4465 & 0.5217 & 0.5913 & 0.6508 & 0.6988 & 0.7367 & 0.7660 & 0.7898 \\
1-step training of 24h model & 0.3142 & 0.4450 & 0.6386 & 0.8227 & 0.9621 & 1.0566 & 1.1184 & 1.1622 & 1.1962 & 1.2279 \\
\bottomrule
\end{tabular}
\end{adjustbox}
\caption{Geometric RMSE}
\end{subtable}

\caption{\textbf{Geometric ACC (a) and RMSE (b) with multi-step fine-tuning for 5-block UNet.} In contrast to 4-block UNet, we see that 5-block UNet benefits from intermediate step supervision for sequential fine-tuning (results in graphical form are presented in Fig.~\ref{fig:results_multistep_ft_5blocks}).}
\label{ft_5layers_ood_num_steps_test_loss.mean}
\end{table*}

%% file: results_v2/compare_multistep_ft_models/compare_multistep_ft_models_ood_num_steps_test_loss.mean.tex
\begin{table*}[t]
\centering

\begin{subtable}{\textwidth}
\begin{adjustbox}{width=\linewidth,center}
\begin{tabular}{lcccccccccc}
\toprule
\multirow{2}{*}{Model} & \multicolumn{10}{c}{Forecasting horizon (hours)} \\
\cline{2-11}
 & 24 & 48 & 72 & 96 & 120 & 144 & 168 & 192 & 216 & 240 \\
\midrule
UNet (4 blocks) / 1-step training & 0.9531 & 0.9099 & 0.8601 & 0.8071 & 0.7553 & 0.7089 & 0.6690 & 0.6370 & 0.6107 & 0.5880 \\
UNet (4 blocks) / 2-step ft (1-step pretraining) & 0.9550 & 0.9148 & 0.8673 & 0.8183 & 0.7721 & 0.7313 & 0.6976 & 0.6698 & 0.6473 & 0.6286 \\
UNet (4 blocks) / 4-step ft (2-step ft + 1-step pt) & 0.9617 & 0.9402 & 0.9144 & 0.8837 & 0.8491 & 0.8134 & \textbf{0.7790} & \textbf{0.7477} & \textbf{0.7202} & \textbf{0.6972} \\
UNet (5 blocks) / 1-step training & 0.9597 & 0.9246 & 0.8831 & 0.8352 & 0.7822 & 0.7293 & 0.6806 & 0.6381 & 0.6025 & 0.5727 \\
UNet (5 blocks) / 2-step ft (1-step pretraining) & 0.9602 & 0.9310 & 0.8946 & 0.8526 & 0.8080 & 0.7646 & 0.7259 & 0.6924 & 0.6642 & 0.6407 \\
UNet (5 blocks) / 4-step ft (2-step ft + 1-step pt) & 0.9578 & 0.9241 & 0.8816 & 0.8354 & 0.7900 & 0.7492 & 0.7140 & 0.6850 & 0.6611 & 0.6412 \\
ResNet-50 + Conv Decoder / 1-step training & 0.9413 & 0.8909 & 0.8359 & 0.7828 & 0.7358 & 0.6965 & 0.6649 & 0.6399 & 0.6193 & 0.6008 \\
ResNet-50 + Conv Decoder / 2-step ft (1-step pretraining) & 0.9505 & 0.9121 & 0.8668 & 0.8194 & 0.7743 & 0.7335 & 0.6986 & 0.6697 & 0.6461 & 0.6278 \\
ResNet-50 + Conv Decoder / 4-step ft (2-step ft + 1-step pt) & 0.9518 & 0.9206 & 0.8839 & 0.8445 & 0.8056 & 0.7693 & 0.7371 & 0.7103 & 0.6880 & 0.6698 \\
Segformer / 1-step training & 0.9404 & 0.8705 & 0.7942 & 0.7299 & 0.6823 & 0.6480 & 0.6228 & 0.6031 & 0.5851 & 0.5676 \\
Segformer / 2-step ft (1-step pretraining) & 0.9589 & 0.9271 & 0.8881 & 0.8454 & 0.8032 & 0.7654 & 0.7336 & 0.7083 & 0.6887 & 0.6737 \\
Segformer / 4-step ft (2-step ft + 1-step pt) & \textbf{0.9631} & \textbf{0.9436} & \textbf{0.9185} & \textbf{0.8872} & \textbf{0.8512} & \textbf{0.8142} & 0.7781 & 0.7455 & 0.7174 & 0.6950 \\
\bottomrule
\end{tabular}
\end{adjustbox}
\caption{Geometric ACC}
\end{subtable}

\begin{subtable}{\textwidth}
\begin{adjustbox}{width=\linewidth,center}
\begin{tabular}{lcccccccccc}
\toprule
\multirow{2}{*}{Model} & \multicolumn{10}{c}{Forecasting horizon (hours)} \\
\cline{2-11}
 & 24 & 48 & 72 & 96 & 120 & 144 & 168 & 192 & 216 & 240 \\
\midrule
UNet (4 blocks) / 1-step training & 0.3014 & 0.4141 & 0.5141 & 0.6049 & 0.6870 & 0.7608 & 0.8299 & 0.8968 & 0.9648 & 1.0346 \\
UNet (4 blocks) / 2-step ft (1-step pretraining) & 0.2956 & 0.3982 & 0.4854 & 0.5564 & 0.6131 & 0.6580 & 0.6929 & 0.7212 & 0.7442 & 0.7638 \\
UNet (4 blocks) / 4-step ft (2-step ft + 1-step pt) & 0.2729 & 0.3367 & 0.3968 & \textbf{0.4564} & \textbf{0.5138} & \textbf{0.5662} & \textbf{0.6123} & \textbf{0.6517} & \textbf{0.6844} & \textbf{0.7107} \\
UNet (5 blocks) / 1-step training & 0.2801 & 0.3802 & 0.4709 & 0.5584 & 0.6445 & 0.7259 & 0.8007 & 0.8676 & 0.9261 & 0.9774 \\
UNet (5 blocks) / 2-step ft (1-step pretraining) & 0.2778 & 0.3628 & 0.4454 & 0.5248 & 0.5986 & 0.6642 & 0.7201 & 0.7684 & 0.8110 & 0.8499 \\
UNet (5 blocks) / 4-step ft (2-step ft + 1-step pt) & 0.2875 & 0.3780 & 0.4598 & 0.5293 & 0.5858 & 0.6302 & 0.6650 & 0.6923 & 0.7141 & 0.7321 \\
ResNet-50 + Conv Decoder / 1-step training & 0.3391 & 0.4559 & 0.5556 & 0.6387 & 0.7065 & 0.7616 & 0.8071 & 0.8457 & 0.8811 & 0.9173 \\
ResNet-50 + Conv Decoder / 2-step ft (1-step pretraining) & 0.3107 & 0.4073 & 0.4951 & 0.5718 & 0.6368 & 0.6911 & 0.7355 & 0.7721 & 0.8022 & 0.8262 \\
ResNet-50 + Conv Decoder / 4-step ft (2-step ft + 1-step pt) & 0.3055 & 0.3853 & 0.4586 & 0.5245 & 0.5815 & 0.6297 & 0.6697 & 0.7019 & 0.7281 & 0.7495 \\
Segformer / 1-step training & 0.3370 & 0.4879 & 0.6102 & 0.6987 & 0.7602 & 0.8048 & 0.8399 & 0.8699 & 0.8993 & 0.9297 \\
Segformer / 2-step ft (1-step pretraining) & 0.2818 & 0.3712 & 0.4562 & 0.5330 & 0.5987 & 0.6515 & 0.6926 & 0.7240 & 0.7476 & 0.7658 \\
Segformer / 4-step ft (2-step ft + 1-step pt) & \textbf{0.2675} & \textbf{0.3276} & \textbf{0.3911} & 0.4580 & 0.5245 & 0.5856 & 0.6404 & 0.6871 & 0.7255 & 0.7555 \\
\bottomrule
\end{tabular}
\end{adjustbox}
\caption{Geometric RMSE}
\end{subtable}

\caption{\textbf{Geometric ACC with multi-step fine-tuning for comparing other promising models alongside UNet including ResNet-50 w/ convolutional decoder and Segformer.} All models are trained with the zenith angle. We see that multi-step fine-tuning improves performance significantly for all models evaluated (results in graphical form are presented in Fig.~\ref{fig:results_multistep_ft}).}
\label{compare_multistep_ft_models_ood_num_steps_test_loss.mean}
\end{table*}

%% file: results_v2/short_term_compare_multistep_ft_models/short_term_compare_multistep_ft_models_ood_num_steps_test_loss.mean.tex
\begin{table*}[t]
\centering

\begin{subtable}{\textwidth}
\begin{adjustbox}{width=\linewidth,center}
\begin{tabular}{lcccccccccccccccc}
\toprule
\multirow{2}{*}{Model} & \multicolumn{16}{c}{Forecasting horizon (hours)} \\
\cline{2-17}
 & 6 & 12 & 18 & 24 & 30 & 36 & 42 & 48 & 54 & 60 & 66 & 72 & 78 & 84 & 90 & 96 \\
\midrule
UNet (4 blocks) / 1-step training & 0.9823 & 0.9709 & 0.9619 & 0.9531 & 0.9426 & 0.9318 & 0.9209 & 0.9099 & 0.8976 & 0.8851 & 0.8725 & 0.8601 & 0.8467 & 0.8333 & 0.8201 & 0.8071 \\
UNet (4 blocks) / 2-step ft (1-step pretraining) & 0.9795 & 0.9705 & 0.9621 & 0.9550 & 0.9449 & 0.9353 & 0.9248 & 0.9148 & 0.9026 & 0.8910 & 0.8788 & 0.8673 & 0.8544 & 0.8422 & 0.8298 & 0.8183 \\
UNet (4 blocks) / 4-step ft (2-step ft + 1-step pt) & 0.9781 & 0.9699 & 0.9650 & 0.9617 & 0.9552 & 0.9497 & 0.9446 & 0.9402 & 0.9333 & 0.9268 & 0.9204 & 0.9144 & 0.9065 & 0.8987 & 0.8910 & 0.8837 \\
UNet (5 blocks) / 1-step training & \textbf{0.9846} & \textbf{0.9751} & \textbf{0.9670} & 0.9597 & 0.9511 & 0.9424 & 0.9336 & 0.9246 & 0.9147 & 0.9043 & 0.8938 & 0.8831 & 0.8715 & 0.8596 & 0.8475 & 0.8352 \\
UNet (5 blocks) / 2-step ft (1-step pretraining) & 0.9805 & 0.9726 & 0.9655 & 0.9602 & 0.9525 & 0.9457 & 0.9381 & 0.9310 & 0.9219 & 0.9132 & 0.9038 & 0.8946 & 0.8841 & 0.8739 & 0.8631 & 0.8526 \\
UNet (5 blocks) / 4-step ft (2-step ft + 1-step pt) & 0.9776 & 0.9687 & 0.9625 & 0.9578 & 0.9487 & 0.9404 & 0.9320 & 0.9241 & 0.9131 & 0.9025 & 0.8918 & 0.8816 & 0.8695 & 0.8576 & 0.8461 & 0.8354 \\
ResNet-50 + Conv Decoder / 1-step training & 0.9763 & 0.9624 & 0.9515 & 0.9413 & 0.9285 & 0.9157 & 0.9032 & 0.8909 & 0.8767 & 0.8627 & 0.8491 & 0.8359 & 0.8217 & 0.8079 & 0.7951 & 0.7828 \\
ResNet-50 + Conv Decoder / 2-step ft (1-step pretraining) & 0.9762 & 0.9660 & 0.9575 & 0.9505 & 0.9404 & 0.9312 & 0.9213 & 0.9121 & 0.9003 & 0.8893 & 0.8776 & 0.8668 & 0.8540 & 0.8423 & 0.8302 & 0.8194 \\
ResNet-50 + Conv Decoder / 4-step ft (2-step ft + 1-step pt) & 0.9737 & 0.9627 & 0.9562 & 0.9518 & 0.9423 & 0.9342 & 0.9268 & 0.9206 & 0.9101 & 0.9008 & 0.8918 & 0.8839 & 0.8726 & 0.8624 & 0.8527 & 0.8445 \\
Segformer / 1-step training & 0.9794 & 0.9660 & 0.9536 & 0.9404 & 0.9241 & 0.9066 & 0.8887 & 0.8705 & 0.8508 & 0.8313 & 0.8124 & 0.7942 & 0.7764 & 0.7595 & 0.7442 & 0.7299 \\
Segformer / 2-step ft (1-step pretraining) & 0.9801 & 0.9719 & 0.9648 & 0.9589 & 0.9508 & 0.9433 & 0.9350 & 0.9271 & 0.9174 & 0.9079 & 0.8979 & 0.8881 & 0.8772 & 0.8666 & 0.8558 & 0.8454 \\
Segformer / 4-step ft (2-step ft + 1-step pt) & 0.9779 & 0.9702 & 0.9659 & \textbf{0.9631} & \textbf{0.9573} & \textbf{0.9524} & \textbf{0.9478} & \textbf{0.9436} & \textbf{0.9371} & \textbf{0.9309} & \textbf{0.9246} & \textbf{0.9185} & \textbf{0.9107} & \textbf{0.9028} & \textbf{0.8949} & \textbf{0.8872} \\
\bottomrule
\end{tabular}
\end{adjustbox}
\caption{Geometric ACC}
\end{subtable}

\begin{subtable}{\textwidth}
\begin{adjustbox}{width=\linewidth,center}
\begin{tabular}{lcccccccccccccccc}
\toprule
\multirow{2}{*}{Model} & \multicolumn{16}{c}{Forecasting horizon (hours)} \\
\cline{2-17}
 & 6 & 12 & 18 & 24 & 30 & 36 & 42 & 48 & 54 & 60 & 66 & 72 & 78 & 84 & 90 & 96 \\
\midrule
UNet (4 blocks) / 1-step training & 0.1869 & 0.2388 & 0.2728 & 0.3014 & 0.3325 & 0.3613 & 0.3884 & 0.4141 & 0.4406 & 0.4663 & 0.4910 & 0.5141 & 0.5381 & 0.5612 & 0.5836 & 0.6049 \\
UNet (4 blocks) / 2-step ft (1-step pretraining) & 0.1988 & 0.2402 & 0.2715 & 0.2956 & 0.3254 & 0.3507 & 0.3759 & 0.3982 & 0.4229 & 0.4448 & 0.4663 & 0.4854 & 0.5056 & 0.5236 & 0.5410 & 0.5564 \\
UNet (4 blocks) / 4-step ft (2-step ft + 1-step pt) & 0.2062 & 0.2416 & 0.2607 & 0.2729 & 0.2938 & 0.3102 & 0.3246 & 0.3367 & 0.3538 & 0.3693 & 0.3838 & 0.3968 & 0.4132 & 0.4285 & 0.4431 & \textbf{0.4564} \\
UNet (5 blocks) / 1-step training & \textbf{0.1740} & \textbf{0.2211} & \textbf{0.2536} & 0.2801 & 0.3078 & 0.3333 & 0.3573 & 0.3802 & 0.4038 & 0.4269 & 0.4494 & 0.4709 & 0.4935 & 0.5155 & 0.5372 & 0.5584 \\
UNet (5 blocks) / 2-step ft (1-step pretraining) & 0.1939 & 0.2311 & 0.2586 & 0.2778 & 0.3024 & 0.3227 & 0.3439 & 0.3628 & 0.3849 & 0.4051 & 0.4260 & 0.4454 & 0.4666 & 0.4862 & 0.5061 & 0.5248 \\
UNet (5 blocks) / 4-step ft (2-step ft + 1-step pt) & 0.2076 & 0.2463 & 0.2701 & 0.2875 & 0.3147 & 0.3378 & 0.3591 & 0.3780 & 0.4012 & 0.4224 & 0.4423 & 0.4598 & 0.4796 & 0.4978 & 0.5146 & 0.5293 \\
ResNet-50 + Conv Decoder / 1-step training & 0.2172 & 0.2733 & 0.3095 & 0.3391 & 0.3727 & 0.4029 & 0.4304 & 0.4559 & 0.4834 & 0.5092 & 0.5332 & 0.5556 & 0.5787 & 0.6003 & 0.6202 & 0.6387 \\
ResNet-50 + Conv Decoder / 2-step ft (1-step pretraining) & 0.2165 & 0.2594 & 0.2889 & 0.3107 & 0.3393 & 0.3629 & 0.3866 & 0.4073 & 0.4319 & 0.4537 & 0.4756 & 0.4951 & 0.5169 & 0.5361 & 0.5552 & 0.5718 \\
ResNet-50 + Conv Decoder / 4-step ft (2-step ft + 1-step pt) & 0.2263 & 0.2692 & 0.2915 & 0.3055 & 0.3321 & 0.3527 & 0.3707 & 0.3853 & 0.4077 & 0.4266 & 0.4441 & 0.4586 & 0.4786 & 0.4957 & 0.5115 & 0.5245 \\
Segformer / 1-step training & 0.2020 & 0.2576 & 0.2992 & 0.3370 & 0.3783 & 0.4172 & 0.4536 & 0.4879 & 0.5222 & 0.5539 & 0.5833 & 0.6102 & 0.6358 & 0.6590 & 0.6797 & 0.6987 \\
Segformer / 2-step ft (1-step pretraining) & 0.1977 & 0.2348 & 0.2614 & 0.2818 & 0.3070 & 0.3288 & 0.3509 & 0.3712 & 0.3940 & 0.4152 & 0.4365 & 0.4562 & 0.4771 & 0.4965 & 0.5155 & 0.5330 \\
Segformer / 4-step ft (2-step ft + 1-step pt) & 0.2074 & 0.2406 & 0.2571 & \textbf{0.2675} & \textbf{0.2865} & \textbf{0.3017} & \textbf{0.3155} & \textbf{0.3276} & \textbf{0.3449} & \textbf{0.3609} & \textbf{0.3765} & \textbf{0.3911} & \textbf{0.4089} & \textbf{0.4259} & \textbf{0.4425} & 0.4580 \\
\bottomrule
\end{tabular}
\end{adjustbox}
\caption{Geometric RMSE}
\end{subtable}

\caption{\textbf{Geometric ACC (a) and RMSE (b) when focusing on short-horizon forecast with multi-step fine-tuning for comparing other promising models alongside UNet including ResNet-50 w/ convolutional decoder and Segformer.} All models are trained with the zenith angle. We see that multi-step fine-tuning improves performance significantly for all models evaluated (results in graphical form are presented in Fig.~\ref{fig:results_short_term_multistep_ft}).}
\label{short_term_compare_multistep_ft_models_ood_num_steps_test_loss.mean}
\end{table*}

%% file: results_v2/hourly/hourly_ood_num_steps_test_loss.mean.tex
\begin{table*}[t]
\centering

\begin{subtable}{\textwidth}
\begin{adjustbox}{width=\linewidth,center}
\begin{tabular}{lccccccccccc}
\toprule
\multirow{2}{*}{Model} & \multicolumn{11}{c}{Forecasting horizon (hours)} \\
\cline{2-12}
 & 6 & 24 & 48 & 72 & 96 & 120 & 144 & 168 & 192 & 216 & 240 \\
\midrule
Circular Pad UNet (4 Blocks w/ 2 input steps, zenith angle, 6h training, 10 epochs) & 0.9823 & 0.9531 & 0.9099 & 0.8601 & 0.8071 & 0.7553 & 0.7089 & 0.6690 & 0.6370 & 0.6107 & 0.5880 \\
Circular Pad UNet (4 Blocks w/ 2 input steps, zenith angle, 6h training, 60 epochs) & 0.9568 & 0.9349 & 0.8964 & 0.8490 & 0.7939 & 0.7349 & 0.6793 & 0.6315 & 0.5905 & 0.5550 & 0.5255 \\
Circular Pad UNet (4 Blocks w/ 2 input steps, zenith angle, 1h training, 10 epochs) & 0.9869 & 0.9630 & 0.9252 & 0.8791 & 0.8275 & 0.7740 & 0.7241 & 0.6805 & 0.6443 & 0.6138 & 0.5885 \\
Circular Pad UNet (5 Blocks w/ 2 input steps, zenith angle, 6h training, 10 epochs) & 0.9846 & 0.9597 & 0.9246 & 0.8831 & 0.8352 & 0.7822 & 0.7293 & 0.6806 & 0.6381 & 0.6025 & 0.5727 \\
Circular Pad UNet (5 Blocks w/ 2 input steps, zenith angle, 6h training, 60 epochs) & \textbf{0.9892} & \textbf{0.9702} & \textbf{0.9426} & \textbf{0.9077} & \textbf{0.8651} & \textbf{0.8170} & \textbf{0.7681} & \textbf{0.7214} & \textbf{0.6795} & \textbf{0.6435} & \textbf{0.6123} \\
Circular Pad UNet (5 Blocks w/ 2 input steps, zenith angle, 1h training, 10 epochs) & 0.9887 & 0.9687 & 0.9397 & 0.9037 & 0.8609 & 0.8133 & 0.7649 & 0.7188 & 0.6771 & 0.6415 & 0.6122 \\
\bottomrule
\end{tabular}
\end{adjustbox}
\caption{Geometric ACC}
\end{subtable}

\begin{subtable}{\textwidth}
\begin{adjustbox}{width=\linewidth,center}
\begin{tabular}{lccccccccccc}
\toprule
\multirow{2}{*}{Model} & \multicolumn{11}{c}{Forecasting horizon (hours)} \\
\cline{2-12}
 & 6 & 24 & 48 & 72 & 96 & 120 & 144 & 168 & 192 & 216 & 240 \\
\midrule
Circular Pad UNet (4 Blocks w/ 2 input steps, zenith angle, 6h training, 10 epochs) & 0.1869 & 0.3014 & 0.4141 & 0.5141 & 0.6049 & 0.6870 & 0.7608 & 0.8299 & 0.8968 & 0.9648 & 1.0346 \\
Circular Pad UNet (4 Blocks w/ 2 input steps, zenith angle, 6h training, 60 epochs) & 0.3013 & 0.3906 & 0.5267 & 0.7242 & 1.0929 & 2.0225 & 4.5709 & 11.3957 & 27.4762 & 62.6718 & 142.4410 \\
Circular Pad UNet (4 Blocks w/ 2 input steps, zenith angle, 1h training, 10 epochs) & 0.1622 & 0.2722 & 0.3842 & 0.4853 & 0.5776 & 0.6606 & 0.7314 & 0.7905 & 0.8392 & 0.8803 & 0.9156 \\
Circular Pad UNet (5 Blocks w/ 2 input steps, zenith angle, 6h training, 10 epochs) & 0.1740 & 0.2801 & 0.3802 & 0.4709 & 0.5584 & 0.6445 & 0.7259 & 0.8007 & 0.8676 & 0.9261 & 0.9774 \\
Circular Pad UNet (5 Blocks w/ 2 input steps, zenith angle, 6h training, 60 epochs) & \textbf{0.1459} & \textbf{0.2416} & \textbf{0.3325} & \textbf{0.4171} & \textbf{0.5002} & \textbf{0.5794} & \textbf{0.6505} & \textbf{0.7128} & \textbf{0.7655} & \textbf{0.8088} & \textbf{0.8460} \\
Circular Pad UNet (5 Blocks w/ 2 input steps, zenith angle, 1h training, 10 epochs) & 0.1516 & 0.2509 & 0.3466 & 0.4365 & 0.5254 & 0.6115 & 0.6917 & 0.7648 & 0.8305 & 0.8876 & 0.9369 \\
\bottomrule
\end{tabular}
\end{adjustbox}
\caption{Geometric RMSE}
\end{subtable}

\caption{\textbf{Geometric ACC (a) and RMSE (b) when training on a larger dataset.} The table highlights that training on the smaller dataset for longer is detrimental to the performance of the 4-block UNet which shows signs of overfitting, but training on the larger dataset improves performance. On the other hand, 5-block UNet effectively achieves similar performance in both cases, highlighting that different models might benefit differently from increased training data (results in graphical form are presented in Fig.~\ref{fig:results_hourly}).}
\label{hourly_ood_num_steps_test_loss.mean}
\end{table*}

%% file: sections/conclusion.tex
\section{Conclusion}

In this paper, we attempt to analyze the role of different design choices on the performance of the resulting weather forecasting system.
Our analysis shows that (unsurprisingly) architectures have a very strong impact on performance, where fixed-grid models outperform their grid-invariant counterparts.
We further show that it is better to perform residual prediction when considering short-horizon prediction such as 6h considered in this work, regardless of the choice of architecture.
In addition, we analyzed the impact of pretraining objectives, the use of image-based pretrained models, the number of past input steps, loss functions for training, as well as multi-step fine-tuning aside from several other more trivial explorations.
We hope that these results would be useful in understanding the marginal contribution of different design choices for the training of weather forecasting systems, and hence, enable the development of better systems in the future.

%% file: sections/acknowledgements.tex
\section{Acknowledgements}

The authors would like to thank Pedram Hassanzadeh for his valuable suggestions, and Thorsten Kurth for help with the datasets.

%% file: sections/appendix.tex
\clearpage
\FloatBarrier

\appendix

\section{Hyperparameters}

In this section, we list all the architecture-specific hyperparameters.
See Section~\ref{sec:methods} in the main paper for information regarding training hyperparameters that were consistent across architectures.

For FNO~\citep{li2020fno}, SFNO~\citep{bonev2023sfno}, and UNO~\citep{ashiqur2022uno}, we use base modes to be $(237, 238)$.
We used domain padding $0.25$, the number of layers was 8, and the number of hidden channels was 64 as per our UNet~\citep{ronneberger2015unet} configuration.